\documentclass{article}
\usepackage[utf8]{inputenc}
\usepackage{authblk}
\usepackage[english]{babel}

\usepackage{amssymb}
\usepackage{amsmath}
\usepackage{graphicx} 
\usepackage{lscape}
\usepackage{longtable}
\usepackage{float}
\usepackage{caption}
\usepackage{subcaption}

\usepackage{mathtools} 
\usepackage{booktabs} 
\usepackage{tikz} 
\usepackage{lineno}
\usepackage{multirow}

\usepackage{fairness}

\graphicspath{ {images/} }

\usepackage{natbib} 
\begin{document}


\author[1]{Alexandra Cimpean}
\author[1]{Nicole Orzan} 
\author[2,3,4]{Catholijn Jonker}
\author[1,5]{Pieter Libin}
\author[1]{Ann Now\'{e}}
\affil[1]{\small Vrije Universiteit Brussel, Brussels, Belgium}
\affil[2]{Technische Universiteit Delft, Delft, The Netherlands}
\affil[3]{Vrije Universiteit Amsterdam, Amsterdam, The Netherlands}
\affil[4]{Universiteit Leiden, Leiden, The Netherlands}
\affil[5]{Universiteit Hasselt, Hasselt, Belgium}

\title{Fairness-Aware Reinforcement Learning (FAReL): A Framework for Transparent and Balanced Sequential Decision-Making} 

\date{}

\maketitle

\begin{abstract}
Equity in real-world sequential decision problems can be enforced using fairness-aware methods. Therefore, we require algorithms that can make suitable and transparent trade-offs between performance and the desired fairness notions. 
As the desired performance-fairness trade-off is hard to specify a priori, we propose a framework where multiple trade-offs can be explored. Insights provided by the reinforcement learning algorithm regarding the obtainable performance–fairness trade-offs can then guide stakeholders in selecting the most appropriate policy.
To capture fairness, we propose an extended Markov decision process, $\fmdpe$, that explicitly encodes individuals and groups. Given this $\fmdpe$, we formalise fairness notions in the context of sequential decision problems and formulate a fairness framework that computes fairness measures over time.
We evaluate our framework in two scenarios with distinct fairness requirements: job hiring, where strong teams must be composed while treating applicants equally, and fraud detection, where fraudulent transactions must be detected while ensuring the burden on customers is fairly distributed. We show that our framework learns policies that are more fair across multiple scenarios, with only minor loss in performance reward. Moreover, we observe that group and individual fairness notions do not necessarily imply one another, highlighting the benefit of our framework in settings where both fairness types are desired.
Finally, we provide guidelines on how to apply this framework across different problem settings.
\end{abstract}

\section{Introduction}

A wide range of real-world decision problems, such as job hiring \citep{Schumann2019,Schumann2020}, epidemic mitigation \citep{Libin2021,Ezekiel2020}, finance \citep{Liu2018} and fraud detection \citep{Soemers2018}, may pose risks of discrimination. When these problems are addressed using automated decision-making systems, it becomes essential to employ methods that incorporate fairness considerations. Additionally, decision problems that require a sequential approach—where decisions must adapt over time—demand automated decision support systems capable of maintaining high performance while adjusting to unseen situations. This highlights the need for automated decision support systems that learn policies capable of balancing both performance and fairness in a dynamic environment. Therefore, such decision problems require methods that take fairness into account. 
Moreover, problems that warrant a sequential approach require automated decision support systems to continuously maintain the desired performance and adapt as needed to unseen situations.
This demonstrates the need for automated decision support systems that learn policies that can balance performance with fairness over the impacted people.

Obtaining a policy with an appropriate performance-fairness trade-off is context-specific and typically relies on multiple fairness notions \citep{Makhlouf2020}.
For example, in a job hiring setting, fairness based on gender is required by law. This requires companies to hire suitable applicants that are qualified for the job, while ensuring all laws are followed.
Therefore, defining appropriate performance-fairness trade-offs requires input from stakeholders and domain experts. 
But, even with laws in place and input from domain experts, defining the balance between fairness, potential biases, and other ethical considerations remains challenging a priori.
Additionally, when a compromise is necessary, determining the extent to which performance can be sacrificed for fairness—or vice versa—is equally challenging to define in advance.
To address this challenge, a framework capable of accommodating multiple fairness notions simultaneously is needed, along with a means to provide stakeholders with a clear overview of possible trade-offs, enabling them to make informed decisions on the most suitable approach for the problem at hand.

Previous work related to fairness mainly focused on the supervised learning setting that operates on a given dataset, such as machine learning \citep{Makhlouf2020,Dwork2012,Mehrabi2019,Friedler2016,Dwork2020} and data mining \citep{Calmon2017,Kamiran2009,Feldman2015}.
In contrast, automated decision support problems are inherently sequential and evolve dynamically over time. This necessitates managing the impact of both short-term and long-term decisions \citep{DAmour2020}.
As such, reinforcement learning (RL) is a suitable approach to enable a decision-making agent to learn a support policy through interaction with its environment \citep{Sutton2018}.
The agent learns through trial and evaluation, by repeatedly interacting with the environment, balancing exploration and exploitation to converge toward an optimal policy \citep{Sutton2018}. Additionally, the agent must adapt to stochastic and non-stationary environments to maintain its performance over time.

%


RL approaches have primarily addressed fairness through application-specific solutions \citep{Joseph2016,Jabbari2017,Weng2019,Siddique2020,Chen2021,RodriguezSoto2021,Satija2023}. These solutions focus on a single fairness notion and rely on reward shaping to define the performance-fairness trade-off \citep{Liu2020, Chen2021}. However, as the desired performance-fairness trade-off is hard to describe a priori by stakeholders, these approaches do not suffice for most real-world problems. Furthermore, managing performance alongside one or multiple fairness notions—each of which may conflict with one another—requires a multi-objective approach to explore the uncertainty in the obtainable trade-offs \citep{Hayes2022}.
First, we highlight related work and situate our framework within the fairness landscape (Section~\ref{section:related_work}). 
Then, we propose a formal framework in Section~\ref{section:fairness_framework}, grounded in multi-objective reinforcement learning (MORL), that learns performance-fairness trade-offs accounting for multiple fairness notions. Section~\ref{section:fairness_framework} defines the fairness notions within the frmework, along with their applicability in terms of the agent-environment interactions. Next, we provide insights into different fairness perspectives. Section~\ref{section:scenarios} describes the two settings, inspired by real-world problems, that have distinct fairness requirements: job hiring and credit card fraud detection. We provide a guideline for implementing this fairness framework in Section~\ref{section:guidelines}, which we then use to define and setup the experiments in Section~\ref{section:experiments}. Finally, we provide a discussion (Section~\ref{Section:discussion}), highlighting the benefits of our framework, along with interesting venues for future work.

\section{Related Work}
\label{section:related_work}

Fairness approaches in RL are mainly focused on a single fairness notion definition, for application-specific solutions \citep{Joseph2016,Jabbari2017,Weng2019,Siddique2020,Chen2021,RodriguezSoto2021,Satija2023}, and often rely on reward shaping to define the performance-fairness trade-off \citep{Liu2020, Chen2021}. Moreover, recent work within the domain of multi-task learning proposes an optimisation objective, by incorporating fairness into the loss function \citep{Ban2024}. However, as the desired performance-fairness trade-off is hard to describe a priori by stakeholders, these approaches do not suffice for most real-world problems.
For example, \cite{Joseph2016,Jabbari2017} define fairness, in a multi-armed bandits and stateful RL setting respectively, as not preferring one action over another unless this action reads to a higher (possibly future) reward. However, this (weakly) meritocratic definition may be unsuitable in other settings. For example, in our fraud detection setting, flagging more transactions from a certain continent may not be appropriate based solely on that continent's higher number of fraudulent transactions.

Another approach in RL for resource allocation problems, focuses on specific fairness notions which are referred to as social welfare functions (SWF) \citep{Siddique2020,Weng2019,Qian2025}. Given a collection of individuals or groups, called users, an SWF requires that a fair agent is Pareto-optimal and equitable with the division of resources, while treating similar users similarly. SWFs have also been extended to settings based on preferences for each user \citep{Siddique2023}. These SWFs are based on multi-objective Markov Decision Processes, where each reward in the reward vector represents the utility of a user. While this implementation is a clear representation of how each user is impacted by the agent’s decision, this setting is only applicable for problems where there is a single objective, namely the SWF for the division of resources.
In the context of the allocation of indivisible items, \cite{Chakraborty2020} propose a weighted envy-based fairness concept, allowing multiple agents with different levels of entitlement over a resource to share in a fair manner.
\cite{Michailidis2024} propose a MORL method that learns a set of non-dominated policies across all objectives. Here fair policies are defined as the trade-offs that distribute rewards fairly across objectives. As in our work, they do not assume an explicit scalarisation of the reward function, thereby allowing decision makers to define their objective criteria post-training.
%

To enforce fairness in job hiring, multi-armed bandits \citep{Schumann2019,Schumann2020} as well as generalisations towards MDP approaches \citep{Jabbari2017} have been explored. However, current solutions do not employ the multi-objective approach that is necessary for learning an appropriate performance-fairness trade-off.
In contrast, \cite{Haydari2024} propose a method for environmentally
friendly traffic scheduling, taking fairness and air quality into account by employing a constrained MORL approach.

%

One fairness notion which focuses directly on explainability is counterfactual fairness. Counterfactual fairness \citep{Kusner2017} uses counterfactuals, established from causal inference \citep{Pearl2009}, to conclude if the outcome for a group or individual would have been the same if they belonged to a different group or had a different sensitive feature value. For example, if a woman is hired for a job with a given probability, counterfactual fairness checks whether she would be hired with the same probability if she were a man instead. Due to its reliance on causal relationships to express counterfactuals, counterfactual fairness requires that these causal relationships are known for the scenario at hand. This reduces its applicability to scenarios where this information can be extracted from the real world, or can be accurately simulated instead, requiring more computation.

\section{Fairness Framework}
\label{section:fairness_framework}

To define the fairness framework \citep{Cimpean2024,Cimpean2024b} in the context of RL, we outline and illustrate its requirements and suitability regarding two distinct real-world settings. We provide a detailed description of these settings in Section~\ref{section:scenarios}, which we will briefly introduce here.
The first setting concerns job hiring, where the aim is to hire qualified candidates while limiting bias towards sensitive features. Additionally, the ground truth regarding candidate performance on the job is only observable for those who are hired.
The second setting involves fraud detection, where fraudulent transactions must be efficiently identified, while considering that verification incurs a costly human effort. Moreover, the agent must accurately detect real fraudulent transactions to avoid inconveniencing genuine customers. Additionally, fraudulent transactions constitute rare events, rendering them challenging to detect. In this setting, the ground truth is only available for transactions that have been flagged for verification.
We emphasize that RL can be applied both directly or indirectly (i.e., through simulations) in the context of real-world problems. 
In this work, we use simulators based on real data distributions, to train the agent. 
We note that, in a deployed setting, the ground truth must be available for certain fairness notions. This availability could possibly be approximated using a supervised learning approach, trained on previous labeled data.

\subsection{\fmdp and fairness history}
\label{section:fairness_history}

A sequential decision process can be formally described as a Markov Decision Process (MDP) \citep{Puterman1994,Sutton2018}. An MDP consists of a tuple $\langle \setrlstates, \setactions, p, \setrewards, \gamma\rangle$, containing the set of states $\setrlstates$, a set of actions $\setactions$, a discount factor $\gamma$, a set of rewards $\setrewards$ and a transition function $p: \setrlstates \times \setrewards \times \setrlstates \times \setactions \rightarrow [0,1]$ describing the probability of a next state $\rlstate_{t+1}$ and reward $\reward_t$ given the current state $\rlstate_t$ and action $\action_t$.

As additional information, such as the presence of the ground truth regarding the correctness of an action, is required for some fairness notions, it must be either obtained through feedback or approximated based on previous interactions.
In the job hiring context, we can obtain this ground truth from hired applicants, by evaluating their performance on the job. In the fraud detection setting, a manual check of a flagged transactions would reveal if the it was indeed fraudulent or not.
To ensure that existing fairness notions can be defined in this sequential process, we propose the fairness MDP extension, which we will abbreviate as $\fmdpe$, defined by the tuple $\langle \setrlstates, \setactions, p, \setrewards, \gamma, \setfeedback \rangle$, with $\setfeedback$ the set of feedback signals. For example, a feedback signal $\feedback_t \in \setfeedback$ may indicate whether the $\action_t$ taken at time $t$ was correct, enabling the computation of fairness notions that require the ground truth.
Furthermore, this signal could contain information from the agent on the probability distribution over the actions for the given state. In the fraud detection setting, feedback on the probability of fraud could be provided through a pre-check.
Note that this feedback is optional and can be partial, sparse or delayed. For example, in the job hiring setting, we can collect a partial and delayed ground truth for job applicants who are hired, as they can be evaluated within the team. However, it is not possible to evaluate job applicants that were rejected.
We refer to an agent-environment interaction in an \fmdp as a tuple $\interaction{}$.

\paragraph{Individuals}
As existing fairness notions typically focus on ensuring fair treatment between individuals or groups, we introduce the following notation.
We will refer to the set of individuals who are involved in the decision process as $\ind$. We use $\indi, \indj \in \ind$ to refer to individuals of that set. Note that the set $\ind$ may change over time if not all individuals are known in advance and are encountered progressively. If the agent has not yet interacted with the environment at the start, this set will be empty.
In the job hiring setting, $\ind$ represents the set of candidates who have applied for the job and require a hiring decision (i.e., hire or reject). 
In the fraud detection setting, $\ind$ represents all customers the agent has encountered so far.

\paragraph{Groups}
To define groups of individuals, we first introduce the set of group characteristics $\group$, which includes at least one feature value that identifies a group. We use $\groupg, \grouph \in \group$ to refer to such group characteristics. For ease of notation, we assume that all possible group divisions are predefined and can be empty. In the job hiring setting, a group of characteristics $g$ can be a feature value $\{$gender: male$\}$, $\{$gender: female$\}$ or a combination of multiple feature values such as $\{$nationality: foreign, gender: female$\}$. Given a group characteristic $\groupg$, we refer to the group of individuals characterised by group $\groupg$ as $\groupindg$.
In the job hiring setting, $\groupindg$ refers to the group of men or women, who applied for a job. 
For the fraud detection setting, $\groupindg$ refers to individuals from the same continent for which the RL agent encounters transactions.
We assume that each individual belongs to at least one group, i.e., $\forall \indi \in \ind\quad\exists \groupg \in \group: \indi \in \ind_\groupg$.
Furthermore, groups are not mutually exclusive, i.e., an individual can belong to multiple distinct groups simultaneously. Note that, based on the current definition of group characteristics $\group$, the group of individuals $\groupindg$ is a subset of all individuals $\ind$.

\paragraph{A history of interactions}
To incorporate the temporal evolution of groups and individuals within the $\fmdpe$ framework, we define a history $\history$ over the agent-environment interactions. We use 
\begin{eqnarray}\label{eq:history}
    \history^{t' \rightarrow t} = \{\interaction{\tau} \}^{t}_{\tau=t'}
\end{eqnarray}
to refer to a history from time $t'$ until time $t$. For ease of notation, we denote the history of interactions at time $t$ as $\history^{t}$.
To improve the readability of the following equations, we introduce fairness notions in Section~\ref{section:fairness_notions} in terms of a history $\history$, where this history is defined over any time period and is implicitly equivalent to $\history^{t' \rightarrow t}$.
In the job hiring setting, the history consists of the encountered applicants and the agent’s corresponding hiring decisions. 
In the fraud detection setting, the history comprises all observed transactions along with whether they were flagged or ignored.

To define individuals and groups based on $\history$, we introduce a notation $\operator{}$ that extracts relevant information from a specified time window. Specifically, $\ind\operator{\history_x^{t'\rightarrow t}}$ denotes the set of individuals encountered in the history from time $t'$ until $t$, where $x \in \{S, A, F \}$ denotes if this history information concerns the states, actions or feedback signals. Analogously, $\ind_\groupg\operator{\history_x^{t'\rightarrow t}}$ refers to information about a group $\groupg$ encountered in the history from time $t'$ until time $t$.
We further expand on the use of $x \in \{S, A, F \}$ when defining fairness notions in the following section.
To account for potential non-stationarity in reinforcement learning environments, group membership is modeled as a time-stamped occurrence. Concretely, an individual’s group membership can change over time, allowing them to belong to different groups throughout the interaction history. For example, a job applicant may age before reapplying for the same job position. Similarly, in the context of fraud detection, a customer may conduct a transaction from a different continent while traveling.

\subsection{Fairness notions}
\label{section:fairness_notions}

We formally define a fairness notion $\fairnessnotion$ as a function that maps a history $\history$ to a non-positive scalar value:
\begin{eqnarray}\label{eq:fairness_notion}
    \fairnessnotion : \history \rightarrow \fairnessscalar^{-}
\end{eqnarray}
Concretely, each fairness notion $\fairnessnotion$ is defined as the negative absolute difference in treatment between groups or individuals.
Therefore, when the output of $\fairnessnotion$ is $0$, the agent has achieved exact fairness with respect to the given fairness notion.
An exact definition of $\fairnessnotion$ can be intractable due to limitations of defining exact fairness \citep{Jabbari2017}. In such cases, we propose to approximate it with $\approximatefairness$.
For a future fairness objective, $\fairnessnotion$, and by extension its approximation $\approximatefairness$ provide a foundation for a reward signal that can be used with a multi-objective RL approach.
The availability of a ground truth concerning the correctness of an action impacts which fairness notions can be calculated for a given scenario.
In the case of binary actions (e.g., hire or reject an applicant), the fairness notion is computed based on a confusion matrix, that is defined as a two-dimensional table, comparing predictions of a model to the actual values. It specifies the number of true positives ($TP$), false positives ($FP$), false negatives ($FN$) and true negatives ($TN$).

Group fairness is enforced through fairness notions that aim to ensure all groups receive equitable treatment relative to one another. If the goal is to ensure fair treatment for each individual, then individual fairness notions should be applied.
Since group fairness notions aim to treat groups similarly based on a set of sensitive features, they are unable to detect unfairness at the individual level, as they disregard all attributes other than the sensitive ones \citep{Dwork2012}. Similarly, individual fairness notions cannot ensure fairness across groups. Ideally, an RL agent conforms to a a combination of group and individual fairness notions to manage this trade-off, which can be managed using a multi-objective RL approach \citep{Hayes2022}. 

\paragraph{Group fairness}
Consider the group fairness notion \emph{statistical parity} \citep{Dwork2012}, where the probability of receiving the preferable treatment of the agent ($\actionhistory = \hpa$, i.e., all individuals want to be hired) should be the same across groups $g$ and $h$:
\begin{eqnarray}
\begin{aligned}\label{equation:statistical_parity}
    \fairnessnotion_{SP} = - | & \mathrm{P}(\ind_\groupg\operator{\actionhistory} = \hpa) 
    - \mathrm{P}(\ind_\grouph\operator{\actionhistory} = \hpa) |
\end{aligned}
\end{eqnarray}
Statistical parity requires that $(TP + FP ) / (TP + FP + FN + TN)$ is equal for both groups $g$ and $h$. 
In the job hiring setting, statistical parity requires that both men and women are hired with the same probability. Statistical parity is suitable when the aim is to treat different groups equally. However, it is susceptible to a self-fulfilling prophecy \citep{Dwork2012}, where decision-makers may randomly select individuals from a group in a way that reinforces existing negative biases. For example, under-qualified women might be hired to meet statistical parity, which could later be used to justify hiring more men, as women may be perceived as less qualified overall.
Because this fairness notion focuses on equal acceptance rate across groups, it can be expressed without knowledge of a ground truth. Other fairness notions require that a ground truth is (partially) known, such as \emph{equal opportunity}:
\begin{eqnarray} 
\begin{aligned}\label{equation:equal_opportunity}
    \fairnessnotion_{EO} = - | & \mathrm{P}(\ind_\groupg\operator{\actionhistory} = \hpa | \ind_\groupg\operator{\trueactionhistory} = \hpa) 
    \\ & 
    - \mathrm{P}(\ind_\grouph\operator{\actionhistory} = \hpa | \ind_\grouph\operator{\trueactionhistory} = \hpa) |
\end{aligned}
\end{eqnarray}
where $\trueactionhistory = \hpa$ corresponds to the correct action as specified by the feedback regarding the ground truth. 
Equal opportunity requires that the recall $TP / (TP + FN)$ is equal across groups and is consequently independent of $FP$. However, to compute it, we require a (partial) ground truth that informs us about $TP$ and $FN$. 
In the context of job hiring, this requires knowing how qualified each candidate is in order to compute the confusion matrix.
In fraud detection, a partial ground truth is available as a proportion of transactions flagged as fraudulent is manually reviewed, allowing the number of $TP$ and $FP$ to be determined. In contrast, unflagged transactions remain unverified unless random checks are conducted or individuals report experiencing fraud.

In the context of fraud detection, it may be desirable to ensure that transactions originating from different continents are treated similarly in terms of flagging.
For the group fairness notion \emph{overall accuracy equality} \citep{Berk2018}, the accuracy of the agent should be the same across the continents, here represented by the groups $\groupg$ and $\grouph$.
\begin{eqnarray} 
\begin{aligned}\label{equation:overall_accuracy_equality}
    \fairnessnotion_{OAE} = - | & \mathrm{P}(\ind_\groupg\operator{\actionhistory} = \ind_\groupg\operator{\trueactionhistory}) 
    \\ & 
    - \mathrm{P}(\ind_\grouph\operator{\actionhistory} = \ind_\grouph\operator{\trueactionhistory}) |
\end{aligned}
\end{eqnarray}

\emph{Predictive parity} \citep{Chouldechova2017} requires that the probability of being fraudulent, given that the agent requested a re-authentication, is the same across groups $\groupg$ and $\grouph$.
\begin{eqnarray} 
\begin{aligned}\label{equation:predictive_parity}
    \fairnessnotion_{PP} = - | & \mathrm{P}(\ind_\groupg\operator{\trueactionhistory} = \hpa | \ind_\groupg\operator{\actionhistory} = \hpa) 
    \\ & 
    - \mathrm{P}(\ind_\grouph\operator{\trueactionhistory} = \hpa | \ind_\grouph\operator{\actionhistory} = \hpa) |
\end{aligned}
\end{eqnarray}

The \emph{predictive equality} fairness notion \citep{Corbett.Davies2017} requires that the false positive error rate is the same across groups. In the job hiring setting, this would mean that unqualified men and women are hired with the same probability:
\begin{eqnarray} 
\begin{aligned}\label{equation:predictive_equality}
    \fairnessnotion_{PE} = - | & \mathrm{P}(\ind_\groupg\operator{\actionhistory} = \hpa | \ind_\groupg\operator{\trueactionhistory} = \hpaz) 
    \\ & 
    - \mathrm{P}(\ind_\grouph\operator{\actionhistory} = \hpa | \ind_\grouph\operator{\trueactionhistory} = \hpaz) |
\end{aligned}
\end{eqnarray}
where $\trueactionhistory = \hpaz$ is the correct action as specified by the feedback regarding the ground truth.

\paragraph{Individual fairness}

Individual fairness notions aim to treat similar individuals similarly, according to a distance metric \citep{Dwork2012}. Given two individuals $i$ and $j$, we assume a distance metric $d(i, j)$ between the individuals. Note that a similarity metric could be used instead as well. Given the probability distributions
$M_i$ and $M_j$ of the agent's policy over the actions for $i$ and $j$ respectively, and a distance metric $D(M_i \vert\vert M_j)$ between these probability distributions, 
\emph{individual fairness} requires:
\begin{eqnarray} 
    \begin{aligned}
    \label{equation:individual_fairness}
    & \fairnessnotion_{IF} = -1 + \frac{1}{n} \sum_{\indi \in \ind\operator{\history}} \sum_{\indj \in \ind\operator{\history}, j \neq i} d(i, j) - D(M_i \vert\vert M_j) 
    \end{aligned}
\end{eqnarray}
where $n$ denotes the number of pairwise combinations of individuals that can be made.

Individual fairness notions assume that an appropriate distance metric is chosen based on domain expertise \citep{Makhlouf2020}. 
To evaluate our framework, we selected distance metrics where we exclude the sensitive features, such that similarity is defined only in terms of non-sensitive features. This ensures that sensitive features are not used in the similarity computation, which could otherwise lead to unfair decisions being justified by allowing individuals to differ in similarity due to their sensitive features. Ultimately, we stress that stakeholders must make this decision when deploying algorithms in real-world contexts.
We employ three different distance metrics. The first one is the Bray-Curtis distance \citep{BrayCurtis1957}, which requires for two feature vectors $i$ and $j$, each with $n$ features, that
\begin{eqnarray} 
    \begin{aligned}
    \label{equation:braycurtis}
    & \mathrm{braycurtis}(i, j) = \frac{\sum_{k=1}^n |i_k - j_k|}{\sum_{k=1}^n |i_k + j_k|}
    \end{aligned}
\end{eqnarray}
This distance metric's output range is $[0, 1]$ if all features are positive, as is the case in our experiments.

To distinguish between nominal features (e.g., ability to speak a language) and numerical features (e.g., years of experience), we employ the Heterogeneous Manhattan-Overlap Metric (HMOM) and the Heterogeneous Euclidean-Overlap Metric (HEOM) \citep{Wilson1997}. 
By using a heterogeneous approach, the distance is computed using the original formula only for the numerical features, while the nominal features are only checked for the number of inequalities.
Concretely, computing HMOM and HEOM goes as follows:
\begin{eqnarray} 
    \begin{aligned}
    \label{equation:HEOM}
    & \mathrm{HEOM}(i, j) = \begin{cases}
        \sum_{k=1}^n\sqrt{(i_k - j_k)^2} & \mathrm{if}\ k\ \mathrm{is\ a\ numerical\ feature},\\
        \sum_{k=1}^n(i_k \neq j_k) & \mathrm{otherwise}
        \end{cases}
    \end{aligned}
\end{eqnarray}
\begin{eqnarray} 
    \begin{aligned}
    \label{equation:HMOM}
    & \mathrm{HMOM}(i, j) = \begin{cases}
        \sum_{k=1}^n|i_k - j_k| & \mathrm{if}\ k\ \mathrm{is\ a\ numerical\ feature},\\
        \sum_{k=1}^n(i_k \neq j_k) & \mathrm{otherwise}
        \end{cases}
    \end{aligned}
\end{eqnarray}
We define a distance metric using HEOM and HMOM in the interval $[0, 1]$ as:
\begin{eqnarray} 
    \label{equation:h_distance_metric}
    d(i, j) = e^{-\lambda \mathrm{HEOM}(i, j)}\\
    d(i, j) = e^{-\lambda \mathrm{HMOM}(i, j)}
\end{eqnarray}
where $\lambda>0$ is a smoothing parameter. We use an exponential function to output values in the range $[0, 1]$, such that all compared individuals have a normalised impact on the outcome of the individual fairness notion.

In fraud detection, we define individual fairness between transactions using the complement of the \emph{consistency score} \citep{Zemel2013}:
\begin{eqnarray} 
\begin{aligned}\label{equation:consistency_score_complement}
    \fairnessnotion_{CSC} = - \frac{1}{||\ind\operator{\history}||} \sum_{\indi \in \ind\operator{\history}} \frac{1}{k} | \action^\indi - \sum_{\indj \in kNN(\indi)} \action^\indj |
\end{aligned}
\end{eqnarray}
given action $\action^\indi$ for an individual $\indi$, where $k$ is the number of nearest neighbours to consider, given a $k$-nearest neighbour algorithm $kNN$ \citep{Mitchell1997}. We assume $\lambda=0.1$ for both heterogeneous distance metrics and $k=5$ for the number of nearest neighbours.

In our implementation, we have defined all the fairness notions within $\fairnessscalar^{-}$, to create an objective to be maximised by an RL agent. Furthermore, we normalise all fairness notions to output values in the range $[-1, 0]$. Note that not all fairness notions output values in $[-1, 0]$ by default. For example, individual fairness depends on the range of the distance metric, whereas group fairness notions such as treatment equality (i.e., FN / FP should be equal across groups) output values in $\mathbb{R}^+$.

While not explicitly defined for group fairness, individual fairness notions such as individual fairness \citep{Dwork2012} can be translated to a group setting, by applying a distance metric aimed at comparing people based on the groups they belong to, rather than on an individual level.

In the context of fairness, group fairness notions often rely on predefined groups. As such, these group notions do not guarantee fairness among any further subgroup divisions. Therefore, it is possible for an algorithm to learn a fair policy for the given groups, while being unfair for subgroups. \citet{Kearns2018} propose a technique to deal with this phenomenon, which is known as gerrymandering. They highlight the need for a more extensive fairness evaluation when it comes to group fairness by enforcing fairness for the subgroups as well. This work aligns with our argument for a multi-objective approach to enforce multiple fairness constraints with regards to existing fairness notions.

\subsection{Defining fairness notions in terms of agent-environment interactions}
\label{section:fairnes_sequential}

Defining fairness in a sequential setting requires knowledge of how fairness notions can be defined given the agent-environment interactions.
Consider the fraud detection setting, where an agent must decide how to efficiently flag transactions for a credit card company, on a daily basis \citep{Zintgraf2017}. Throughout the day, each individual client may decide to make transactions. The agent aims to flag suspicious transactions, in a way that every continent is subject to a similar proportion of re-authentication requests.

\begin{figure}[h!]\centering
  \begin{subfigure}[h!]{0.2395\linewidth}\centering
    \includegraphics[height=7em]{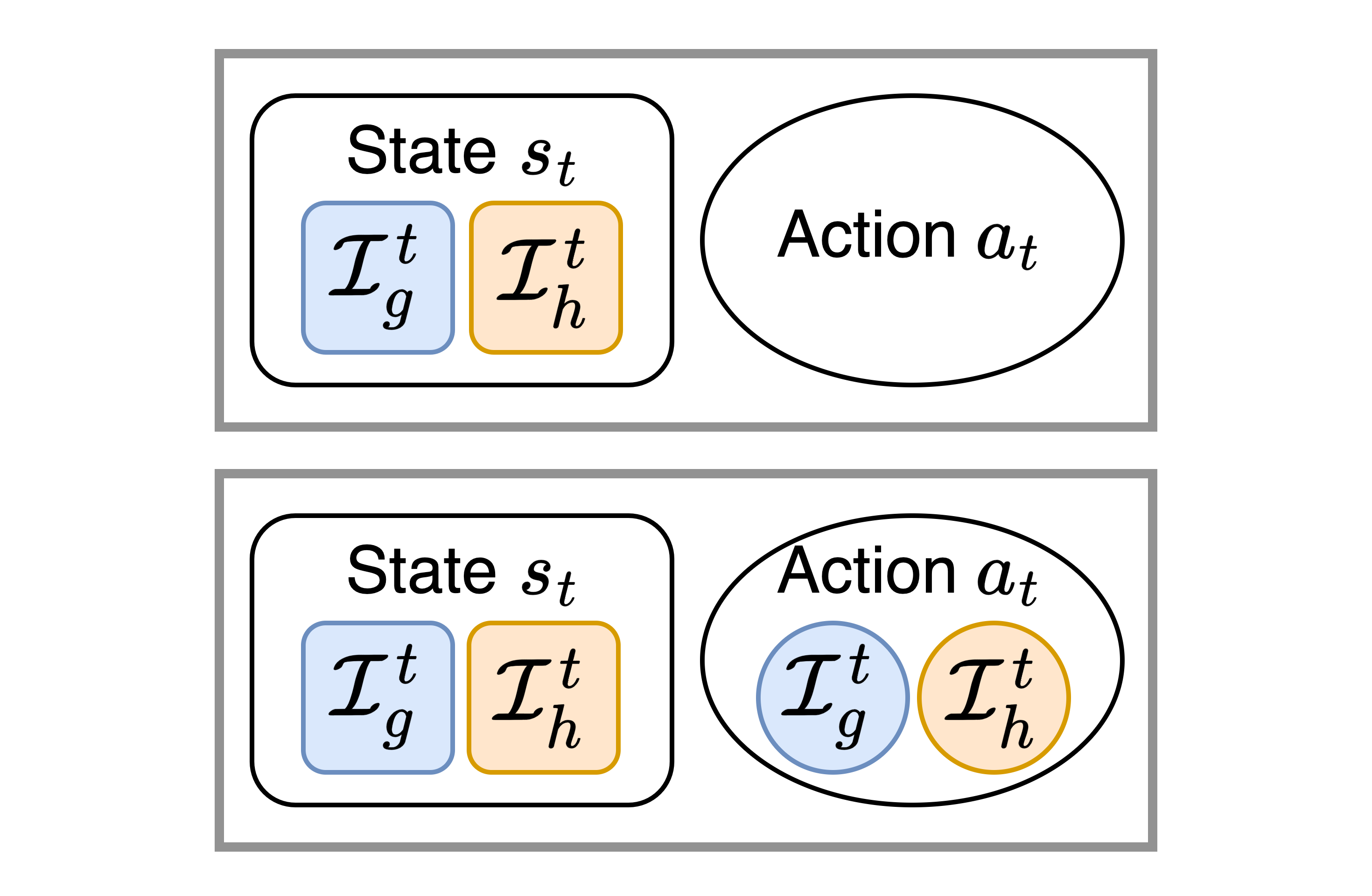} 
    \caption{}
    \label{figure:groups_t}
  \end{subfigure}
    \begin{subfigure}[h!]{0.2395\linewidth}\centering
    \includegraphics[height=7em]{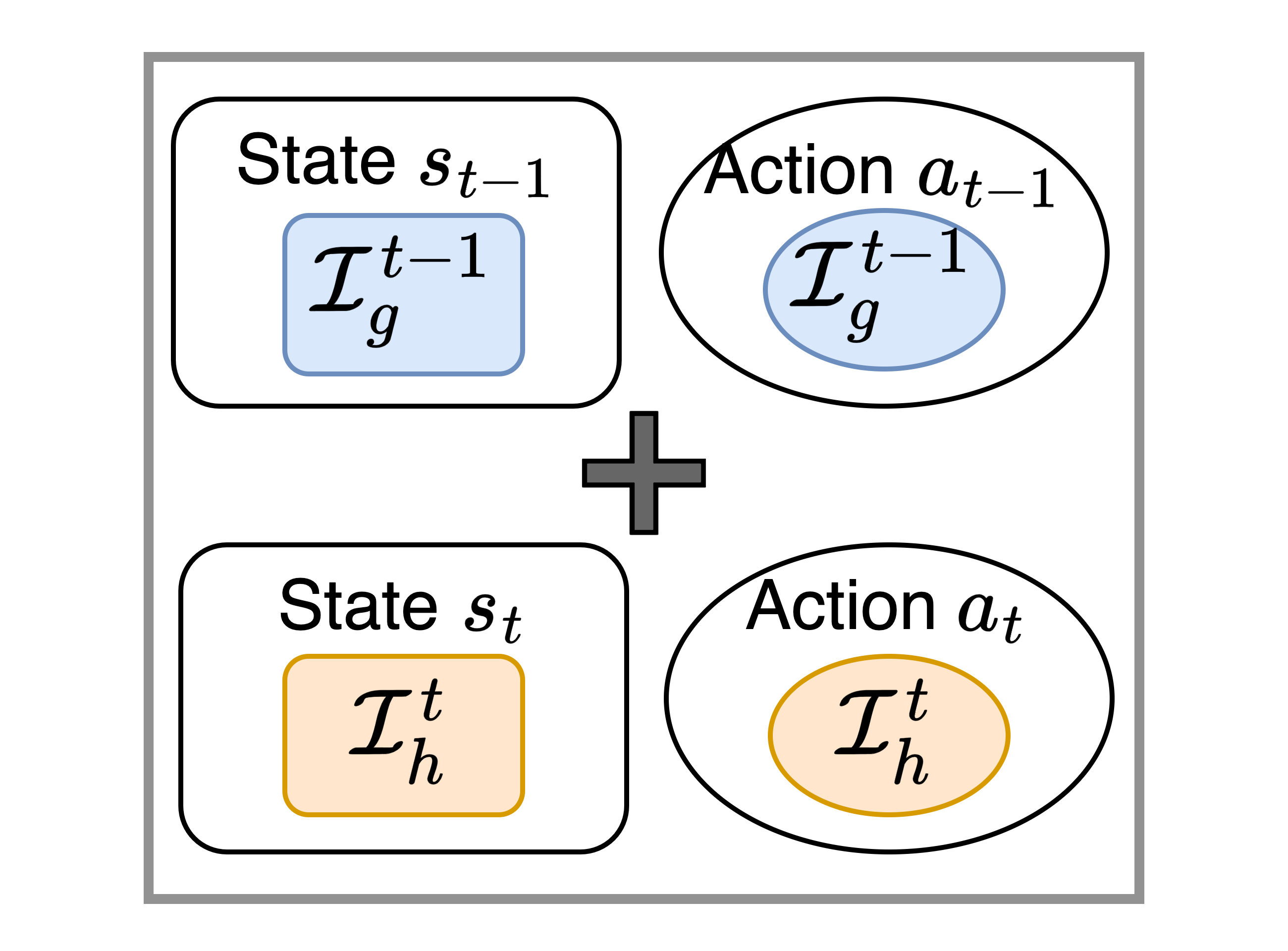} 
    \caption{}
    \label{figure:groups_horizon}
  \end{subfigure}
  \begin{subfigure}[h!]{0.2395\linewidth}\centering
    \includegraphics[height=7em]{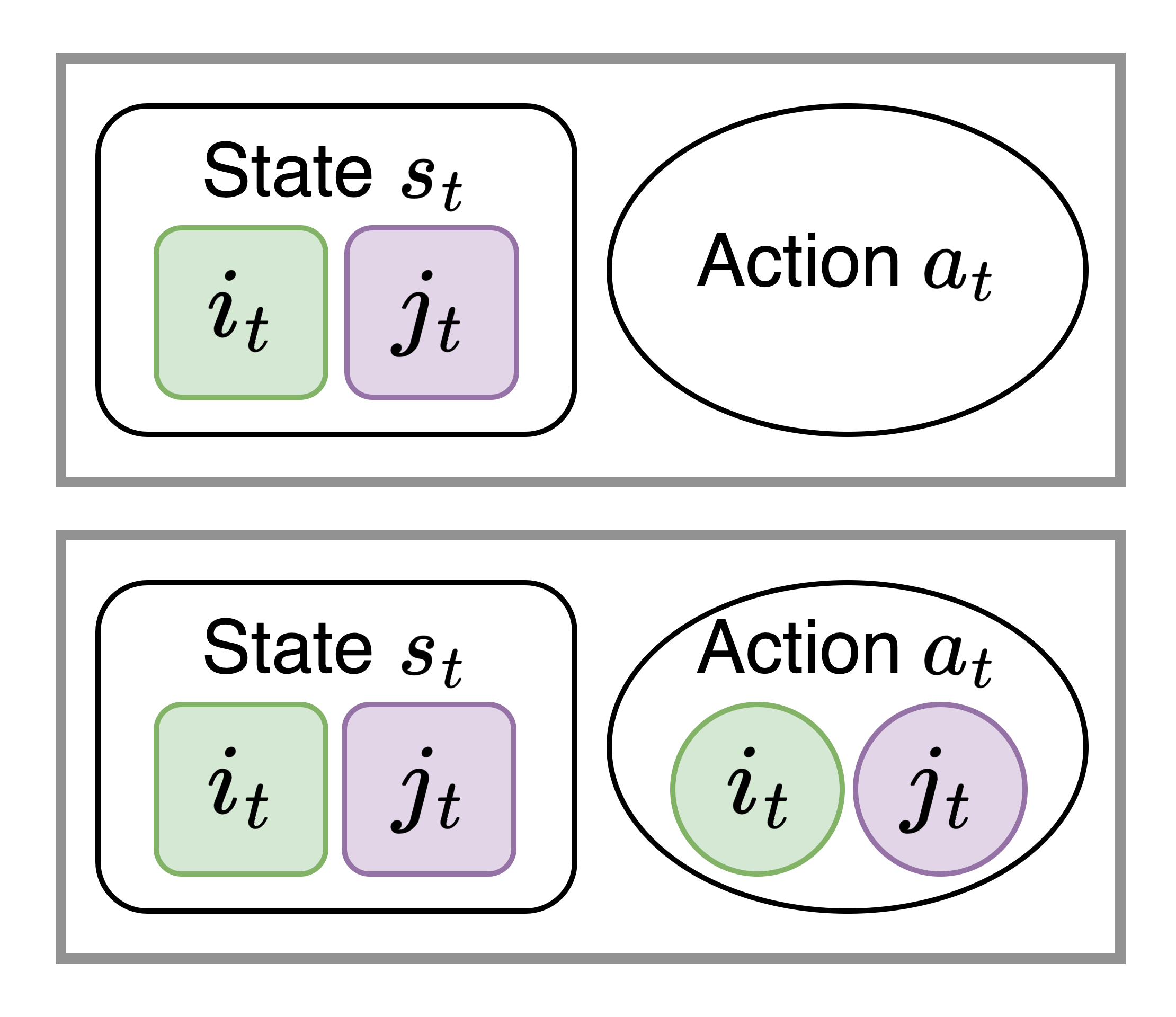} 
    \caption{}
    \label{figure:individuals_t}
  \end{subfigure}
    \begin{subfigure}[h!]{0.2395\linewidth}\centering
    \includegraphics[height=7em]{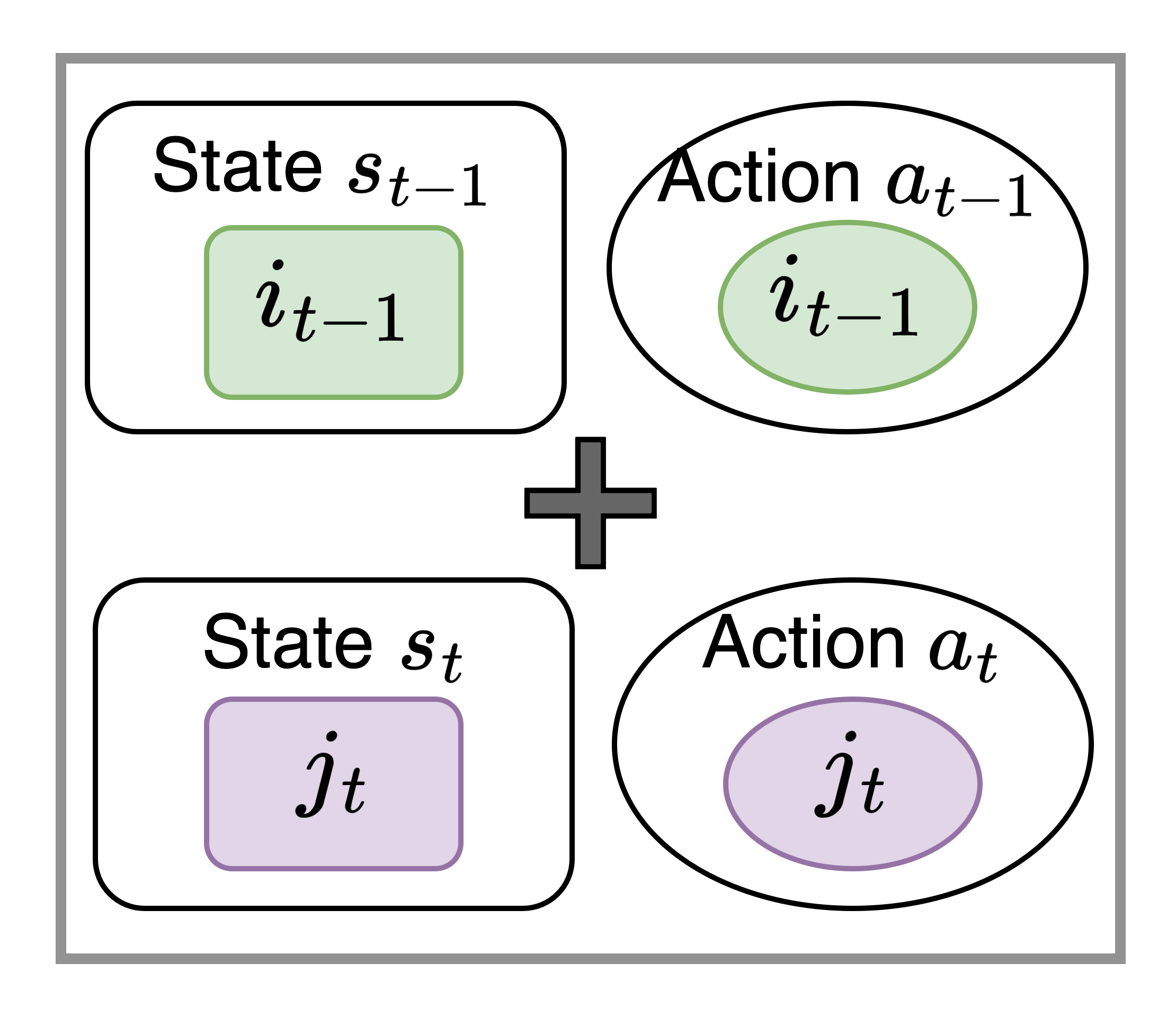} 
    \caption{}
    \label{figure:individuals_horizon}
  \end{subfigure}
  \caption{(a) (b) Scenarios where group fairness can be calculated. (a) All groups $\groupg, \grouph \in \group$ are encountered at each time $t$. Top: action $a_t$ is an action over all groups $\group$. Bottom: action $a_t$ encodes a specific action for each group $g$. (b) All groups $\group$ are encountered over a history until time $t$. The $+$ symbol indicates a union over states and actions.
  (c) (d) Scenarios where individual fairness can be calculated. (c) All individuals $\indi, \indj \in \ind$ are encountered at each time $t$. Top: action $a_t$ is an action over all individuals $\ind$. Bottom: action $a_t$ encodes a specific action for each individual. (d) All individuals $\ind$ are encountered over a history until time $t$.
  }
  \label{figure:Groups_State_Action}
\end{figure}

Suppose in our fraud detection setting, that each hour the agent encounters transactions from different continents.
Then at each time $t$, given an observed state $\rlstate_t$ and chosen action $\action_t$, a group fairness notion can be defined if $\rlstate_t$ contains all respective groups $\groupg, \grouph \in \group$ and the chosen action $\action_t$ represents the action taken towards each group.
Figure \ref{figure:groups_t} visualises the possible scenarios with regards to the available action. This may involve a common action across all groups or tailored actions specific to each group $\groupg$. Note that if individuals are defined within the state representation, then all individuals can be grouped under their respective groups.

Continuing with the fraud detection setting, consider a scenario where the agent encounters certain continents only during specific hours, as a result of time zone differences. As a result, the time horizon must be long enough to guarantee exposure to all continents.
Concretely, if the state $\rlstate_t$ contains only information on a strict subset $\subgroups \subset \group$ of the respective groups impacted by the decision at time $t$, a fairness notion can only be defined over the history $\history$ until $t$, to contain sufficient information about all impacted $\group$ groups for time $t$.
Similarly, we require multiple timesteps if the action $\action_t$ does not apply to all groups.
Figure \ref{figure:groups_horizon} visualises the scenario where only a subset of the groups is available at each time $t$, requiring a history of timesteps in order to express group fairness notions.
If individuals are defined within the state representation of the environment, the scenarios in Figures \ref{figure:groups_t} and \ref{figure:groups_horizon} can be extended to consider cases where a subset of individuals is encountered.

Following up on the same fraud detection setting, when the agent encounters all customers each hour, then individual fairness notions can be calculated for the transactions.
To define an individual fairness notion for individuals $\indi, \indj \in \ind$ at time $t$, given an observed state $\rlstate_t$ and a chosen action $\action_t$, we require that the state and action information concerning all individuals is defined.
Figure \ref{figure:individuals_t} visualises the scenarios where individual fairness can be calculated at each time $t$.
Note that the action can be fine-grained, targeted to each individual, or coarse-grained, targeted towards their respective countries or continents.

If state $\rlstate_t$ does not contain all $\ind_t$ individuals but rather a strict subset $\subinds_t \subset \ind_t$, an individual fairness notion can be defined over a history $\history$ until $t$ when subsets of the individuals are encountered.
For instance, in fraud detection, one might periodically sample transactions from subsets of customers within specific continents throughout the day, allowing monitoring for suspicious activity patterns based on customers' local timezones. In this scenario, a fair agent should balance how frequently each continent is monitored over time, ensuring that customers from certain regions are not disproportionately subjected to repeated authentication requests.
Figure \ref{figure:individuals_horizon} visualises the scenario where individual fairness can be expressed over a sequence of time steps.

Note how both group and individual fairness notions can be expressed if the encountered states contain all necessary information about the respective groups and individuals. Regardless of whether the action was specifically assigned to them, their group, or the entire population, we can compare the action which affects them to calculate fairness notions.
In this work, we consider the scenarios from Figures \ref{figure:groups_horizon} and \ref{figure:individuals_horizon}, where all groups and individuals are encountered over a history, respectively.

We consider each fairness notion by computing its approximation $\approximatefairness$, through the history by considering a sliding window of the most recent interactions. Note that this approximation is necessary because exact fairness across the entire transaction history is computationally intractable under typical fairness definitions. On the one hand, we require enough interactions to guarantee exact fairness \citep{Jabbari2017}. On the other hand, considering the full history makes computing the fairness notions intractable. Individual fairness notions in particular become intractable due to the pairwise comparisons needed between each new individual and all those previously encountered.
In the context of data mining, approaches focus on over-representing minorities or rare events \citep{Nargesian2021} in the training data. Similarly, recommender systems suffer from uneven data distributions which impacts fairness and as such requires re-distributing the data to appropriately compare groups and individuals \citep{Wang2022}. We consider such approaches as future work to learn better approximations for fairness notions over a full history.


In this work, we assume that the states in the history encompass all groups and individuals necessary to compute the relevant fairness notions. However, to meet this assumption, the relevant states need to be encountered, which is highly dependent on how the agent interacts with the environment. To establish this, we need an appropriate exploration strategy that ensures that sufficient information is collected about all groups and individuals. On the one hand, to guarantee optimality, this exploration strategy will need to collect information on groups and individuals as broadly as possible. On the other hand, to keep the process computationally tractable, the exploration strategy should be effective and targeted. 
To support decision makers, policies can be learned in simulated environments or directly in the real-world. This choice depends on the problem at hand, and particularly how the agent's actions would impact the groups and individuals.
Furthermore, the availability of a simulated environment may provide insights on which performance-fairness trade-offs are possible prior to deploying them in the real world. 
This facilitates a model-based reinforcement learning loop that could mitigate the hurdle of computationally intensive exploration strategies.

\subsection{Different fairness perspectives}
Most fairness notions are expressed in terms of the action \citep{Makhlouf2020, Dwork2018}.
Depending on the setting, it might be more appropriate  to evaluate fairness based on the impact of the agent's actions, rather than merely assessing the fairness of the actions themselves.
So far, we have discussed all fairness notions above from the perspective of actions. However, both the immediate and estimated effect follow similar rules as information about them must also be available in the agent-environment interactions.
An example in the context of fraud detection, where the (future) impact of the action is considered more important, is when the agent should avoid burdening customers from different continents equally to not lower customer satisfaction.

The job and fraud detection fairness examples we described so far define fairness notions based on the the action directly impacting the individuals or groups, such as hiring/rejecting a job applicant or ignoring/flagging a transaction. However, some settings may define fairness in terms of the benefit of the company employing the RL agent. For instance, the weakly meritocratic fairness notion requires that one group of people is not preferred over another when hiring for a job, unless the estimated future reward is higher \citep{Jabbari2017, Joseph2016}. However, when this reward is defined for the company doing the hiring process, the definition of fairness does not necessarily match being fair from the perspective of the person being impacted.
Depending on how aligned the agent's rewards and actions are with fairness from the perspective of the impacted people, fairness definitions such as weakly meritocratic may be unsuited. An other approach in the context of resource allocation uses social welfare functions \citep{Weng2019, Fan2023} to define both fairness and efficiency in a single scalar output. This approach requires that the rewards from the MDP must be efficiently and fairly divided over multiple individuals or groups. However, if the MDP is not defined for this specific scenario, a more general approach of optimising the performance and fairness separately is warranted.

\section{Scenarios}
\label{section:scenarios}

Automated decision-making in real world applications has many advantages. When dealing with a large number of individuals, using algorithms to quickly filter through data saves cost by removing the human from this initial screening process. Moreover, algorithms can handle a larger number of features than humans, enabling them to perform fine-grained comparisons between instances to better optimize the desired objectives.
In this section, we introduce a job hiring and fraud detection scenario, that we use in our simulation experiments, along with their distinct fairness implications.

\subsection{Job hiring}
\label{section:jobhiring}

Job hiring is a reoccurring process throughout a company's lifetime. This allows companies to use previous data when training algorithms. However, the training data may be subject to historical bias, which is then further exacerbated by the algorithm \citep{Makhlouf2020}. 
Additionally, the job hiring process is sequential, typically consisting of multiple decision stages, i.e., resume screenings and possibly multiple rounds of interviews \citep{Bogen2018}, which warrants a sequential approach. Moreover, unfairness at one stage may be propagated to consecutive stages. 
In job hiring, gender-based discrimination ranges from stereotypes and employer beliefs \citep{VanBorm2022,Barron2022} to occupational-specific characteristics \citep{Holdman2018,Kubler2018,Ahmed2021,Cortina2021,Adamovic2023}.
Age-based discrimination depends on supply and demand \citep{Neumark2021}, and impacts job opportunities which decline with age \citep{Busch2011}. 
Ethnic discrimination has been studied from an immigration perspective \citep{Zschirnt2016} and is based on implicit interethnic attitudes \citep{Blommaert2012}. 
In addition, discrimination is known to arise when multiple sensitive features are considered jointly, as certain combinations of these attributes can lead to unfair treatment or outcomes \citep{Petit2007,Baert2018,Derous2019}.

Structured interviews are shown to be more resilient against discrimination when it comes to human assessment \citep{Kith2022}. However, there are no guarantees automated decision-making techniques follow this trend as removing sensitive features from the data does not mitigate fairness concerns. An explanation for this phenomenon is the use of proxy features by the algorithm to re-introduce bias \citep{Kilbertus2017}. Therefore, we require an algorithm that is inherently fair, without the need to pre-process the data. As the full dataset is typically not available prior to training in an RL setting, it makes pre-processing techniques challenging to attempt. To this end, we evaluate the performance of algorithms with regards to the main objective (i.e., hiring qualified candidates) and appropriate fairness notions.

\paragraph{Job hiring $f$MDP}
\label{section:jobhiring_mdp}

We define the job hiring setting as an $\fmdpe$ (introduced in Section~\ref{section:fairness_history}), where an agent must learn to build a well-performing team of employees, when presented applicants sampled from the Belgian population \citep{STATBELBelgianPop}. 
Given an applicant and the current team composition, the agent must decide on the appropriate action $\action_t$, i.e., to hire or reject the applicant, based on their estimated qualifications. 
To calculate the qualification of each applicant, we define an objective but noisy goodness score $\goodness \in [-1, 1]$, that estimates how much the applicant will improve the company based on their skills. Concretely, a positive goodness score indicates that a candidate has desired qualities that makes them suitable for hire.
Using a threshold $\epsilon = 0.5$, the ground truth action $\trueaction_t$ states to hire the applicant if $\goodness_t >= \epsilon$, otherwise reject. We define this goodness score as the ground truth for our experiments based on which the $\fmdpe$ classifies applicants. 
Based on the company's preferences, the goodness score is a weighted sum, allowing the skills or diversity to be prioritised.
To extend the MDP to an $\fmdpe$, we implement the feedback signal $\feedback_t$ as the correct action $\trueaction_t$ based on the goodness score:
\begin{eqnarray}
    \feedback_t = \trueaction_t
\end{eqnarray}
We provide additional details in \supplementary{\ref{suppl:jobhiring_mdp}} on the job hiring $\fmdpe$ and the applicant generation.

\paragraph{Fairness notions in job hiring}
In this work, we consider fairness concerns in job hiring based on discrimination grounded in two sensitive features: gender and nationality. 
We cover distinct job hiring scenarios to highlight various fairness implications when it comes to the agent-environment interactions. 
As the agent observes an applicant in the state $\rlstate_t$ at each timestep $t$, both individual and group fairness notions are applicable (Section~\ref{section:fairness_notions}). 
We consider the context of unfairness based on gender, where an applicant $\indi_t \in \ind_t$ can belong to the group of $\ggmen$ or $\ggwomen$.

\subsection{Fraud detection}

Fraudulent credit card transactions result in significant losses when undetected \citep{Delamaire2009}. Although manual investigations can accurately detect fraud, they are impractical for large volumes of transactions, as they inevitably result in significant delays. Moreover, fraudsters frequently alter their behaviour over time to evade detection \citep{Dal2014}, necessitating an online approach capable of continuously adapting to emerging fraudulent strategies. Since customers typically perform multiple transactions within a given time period, credit card companies must carefully balance fraud-prevention authentication steps against customer satisfaction and tolerance \citep{Zintgraf2017}.
As transactions typically include personal and location data, algorithms may learn to discriminate based on sensitive features. For example, customers from countries with a higher proportion of fraudulent transactions may face more frequent checks solely based on their geographical location \citep{Makhlouf2020}. Therefore, fraud detection methods require fairness criteria that explicitly account for these differences in fraud base rates to accurately identify suspicious transactions.

\paragraph{Fraud detection $f$MDP}
\label{section:frauddetection_mdp}

The fraud detection setting concerns online credit card transactions where multi-modal authentication is used to identify and reject fraudulent transactions. To simulate customer behaviour, we use the MultiMAuS simulator \citep{Zintgraf2017}, which is based on a database of real-world credit card transactions.
We extend this simulator to a $\fmdpe$, by encoding information regarding each transaction into a state. Furthermore, we provide the current company's fraudulent transactions percentage and customer satisfaction along with the transaction in the state at each time step. The feedback signal $\feedback$ is defined based on the gain or loss in reward, indicating if revenue was affected by fraudulent activity. Concretely, the agent receives a $+1$ for every successful genuine transaction and $-1$ for uncaught fraudulent or cancelled transactions.
The reward $\reward_t$ at timestep $t$ specifies the correctness of the action if the agent requests authentication. Consequently, if the reward is positive the transaction is considered genuine, while a negative reward indicates an unsuccessful transaction, caused by a loss in commission or by stolen money requiring the credit company to repay the losses to the client.
To implement a feedback signal $\feedback$, we infer the correctness when authenticating to observe the amount of true positives and false positives.
We provide additional details on the MultiMAuS simulator and the $\fmdpe$ in \supplementary{\ref{suppl:frauddetection_mdp}}.

\paragraph{Fairness notions in fraud detection}
We investigate fairness in fraud detection based on the continent of the customers.
As the agent observes a new transaction in state $\rlstate_t$ at timestep $t$, both individual and group fairness notions are applicable. 
For simplicity, we consider two continents, $\cA$ and $\cB$, with the most transactions. We define group fairness notions as follows: 
Given transactions $\indi_t \in \ind_t$, where transaction $\indi_t$ can belong to continent $\cA$ or $\cB$, all group fairness notions require that the difference in treatment between the groups $\gcontA$ and $\gcontB$ is minimised.

\section{Guidelines}
\label{section:guidelines}

Applying the fairness framework to an existing \mdp requires some assumptions to be made to extend it into an \fmdps.
First, \textbf{interactions from the \mdp must be made accessible for the fairness framework}. Moreover, each interaction must specify which groups and/or individuals are impacted by it to express any fairness notion. 
For example, in our fraud detection setting, the state includes information about the continent associated with the transaction in question. This allows the agent’s chosen action to be evaluated using fairness metrics that take the transaction’s continent into account.
Second, \textbf{information concerning the computation of the requested fairness notions must be available}. As some fairness notions mentioned in previous sections, such as Equal Opportunity, require that the correct action is known, this information must be available through the \fmdps. 

\paragraph{Selecting an appropriate history}
As the fairness framework uses the history to compute all fairness notions, the maximum history size must be decided a priori, as it defines the time scale on which to compute fairness.
Using an overly large history size may lead to policies that put less emphasis on recent interactions. Concretely, new actions are more often chosen to be suitable for fairness, given the large number of previous interactions. If fairness notions are computed over a larger history, they provide no fairness guarantees over a subset of this history. Inversely, using a small history does not guarantee fairness over a larger number of interactions.
Note that choosing a larger history size requires more storage and more computation.

In our experiments, we provide results for a selection of history sizes, which are implemented as a sliding window of the most recent interactions. While for simplicity we have chosen a single history\footnote{There are currently two histories shared across all fairness notions: a long history of the user-specified size, and a history for the current timestep only. This matches our definitions in Section~\ref{section:fairnes_sequential} for defining fairness over a history or for the current timestep only.} that is to be shared across all computed fairness notions, our fairness framework can be easily extended to consider a different history size per fairness notion.
Furthermore, we explore using a dynamic history size to facilitate finding an appropriate history size by detecting possible concept drifts that impact fairness notions. Given the granularity of individual fairness, which involves comparing all individuals in the history to one another, we use it as the guiding fairness notion for determining an appropriate history size. The discounted history is defined using four parameters: a minimum window size $w$, a discount factor $\gamma \in ]0, 1]$, a discount threshold and a discount delay. The minimum window size allows to specify a minimum number of interactions to consider for the fairness computation. The discount factor allows to assign a decreasing weight to earlier interactions such that recent interactions contribute more to the fairness computations. While in a sliding window history, each interaction has the same weight for the fairness computation (equivalent to a discount factor $\gamma=1$), the discounted history employs a discounted sum, such that each interaction contributes $\gamma^{w-t}$, with $w$ the minimum window size and $t$ the respective timestep of the interaction. 
The discount threshold defines the minimum change in the fairness notion required for an earlier interaction to be considered relevant. The discount delay indicates how many subsequent interactions can occur before all earlier interactions are considered irrelevant.
If the fairness computation does not change more than the specified threshold for the given delay number of consecutive interactions, then all earlier interactions are disregarded to ease computational and memory resources.

\paragraph{Choosing fairness notions}
As all objectives to optimise must be known a priori for most RL algorithms, the fairness notions that should be optimised must be specified before training. Note that the fairness framework allows other fairness notions to be computed after training has started, as long as the necessary information for these fairness notions is available, in real-time or gathered with a delay. For example, computing a fairness notions with a smaller history size or computing additional fairness notions for which the correct action is already known is possible. 
However, if the chosen RL algorithm is unable to change objectives during training, these new fairness notions cannot be optimised if added after the start of the training process. 
We emphasise that even with this limitation, computing additional fairness notions for a retrospective analysis can provide insights on the overall fairness of an algorithm. Consequently, we will present our results in the following sections with additional measurements to highlight the benefit of these insights. We refer to \cite{Makhlouf2020} for an overview on how to appropriately select most fairness notions based on given selection criteria.

\section{Experiments}
\label{section:experiments}

First, we showcase a collection of baseline scenarios for the job hiring and fraud detection setting. Unless specified otherwise, all experiments in this section and the following sections are based on 10 seeds (i.e., runs), for a total of 500.000 time steps, where any individual fairness notions use the HEOM distance metric (Equation~\ref{equation:HEOM}). For all experiments, we report the learned non-dominated coverage sets for all objectives \citep{Hayes2022}.
The reward vector consists of the following objectives: the performance reward (R), statistical parity (SP), equal opportunity (EO), overall accuracy equality (OAE), predictive parity (PP), predictive equality (PE), individual fairness (IF) and consistency score complement (CSC). We refer to Section~\ref{section:fairness_notions} for the definition of these fairness notions.
When visualised in radar plots, all objectives are normalised, such that 0 is the maximum cumulative return for each of the objectives. Consequently, the closer a fairness notion is to 0, the more fairly the agent behaves with respect to that particular fairness notion. 
To establish an empirical upper bound for the performance reward in both scenarios, we consider a single-objective reinforcement learning agent that only optimises the performance reward. We then normalise the radar plots in this section based on the learned maxima. To this end, we train a Deep Q-Network (DQN) algorithm \citep{Mnih2013,Mnih2015} and aggregate the results across 10 runs. DQN is a deep neural network implementation of the standard $Q$-learning algorithm \citep{Watkins1989,Watkins1992,Sutton2018} and uses an experience replay buffer to avoid problems created by training on correlated data and non-stationary data distributions \citep{Mnih2013}.
We use an $\varepsilon$-greedy action selection policy, with $\varepsilon=0.1$, meaning the agent will act greedily (i.e., select the action with the highest estimated future reward) 90\% of the time and act randomly 10\% of the time. The neural network consists of 1 hidden layer of 64 neurons and a RELU activation function \citep{Goodfellow2016}.
Based on the results across 10 runs, a DQN agent reaches a maximum reward of 46.53243 and 906.0 for the job hiring and fraud detection setting, respectively. Therefore, we normalise the performance reward results based on these maxima. 
We present the results visually in the following sections and refer to \supplementary{\ref{appendix:job_results}} and \supplementary{\ref{appendix:fraud_results}} for a concrete overview of the numeric results for the learned job hiring policies and fraud detection policies, respectively.

As both scenarios deal with a performance reward and multiple fairness objectives, the number of policies with suitable trade-offs can scale exponentially. To learn all policies would therefore be computationally intractable, to explore the entire state space. To this end, we use Pareto Conditioned Networks (PCN) \citep{Reymond2022pcn}. 
PCN trains a single neural network to approximate all non-dominated policies, by applying supervised learning techniques to improve the policy.
A Pareto Conditioned Network (PCN) \citep{Reymond2022pcn} applies supervised learning techniques to approximate all non-dominated policies within a single neural network.
PCN takes as input a tuple $\langle \rlstate, \desiredHorizon, \desiredReturn \rangle$, representing the observed state $\rlstate$, the desired return $\desiredReturn$ to reach at the end of the episode and the desired horizon $\desiredHorizon$ indicating the number of timesteps that should be executed before reaching $\desiredReturn$.
Both $\desiredHorizon$ and $\desiredReturn$ are chosen by the decision maker at the start of an episode. Consequently, at every timestep $t$, the desired return is updated by the received reward $\desiredReturn \leftarrow \desiredReturn - \reward_t$ and the desired horizon is decreased by one timestep $\desiredHorizon \leftarrow \desiredHorizon - 1$. PCN learns policies similar to classification techniques, where $\langle \rlstate_t, \Horizon_t, \Return_t \rangle$ is the input at timestep $t$ and the chosen action $\action_t$ is the output.
We employ a dense neural network with state, horizon and return embeddings, with each consisting of a hidden layer of 64 neurons and a sigmoid activation function. Their outputs are fed through a fully connected neural network of 2 layers with a RELU activation on the first layer. This last network produces outputs for each action.

\paragraph{Job hiring}

For the job hiring scenario, we train an agent to hire and maintain a well-performing team of 100 employees, where each episode lasts for a maximum of 1000 timesteps.
We consider the Belgian population, informed by the official statistics registry of Belgium, STATBEL \citep{STATBELBelgianPop}, where the agent is required to ensure fairness between men and women in the hiring process.

\paragraph{Fraud detection}

For the fraud detection scenario, we use the default parameters of the MultiMAuS simulator \citep{Zintgraf2017}, but increase the frequency of fraudulent transactions to ensure that both genuine and fraudulent transactions are sufficiently represented for the anonymised continents $\cA$ and $\cB$. This results in approximately 10\% fraudulent transactions and allows us to gain compute all previously mentioned fairness notions. Note that continents $\cA$ and $\cB$ have different base rates of fraudulent transactions. We let the agent check transactions for a week, resulting in at most 1000 transactions per episode, where the agent must be fair towards requesting re-authentications from both continents.

\subsection{Single-objective scenarios}

\begin{figure}[h!]
    \centering
    \includegraphics[width=1\linewidth]{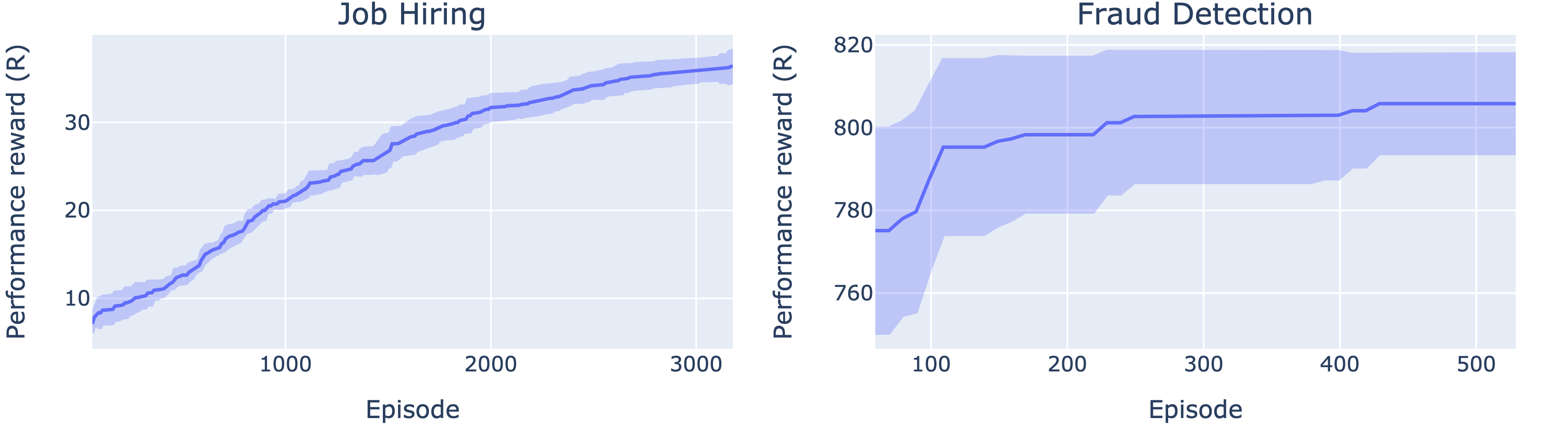}
    \caption{Mean and standard deviation of the obtained performance reward of PCN across 10 runs of 500 000 timesteps, for the job hiring and fraud detection setting. For job hiring, the PCN agent shows a continuous improvement over time. In contrast, while the initial performance in the fraud detection setting is high, the continuous improvement during training is much slower.}
    \label{fig:job_fraud_single}
\end{figure}

When optimising for the performance reward (Figure~\ref{fig:job_fraud_single}), the agent learns continuously improving policies. The achieved performance reward for job hiring is more consistent across the runs, as opposed to the performance for fraud detection. We attribute this to the distinct problem definitions of both settings. In the job hiring setting, the agent encounters job applicants where their features and the preferred action (i.e., hire or reject) are strongly correlated. Concretely, if two job applicants share the same features, it is more likely that the same action is preferred. In contrast, in the fraud detection setting the fraudulent transactions represent a minority of all transactions, which makes them more challenging to detect. Consequently, the agent must learn to flag this small subset of transactions, while ignoring the majority of genuine transactions. As these fraudulent transactions are less represented during simulations, the agent learns to detect these slowly, resulting in a slower optimisation process.

\begin{figure}[h!]
    \centering
    \includegraphics[width=1\linewidth]{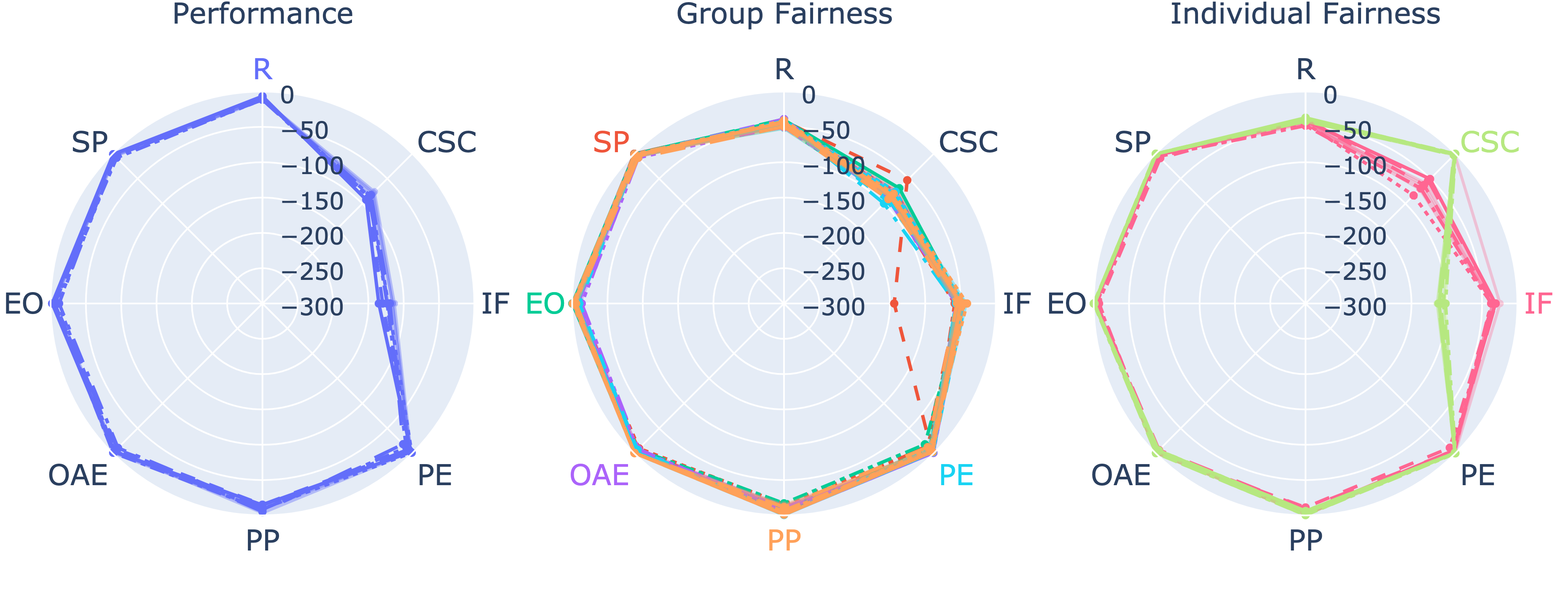}
    \caption{Representative set of learned job hiring policies when optimising a single objective, split per type of objective.
   Left: Optimising for the performance reward (R) only. Center: Optimising a group fairness notion (SP, EO, OAE, PP or PE). Right: Optimising an individual fairness notion (IF or CSC). 
   Different fairness notions are indicated by different colours. Lines in the same colour represent outcomes of different runs.
   Policies that optimise the performance reward achieve similar outcomes across the fairness notions. Most notably, both individual fairness notions are lower for the performance policies compared to policies that focus on a fairness notion. In contrast, policies that do well for fairness notions (center and right plots) achieve a lower performance reward.}
    \label{fig:job_single}
\end{figure}

While the performance reward specifies how efficient a policy is, it provides no insights into the impact of the policy with regards to other objectives.
To this end, to highlight the importance of using a multi-objective approach, we first present the results of single-objective policy runs, plotted against all other objectives. 
Figure \ref{fig:job_single} shows the results for the job hiring setting.
When asked to optimise the performance reward (R), the agent obtains a reward close to the maximum (0).
Note how the learned policies are close to 0 for all group fairness notions. This indicates that the agent has learned policies which can satisfy multiple fairness notions, despite none being requested. However, all policies achieve lower individual fairness.
When the agent is required to optimise one of the group fairness notions, other group fairness notions also receive a high value. As most group fairness notions are based on similar statistics, there are overlaps regarding their outcome. 
In other words, within this single-objective setting, optimizing one fairness notion often leads to improvements in other fairness notions as a side effect, due to their partial alignment. Note that optimising for statistical parity (SP) results in a policy that underperforms on individual fairness (IF). This indicates that, for the job hiring setting, these two fairness notions are less aligned.
In contrast, individual fairness notions make pair-wise comparisons of similar individuals. As such, it is possible for the agent to find larger differences in the non-dominated values, as IF considers the probability distributions over the actions, while CSC only considers the action. Note how CSC-policies obtain a low reward for IF and vice versa. This may indicate that these two fairness notions may be possibly conflicting in this setting.
Optimising for any fairness notion results in a lower performance reward. However, the agent is able to learn policies that provide high group fairness. 
Nevertheless, we observe lower individual fairness when optimising for the performance reward or some group fairness notions in particular. 
While the agent learns fair policies regarding gender, it does so without considering the combination of other sensitive attributes, such as nationality. This highlights the need for individual fairness notions to detect feature intersectionality \citep{Makhlouf2020}.

\begin{figure}[h!]
    \centering
    \includegraphics[width=1\linewidth]{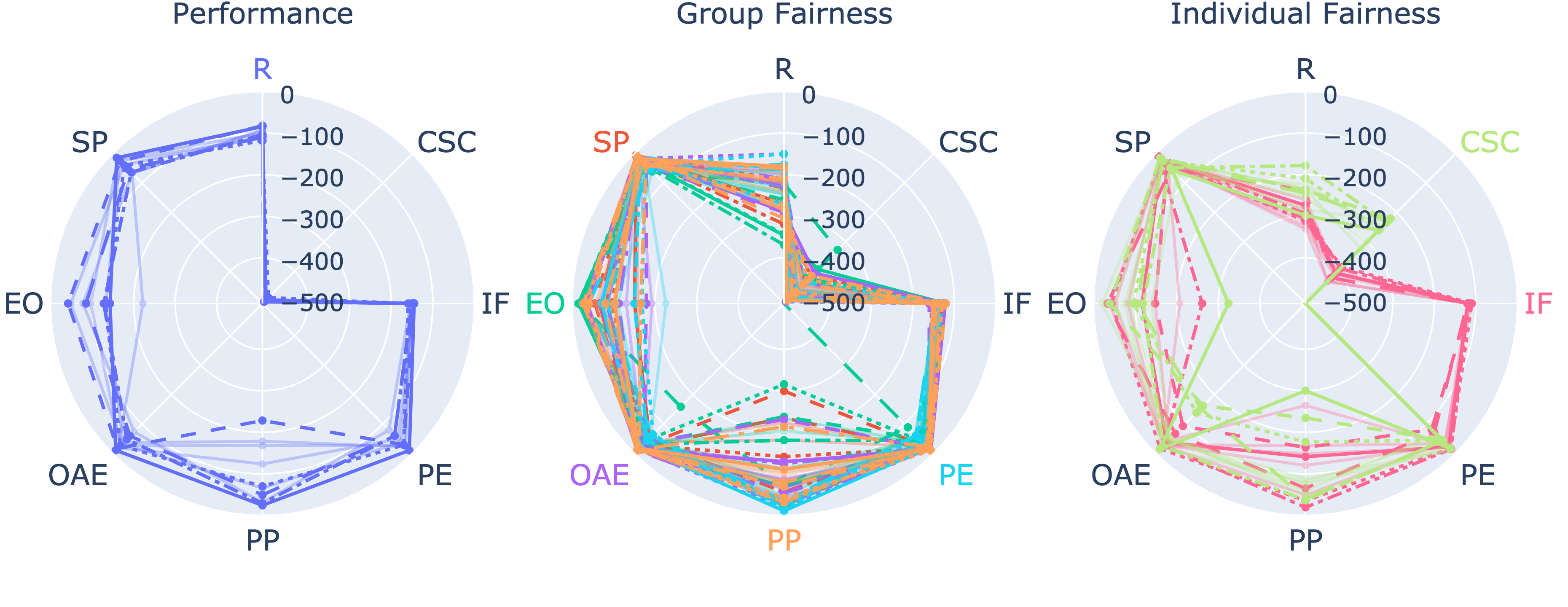}
    \caption{Representative set of learned fraud detection policies when optimising a single objective, split per type of objective.
   Left: Optimising for the performance reward only. Center: Optimising a group fairness notion. Right: Optimising an individual fairness notion. 
   Different fairness notions are indicated by different colours. Lines in the same colour represent outcomes of different runs. 
   When optimising for the performance reward, small variations in the performance reward lead to high variations with respect to EO and PP. This indicates that interesting trade-offs can be found in this use case. 
   Overall, we observe that the agent is unable to find policies which maximise CSC, regardless of the objective. We attribute this to the difficulty of the environment, when it comes to treating similar transactions similarly. In contrast to the job hiring setting, policies optimising for a group fairness notion appear less aligned.}
    \label{fig:fraud_single}
\end{figure}

Figure \ref{fig:fraud_single} shows the learned single-objective policies for the fraud detection setting. When optimising for the performance reward (R), learned policies achieve high fairness outcomes across most fairness notions. Most notably, policies achieve diverging outcomes on the group fairness notions EO and PP. This suggests that the performance reward and these two fairness notions are not well aligned, as achieving a higher reward does not imply better fairness outcomes.
When optimising a group fairness notion, we observe high variability across the objectives. While optimising for the performance reward (R) results in a low CSC outcome, optimising for a group fairness notion improves CSC in some cases. In contrast, optimising for an individual fairness notion results in a lower performance reward, indicating these objectives are conflicting. Moreover, note that policies optimised for IF underperform for CSC and vice versa. This highlights that transactions, in this setting, are less correlated regarding their type (i.e., genuine or fraudulent). We hypothesise that this may be caused by the sensitive features of the transaction, including the continent where the transaction originates from.

\begin{figure}[h!]
    \centering
    \includegraphics[width=0.45\linewidth]{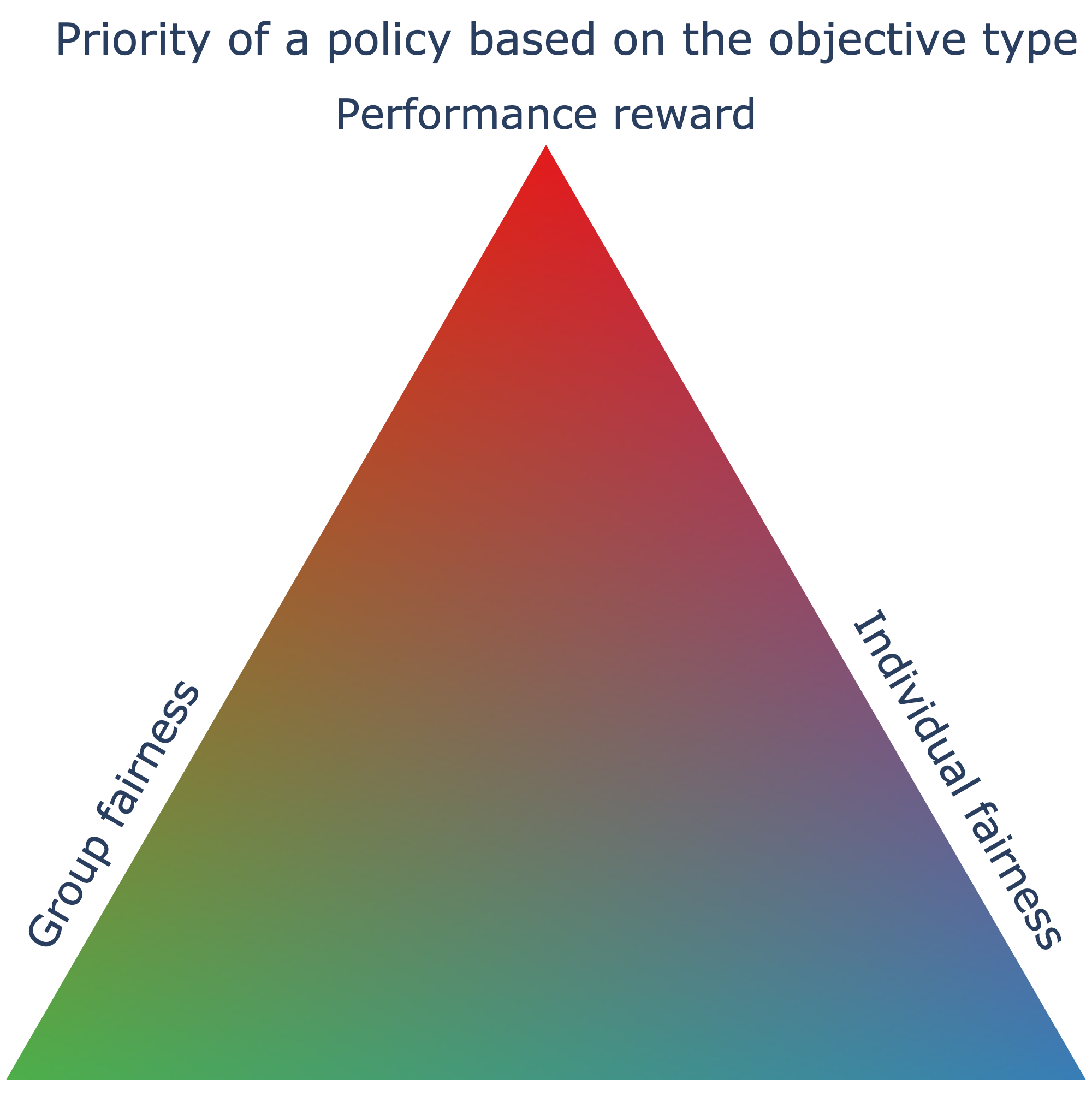}
    \caption{Colour scheme used to highlight how objective types are prioritised by a policy. Within an scenario, a policy with a high performance reward and low group and individual fairness will have a bright red hue. In contrast, a policy with high group and individual fairness but a low performance reward will have a blue-green hue.}
    \label{fig:colour_legend}
\end{figure}

\subsection{Multi-objective scenarios}

As the number of trade-off policies learned is quite high in the multi-objective scenarios, we select a representative subset of policies which provide a good approximation of the Pareto front in the results below. For completeness, we plot the representative policies in colour and all remaining policies in transparent grey. We provide additional details on the visualisation in \supplementary{\ref{suppl:policy_visualisation}}.
Figure~\ref{fig:colour_legend} illustrates the weighted colour scheme used for the policy trade-offs, based on how the different objective types (i.e., performance reward, group fairness and individual fairness) are prioritised.

\begin{figure}[h!]
    \centering
    \includegraphics[width=1\linewidth]{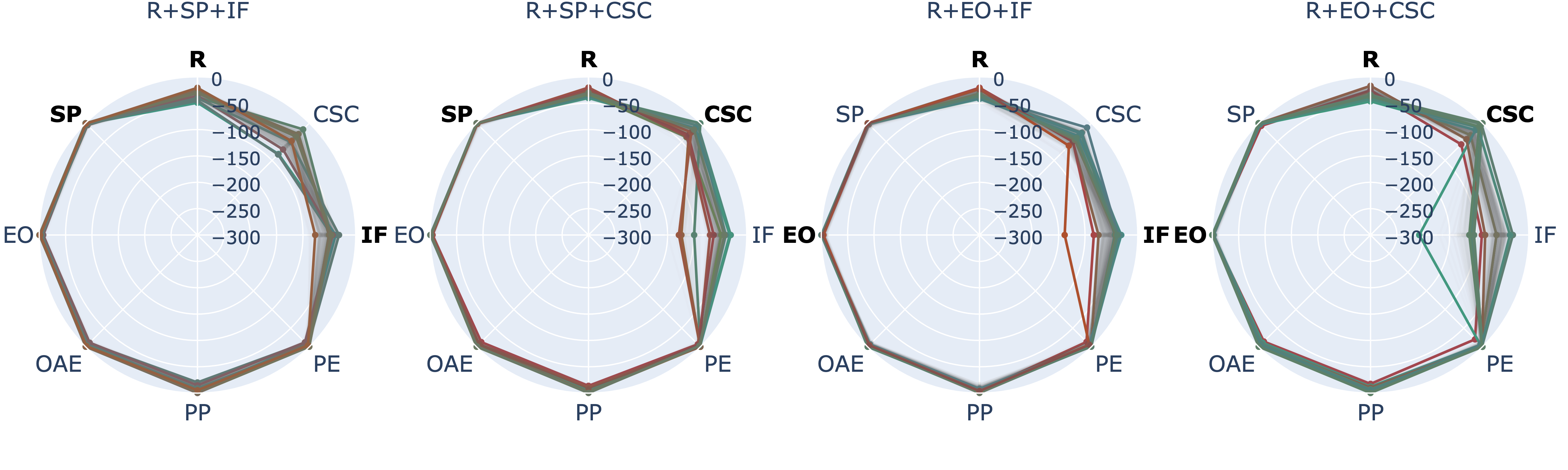}
    \caption{Representative set of learned job hiring policies when optimising the performance reward (R), a group fairness notion (SP or EO) and an individual fairness notion (IF or CSC).
    On the one hand, the requested group fairness notions has a limited effect on the outcome of the other group fairness notions. This indicates that high overall fairness is achievable in the job hiring setting, without explicitly optimising for these additional objectives. On the other hand, requesting one individual fairness notion impacts the policy's outcome regarding the other. Moreover, policies which obtain a high CSC underperform on IF and vice versa, denoting these notions a weaker alignment compared to the group fairness notions.}
    \label{fig:job_multi}
\end{figure}

To expand upon the single objective results, we ask a PCN agent to learn multi-objective policies. To improve the interpretability of the results and highlight the versatility of our framework, we request multi-objective policies that always consider the performance reward, one group fairness notion and one individual fairness notion simultaneously.
Figure \ref{fig:job_multi} shows the learned multi-objective policies for the job hiring setting, where we evaluate 4 distinct combinations of these objectives. Overall, the agent achieves high group fairness outcomes with a minor trade-off in performance reward, compared to a policy optimising only the performance reward (Figure~\ref{fig:job_single}). In general, all group fairness notions—including those not explicitly targeted—tend to be naturally satisfied by the agent, as they are well-aligned and require little additional optimisation effort. We observe the largest differences in the learned trade-offs regarding the performance reward and either individual fairness notion. Note that the requested individual fairness notion impacts which trade-offs can be found, indicating the combination of requested fairness notions influences the overall fairness a policy can provide.

\begin{figure}[h!]
    \centering
    \includegraphics[width=1\linewidth]{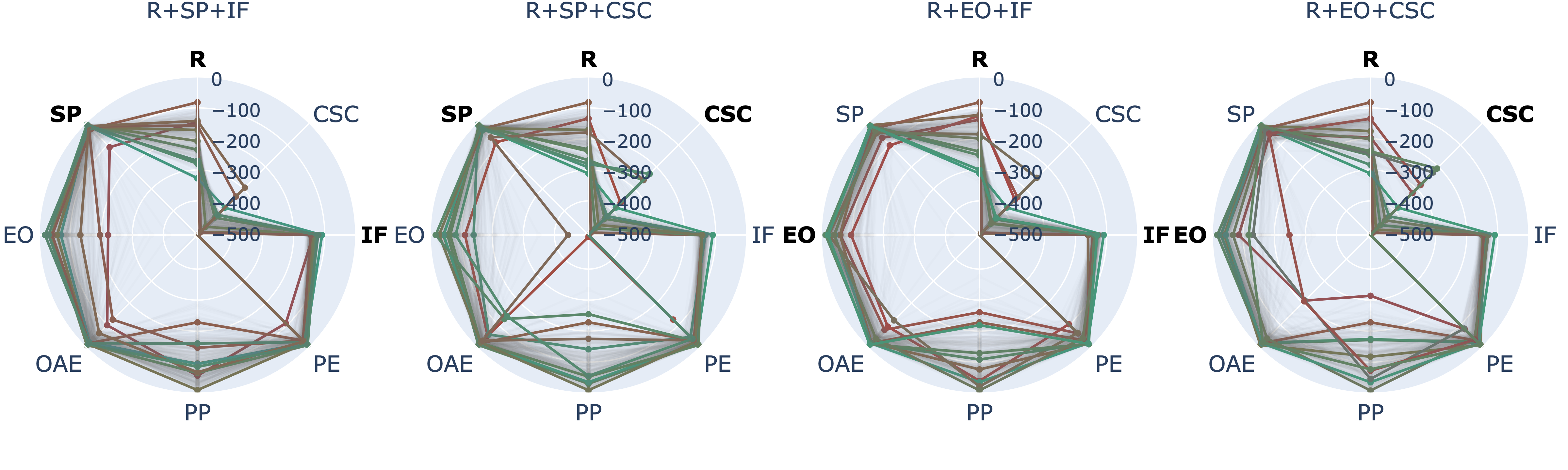}
    \caption{Representative set of learned fraud detection policies when optimising the performance reward (R), a group fairness notion (SP or EO) and an individual fairness notion (IF or CSC). In contrast to the job hiring setting, learned fraud detection policies have larger differences across the objectives. Moreover, prioritising R results in a lower PP outcome across all scenarios.}
    \label{fig:fraud_multi}
\end{figure}

In the fraud detection setting (Figure \ref{fig:fraud_multi}), we note that most trade-offs have larger differences across the performance reward R and group fairness notions such as EO and PP. 
Note how IF does not change considerably, regardless of the requested objectives, compared to CSC.
We emphasise that the combination of requested objectives impacts the obtainable trade-offs. This further highlights the need for multiple fairness notions, which should be chosen by stakeholders with the necessary domain expertise. 
While the performance reward (R) and consistency score complement (CSC) were easier to optimise in the job hiring setting, we observe the opposite effect in the fraud detection setting. We hypothesise this is caused by the context of the problem. Concretely, in job hiring it is easier to find applicants with similar attributes that must be treated similarly. For example, individuals with the same qualifications should receive the same decision as their most similar neighbours under CSC. As IF considers the probability distribution over the actions, and compares all individuals, it is more difficult for the agent to provide the appropriate treatment.
In contrast, in the fraud detection setting it is not necessarily the case that similar transactions are all fraudulent or all genuine. This makes it more difficult to ensure equal treatment with regards to CSC. Moreover, fraud detection constitutes anomalies, making it more likely that most transactions are ignored, providing better (but possibly misleading) results for IF.

\subsection{Influence of distance metric on individual fairness}

\begin{figure}[h!]\centering
  \begin{subfigure}[h!]{0.4\linewidth}\centering
    \includegraphics[width=\linewidth]{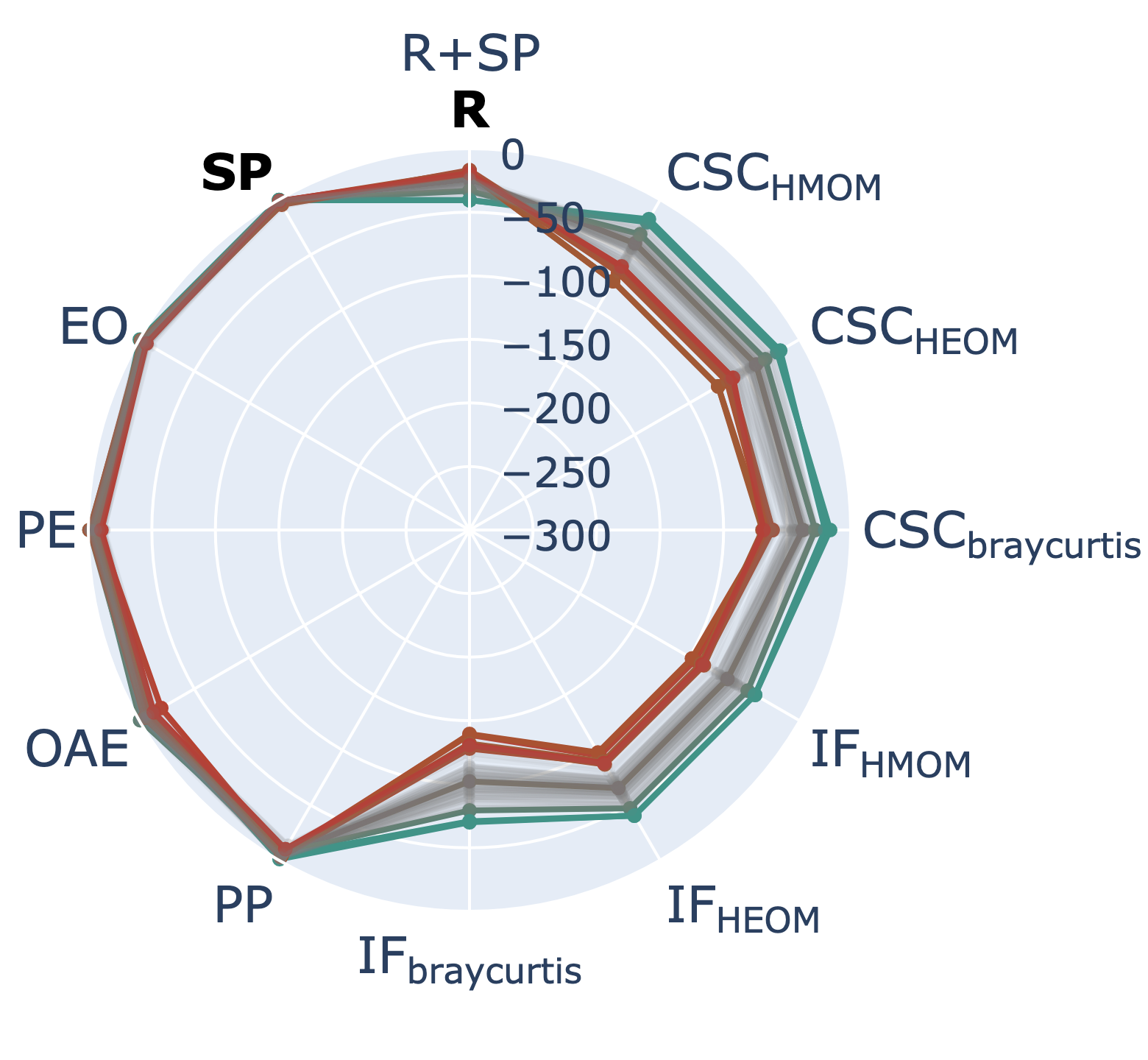}
    \caption{Job hiring}
    \label{fig:job_dist_all}
  \end{subfigure}
    \begin{subfigure}[h!]{0.4\linewidth}\centering
    \includegraphics[width=\linewidth]{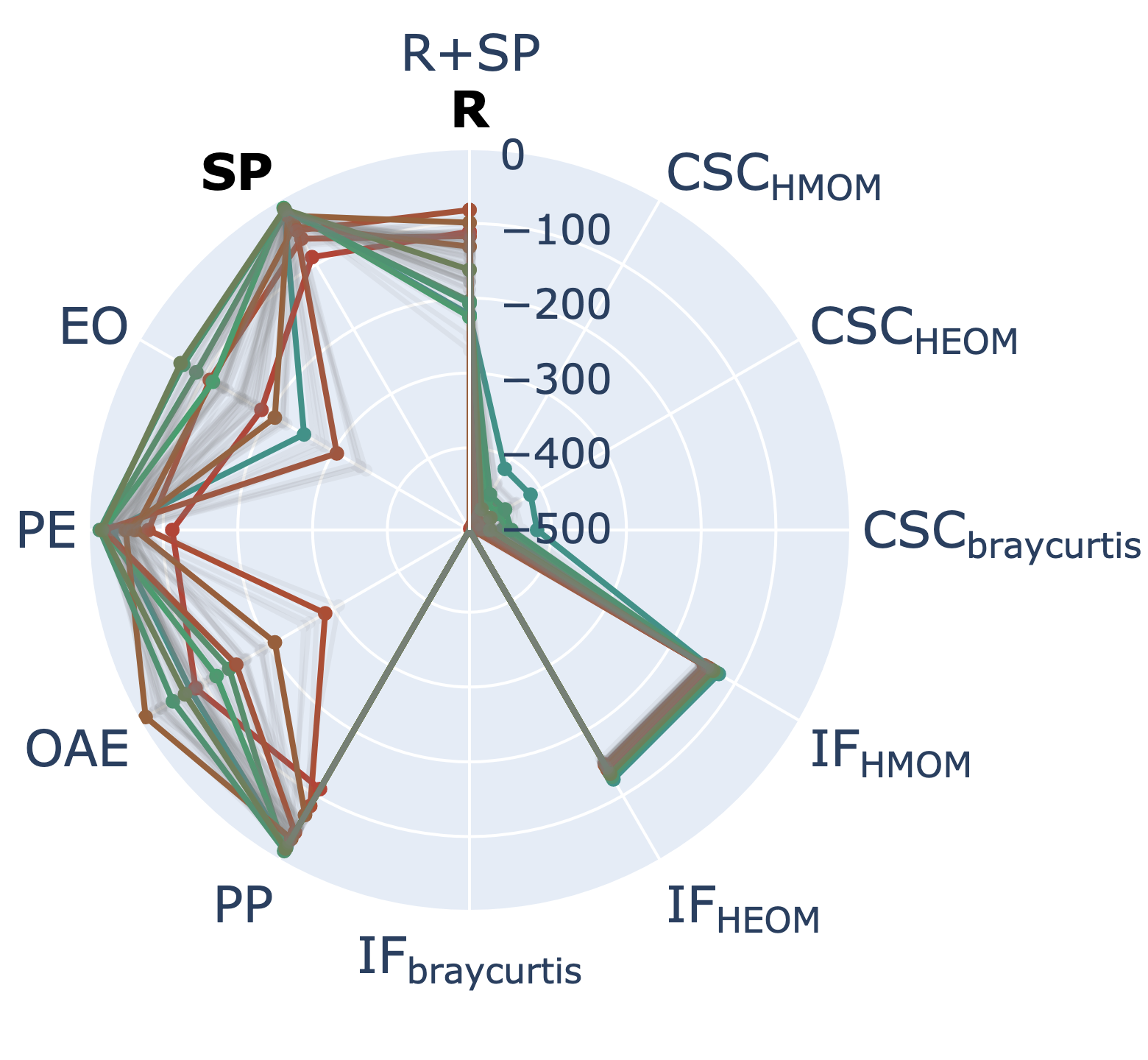}
    \caption{Fraud detection}
    \label{fig:fraud_dist_all}
  \end{subfigure}
  \caption{Representative set of learned policies when optimising the performance reward (R) and statistical parity (SP). Showing results for IF and CSC under different distance metrics. Both individual fairness notions achieve similar outcomes under different distance metrics in the job hiring setting. This may indicate that the choice in distance metric has a limited impact on the measured fairness. In contrast, in the fraud detection setting, individual fairness IF under the braycurtis distance metric provides a different outcome. We highlight that the choice in distance metric in this scenario is more critical, as the measured individual fairness achieves different outcomes and can therefore negatively influence the interpretability of the results.}
    \label{figure:dist_all}
\end{figure}

To investigate the impact of the chosen distance metric on individual fairness notions, we run an experiment where the agent is asked to optimise for the performance reward (R) and statistical parity (SP). In the job hiring setting (Figure \ref{fig:job_dist_all}), we show a representative set of the learned policies, along with their performance for the individual fairness notions. We observe similar results across the individual fairness notions, with the exception of individual fairness (IF) under the braycurtis distance metric. Due to the similarity of all individual fairness results, the learned trade-offs indicate that in this job hiring setting, the choice of distance metric is less impactful.
Note that job hiring policies which prioritise the performance reward and statistical parity only consistently reach low individual fairness across all distance metrics. However, all group fairness notions are well aligned in this scenario.

For the fraud detection setting (Figure \ref{fig:fraud_dist_all}), we observe a larger effect on individual fairness notions under the different distance metrics. On the one hand, there is a greater difference between the fairness notions IF and CSC. On the other hand, we observe a large difference for IF if we compare using braycurtis against HEOM or HMOM as a distance metric. Even though the same fairness notion is used, the outcome implies significantly different results on the fairness of the policy. This emphasises the need for policy makers to choose an appropriate distance metric to enforce fairness between the desired individuals, such that fairness for the desired comparisons are reported.
An overview of the learned policies under different distance metrics for IF can be found in \supplementary{\ref{appendix:job_results}} and \ref{appendix:fraud_results}. There, we also note a low variability across the learned policies for job hiring, while fraud detection policies are more impacted by the chosen distance metric.

\subsection{Historical bias}

\paragraph{Different population distributions}

For the job hiring setting, we consider the Belgian population informed by the official statistics registry of Belgium \citep{STATBELBelgianPop}, and 2 skewed population distributions. The first skewed population distribution (gender) skews the originally equal proportions for gender such that 70\% of applicants are men and 30\% women. The second skewed population distribution (nationality-gender) focuses on the combination of nationality and gender, such that foreign women become a minority. 
Concretely, we skew the true proportions to $\{ (\mathrm{nationality} = Belgian, \mathrm{gender} = man): 30\% \rightarrow 40\%, (\mathrm{nationality} = Belgian, \mathrm{gender} = woman): 31\% \rightarrow 40\%, (\mathrm{nationality} = foreign, \mathrm{gender} = man): 20\% \rightarrow 15\%, (\mathrm{nationality} = foreign, \mathrm{gender} = woman): 19\% \rightarrow 5\% \}$. Note that, while foreign women are a minority in this population, the total percentage of women (Belgian and foreign) in the population is approximately 45\%.
%

%
%

\begin{figure}[h!]
    \centering
    \includegraphics[width=1\linewidth]{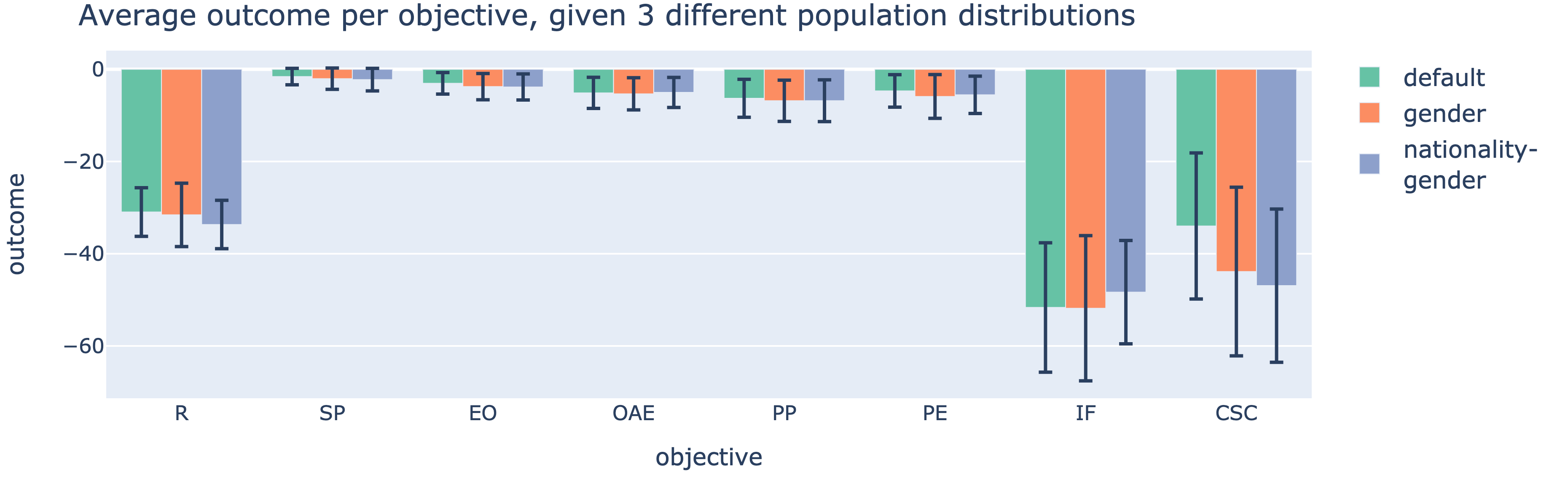}
    \caption{Average outcome with standard deviation per objective of learned job hiring policies when optimising the performance reward (R), statistical parity (SP) and individual fairness (IF) under different population distributions. 
    We note that encountering a different population distribution does not have a substantial negative impact on the fairness of the agent. The outcome for CSC is most influenced by the population distribution. Based on individual fairness definitions, similar individuals should receive a similar treatment. Consequently, we hypothesise that a skewed population distribution makes it more likely for minorities to be nearest neighbours with less similar individuals.
    }
    \label{fig:job_populations_bar}
\end{figure}

As the learned policies have a low variability over the achieved trade-offs, we present the average outcome per objective, along with its standard deviation. Figure \ref{fig:job_populations_bar} shows the average outcome of learned job hiring policies under the different population distributions. 
Overall, the agent is able to achieve higher outcomes under the default population, highlighting that it is easier to act fairly if everyone in the population is well represented, i.e., there is a similar amount of people in both groups and individuals have enough similar individuals to be compared to. When there is a 70\% majority of men in the gender skewed population, we observe lower outcomes across all objectives.
Analogously, in the nationality-gender population, where foreign women are a minority, the outcomes are slightly lower, with IF as an exception. We hypothesise that, due to individual fairness comparing all individuals in a pairwise fashion, and the exclusion of sensitive features in the distance metric used, it is more robust against changes in minorities of the population.
The representative policies for these results can be found in \supplementary{\ref{appendix:job_results}}.

\paragraph{Reward discrepancy}

In real-world problems, reward signals may be complex and difficult to understand. Consequently, they may be, often unintentionally, defined in a biased manner. As mentioned in Section \ref{section:jobhiring}, job hiring processes and their objectives have historically been biased and as such can cause undesired behaviour to be learned by agents trained on this data.
To investigate the strengths and shortcomings of all fairness notions and our framework, we consider two additional reward configurations next to the currently objective configuration, which assigns a similar reward to individuals who apply to the same team with the same qualifications. The second configuration assigns a $+0.1$ bias to men, while the third configuration assigns a $+0.1$ bias to Belgian men.

\begin{figure}[H]\centering
  \begin{subfigure}[h!]{1\linewidth}\centering
    \includegraphics[width=\linewidth]{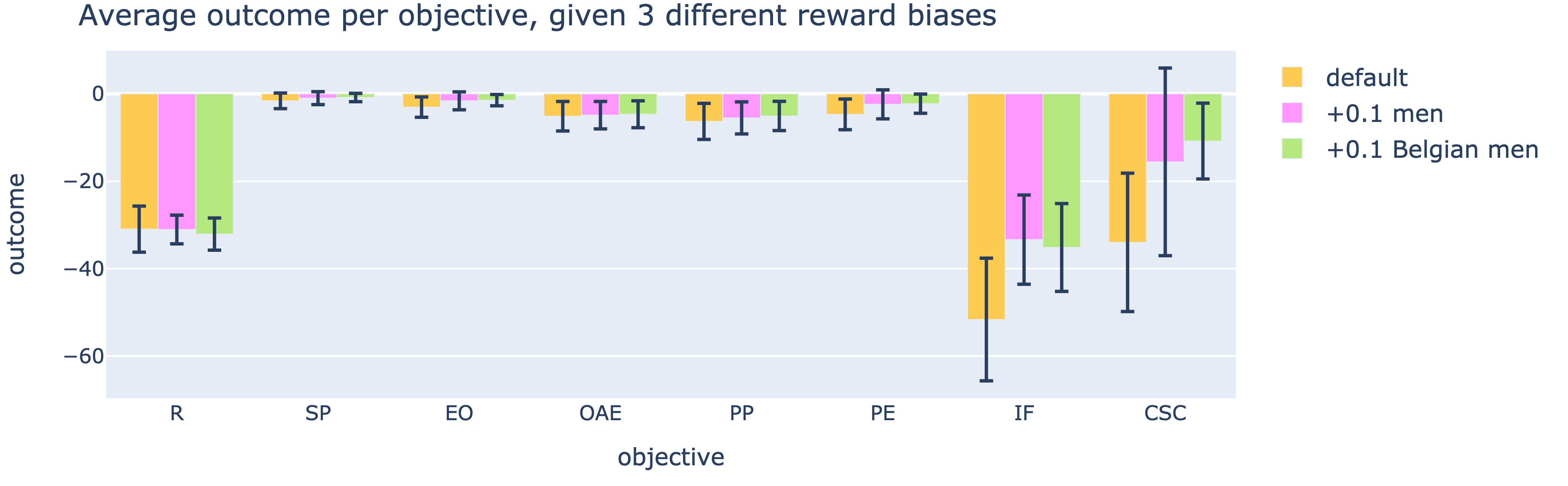}
    \caption{The default Belgian population}
    \label{fig:job_bias_default_bar}
  \end{subfigure}
    \begin{subfigure}[h!]{1\linewidth}\centering
    \includegraphics[width=\linewidth]{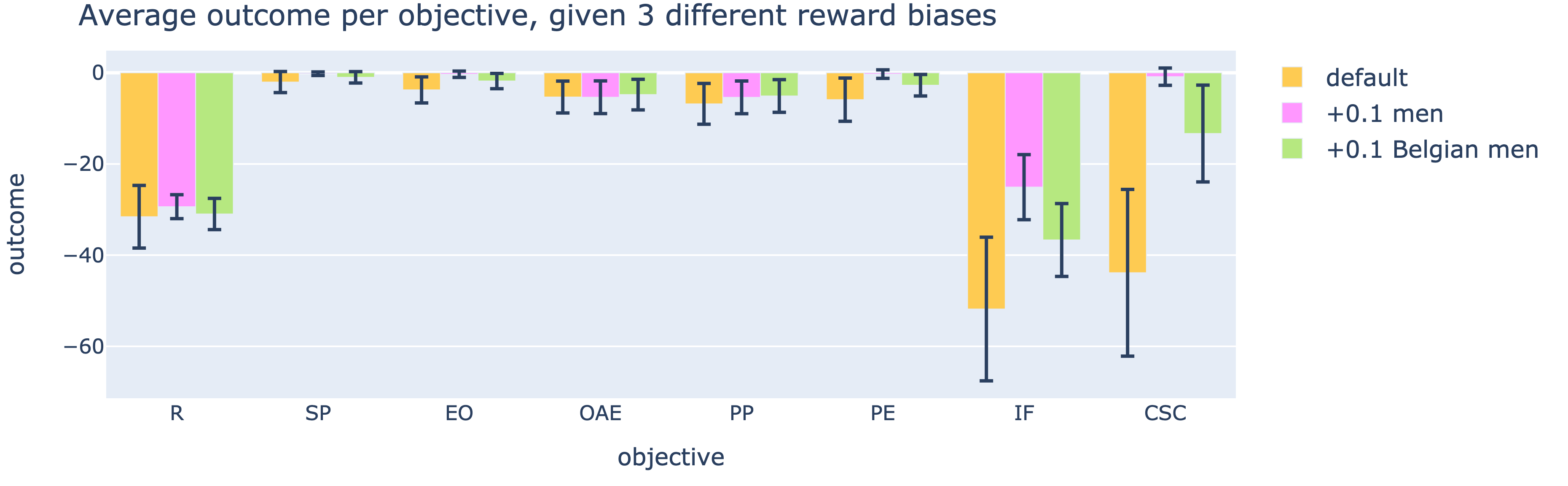}
    \caption{Belgian population, with a majority of 70\% male job applicants.}
    \label{fig:job_bias_gen_bar}
  \end{subfigure}
      \begin{subfigure}[h!]{1\linewidth}\centering
    \includegraphics[width=\linewidth]{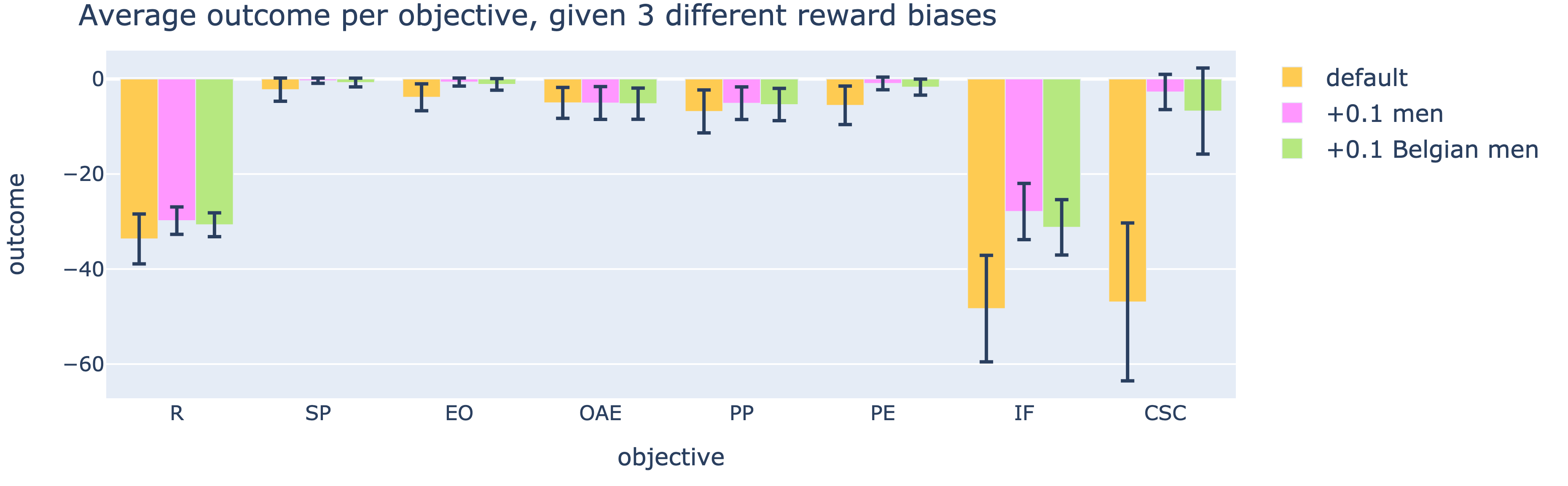}
    \caption{Belgian population, with a 5\% foreign women minority.}
    \label{fig:job_bias_nat_gen_bar}
  \end{subfigure}
  \caption{Average outcome with standard deviation per objective of learned policies when optimising the performance reward (R), statistical parity (SP) and individual fairness (IF) under scenarios with different reward biases.}
    \label{figure:job_bias_all_bar}
\end{figure}
Figure \ref{figure:job_bias_all_bar} shows the learned policies when optimising for R, SP and IF for each reward configuration and population distribution. As in the previous results, the agent achieves a high group fairness for most of the policies. Depending on the (lack of) bias, the agent finds different trade-offs. The policies in the default unbiased configuration result in the lowest individual fairness across all populations. Counterintuitively, the agent is able to obtain higher individual fairness in the biased configurations. The average outcome for CSC in particular is higher in the skewed populations under biased scenarios.
Despite that all group fairness notions check for unfairness between men and women, they remain mostly unaffected by the population distribution and bias scenario. Furthermore, we note a small improvement for the performance reward and across all fairness notions in the biased scenarios when compared with the default scenario.
Moreover, in nearly all representative policies in the gender-skewed population where men receive a $+0.1$, CSC achieves near-perfect fairness.
As both IF and CSC compare individuals directly, they are better suited to deal with multiple sensitive attributes. Consequently, by optimising for IF, the agent is able to learn policies that reach a high individual fairness (IF and CSC) in all 3 reward configurations. We hypothesise, based on the previous experiments, that group fairness notions and individual fairness notions are mostly aligned in this job hiring setting. Therefore, improving one type of fairness notion results in some improvement in the other type.

\begin{figure}[h!]
    \centering
    \includegraphics[width=1\linewidth]{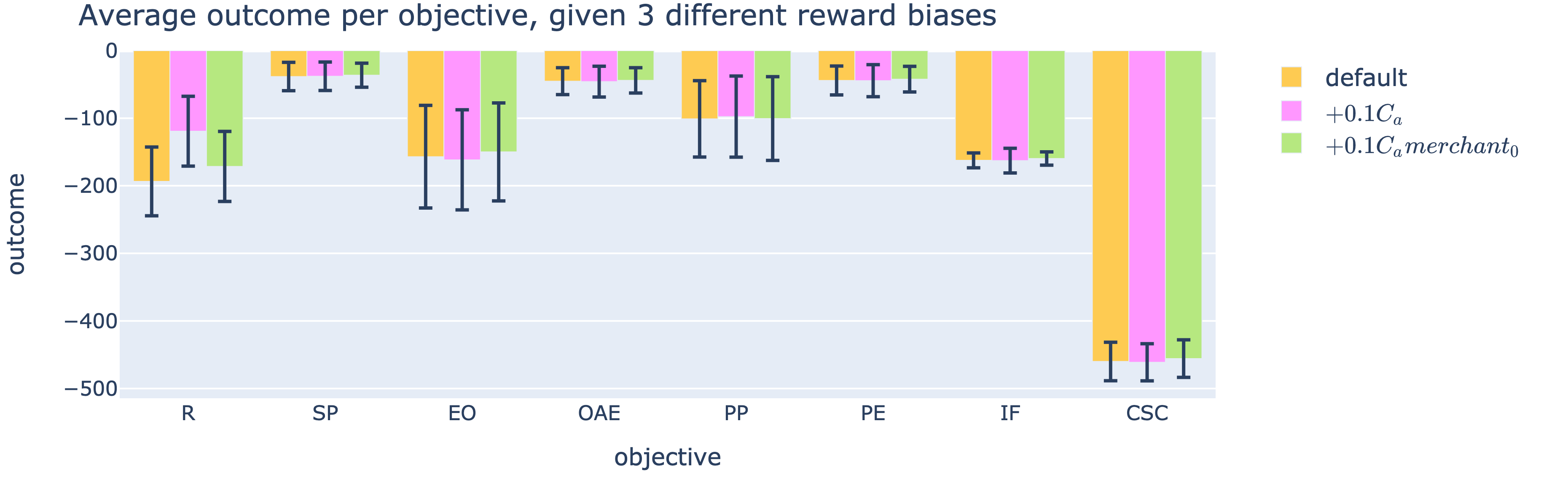}
    \caption{Average outcome with standard deviation per objective of learned policies when optimising the performance reward (R), statistical parity (SP) and individual fairness (IF). Showing results under different reward biases.}
    \label{fig:fraud_bias_bar}
\end{figure}

We explore the impact of bias in fraud detection, under two different reward biases. The first bias scenario assigns $+0.1$ to transactions from continent $\cA$, while the second bias scenario assigns $+0.1$ to transactions from continent $\cA$ to merchant $0$. Figure \ref{fig:fraud_bias_bar} shows the average outcome for the bias scenarios compared to the default. Interestingly, the agent is able to learn policies that achieve a higher performance reward under the biased scenarios.\footnote{Continent $\cA$ has less fraud with a 44\% chance of a fraudulent transaction if it originates from there. In contrast, continent $\cB$ transactions are 73\% likely to be fraudulent. Consequently, the agent is encouraged to proportionally let more transactions from $\cA$ go unchecked compared to continent $\cB$.}
Despite an introduced bias, the agent maintains a similar performance regarding all fairness notions. This highlights that the policies are unaffected by the biased reward when optimising for fairness notions alongside it.

By using a multi-objective approach, where a (possibly biased) performance reward is optimised together with group and individual fairness notions, we show that the agent is still able to learn fair policies across multiple fairness notions, at little cost to the performance reward. 
Furthermore, there are few significant differences in treatment across the scenarios, even though the group fairness notions do not consider sub-group implications created by the different population distributions and reward configurations combined. As the agent also optimised for individual fairness next to group fairness, any gerrymandering attempts by the agent are discouraged \citep{Kearns2018}, resulting in more overall fair policies.

\subsection{History size}

\begin{figure}[h!]
    \centering
    \includegraphics[width=1\linewidth]{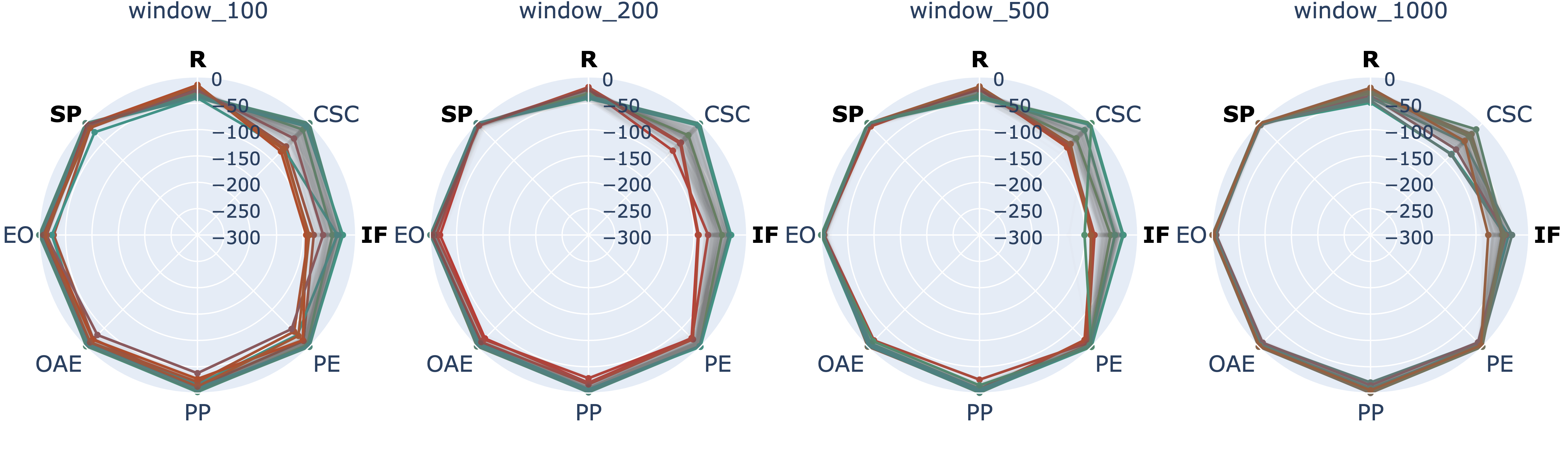}
     \caption{Representative set of learned job hiring policies when optimising the performance reward (R), statistical parity (SP) and individual fairness (IF). Showing results for histories with different sliding window sizes. As the window size increases, the agent learns policies with consistently higher group fairness. Moreover, increasing the window size allows the agent to learn policies that achieve a higher individual fairness while maintaining a higher reward.}
    \label{fig:job_windows}
\end{figure}

\paragraph{Sliding window history}
Figure \ref{fig:job_windows} shows the impact of different history sizes on the learned job hiring policies. In general, we observe that group fairness notions are the least impacted by the chosen history sizes. As group fairness notions focus more on statistical measures over groups of individuals, it is easier for the agent to provide equal treatment over the groups. In contrast, individual fairness notions (IF and CSC) are more impacted by the sliding window size of the history. Note that for both individual fairness notions, across all window sizes, increasing the reward often reduces individual fairness and vice versa. This indicates that these objectives may be conflicting.
We refer to \supplementary{\ref{appendix:job_results}} for an overview of these policies.

\begin{figure}[h!]
    \centering
    \includegraphics[width=1\linewidth]{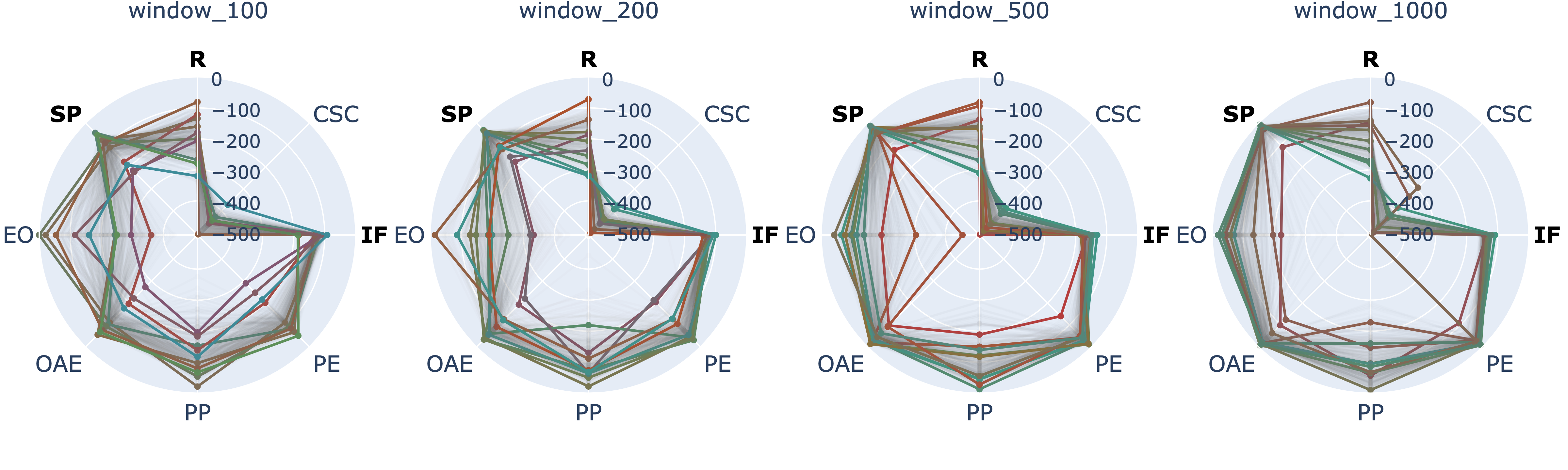}
    \caption{Representative set of learned fraud detection policies when optimising the reward  (R), statistical parity (SP) and individual fairness (IF). Showing results for histories with different sliding window sizes. As the window size increases, the performance across group fairness notions improves. We attribute this to the reliance of group fairness on statistical measures over the groups, which give the agent more time to correct any unfairness with regards to previous actions.}
    \label{fig:fraud_windows}
\end{figure}

Figure \ref{fig:fraud_windows} shows the learned fraud detection policies when optimising under different history sizes. We note larger differences across the objectives compared to the job hiring setting. Note how the sliding window size impacts which trade-offs can be found across all objectives. This highlights that the fraud detection setting requires a more careful consideration when choosing a history window size, as the learned trade-offs are strongly influenced by it. Furthermore, the agent learns less similar policies across all group fairness notions. We attribute this to the different fraudulent base rates, making it challenging for the agent to correctly identify fraudulent transactions for both continents equally.

\begin{figure}[b!]
    \centering
    \includegraphics[width=1\linewidth]{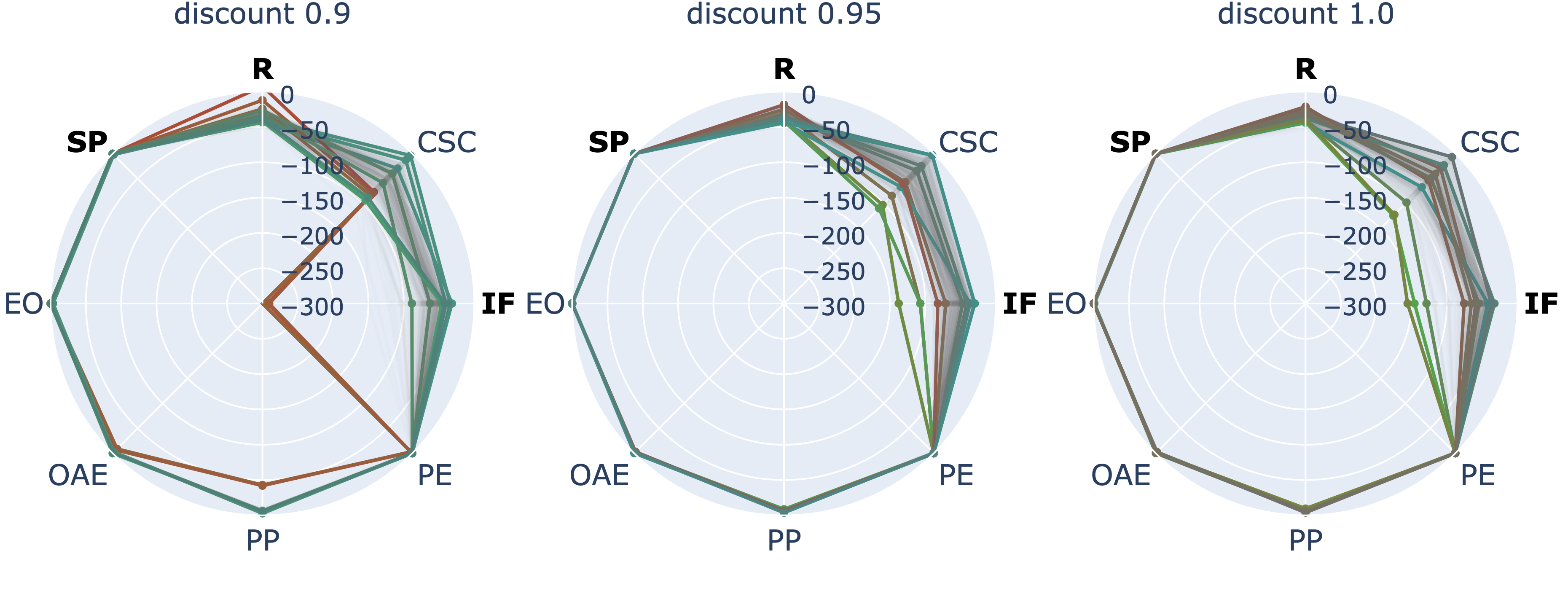}
    \caption{Representative set of learned job hiring policies when optimising the performance reward (R), statistical parity (SP) and individual fairness (IF). Showing results for a discount history with different discount factors.}
    \label{fig:job_windows_discount}
\end{figure}

\begin{figure}[h!]
    \centering
    \includegraphics[width=1\linewidth]{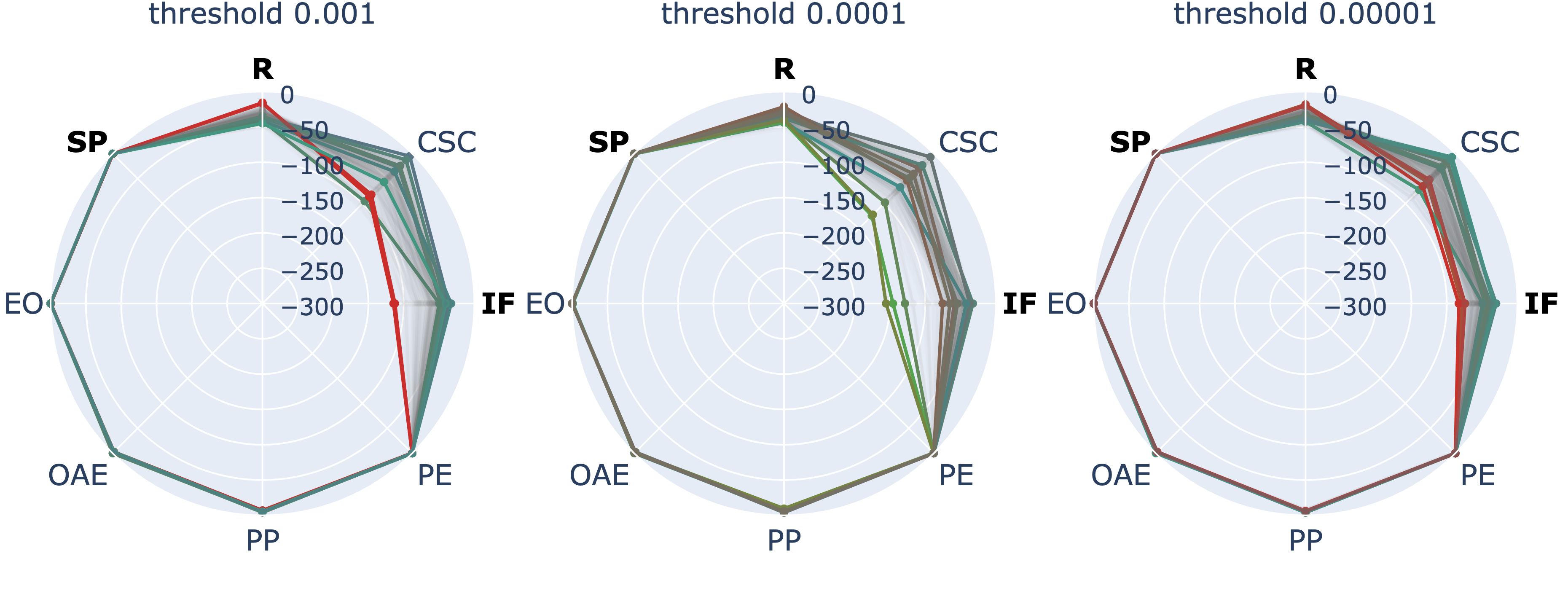}
    \caption{Representative set of learned job hiring policies when optimising the performance reward (R), statistical parity (SP) and individual fairness (IF). Showing results for a discount history with different discount thresholds.}
    \label{fig:job_windows_threshold}
\end{figure}

\paragraph{Discounted history}
We explore the impact of using a discounted history on the learned policies. Figure \ref{fig:job_windows_discount} shows the learned job hiring policies under different discount factors. We observe similar results across the trade-offs for a discount factor of 0.95 and 1.0. For a 0.9 discount factor, the agent learns some trade-offs with lower individual fairness in favour of a higher reward and statistical parity. When using different discount thresholds (Figure \ref{fig:job_windows_threshold}), we observe similar performance across the group fairness notions. In contrast, the agent achieves different trade-offs regarding the performance reward and individual fairness notions. Note that policies that achieve a high performance reward do this at the cost of a lower individual fairness.

\begin{figure}[h!]
    \centering
    \includegraphics[width=1\linewidth]{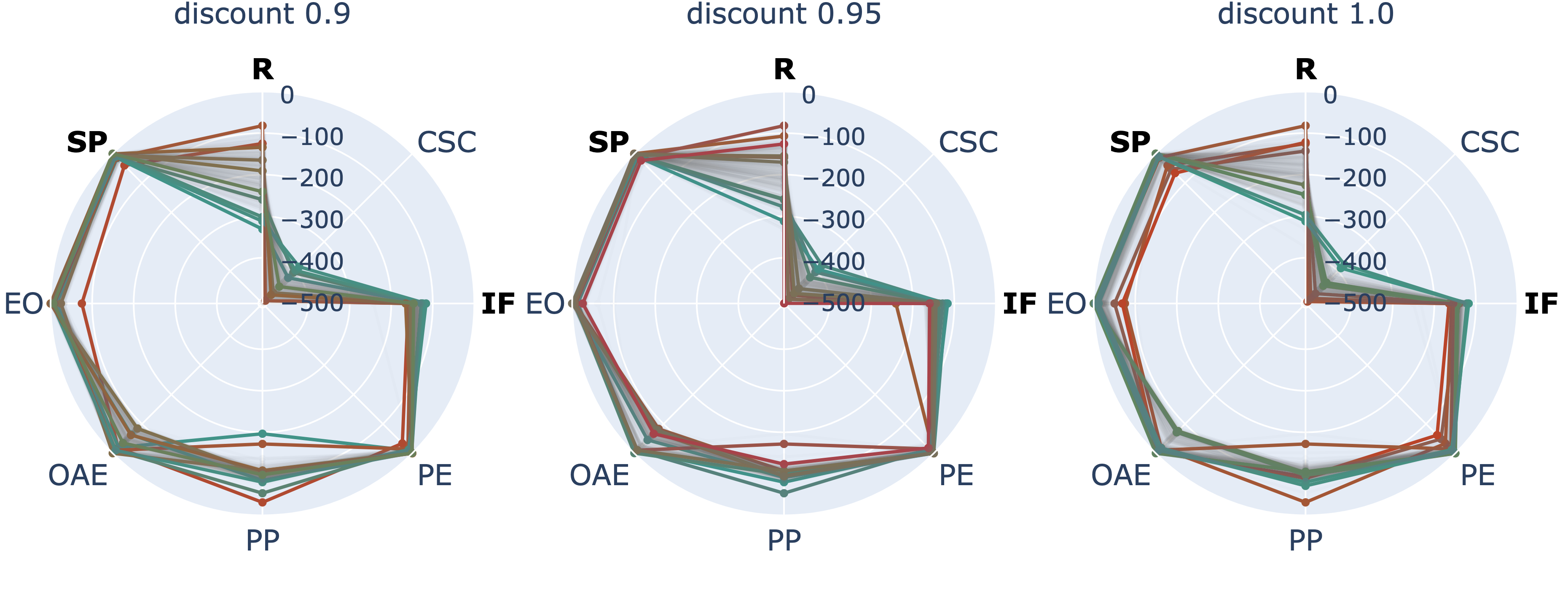}
    \caption{Representative set of learned fraud detection policies when optimising the performance reward (R), statistical parity (SP) and individual fairness (IF). Showing results for a discount history with different discount factors.}
    \label{fig:fraud_windows_discount}
\end{figure}

\begin{figure}[h!]
    \centering
    \includegraphics[width=1\linewidth]{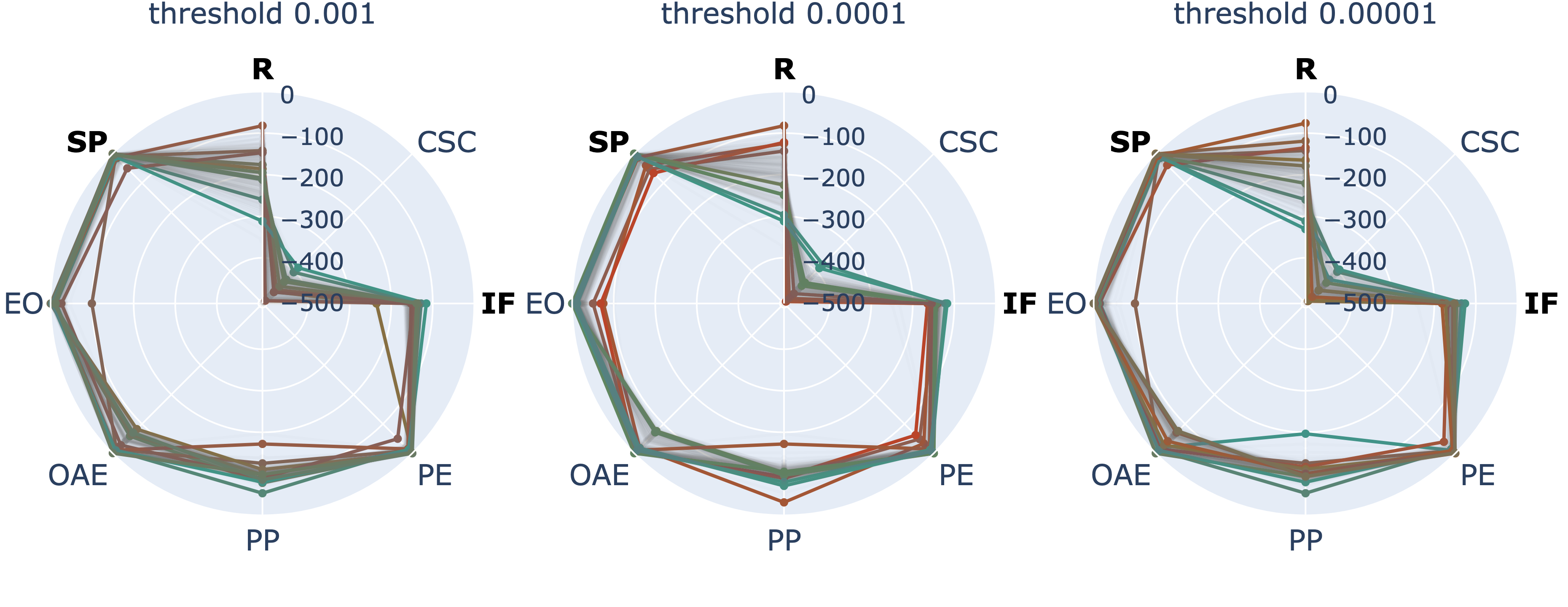}
    \caption{Representative set of learned fraud detection policies when optimising the performance reward (R), statistical parity (SP) and individual fairness (IF). Showing results for a discount history with different discount thresholds.}
    \label{fig:fraud_windows_threshold}
\end{figure}

In the fraud detection scenario, we observe similar trade-offs across the different discount factors (Figure \ref{fig:fraud_windows_discount}) and thresholds (Figure \ref{fig:fraud_windows_threshold}). The trade-offs differ most by the performance reward, predictive parity and consistency score complement. Moreover, improving the performance reward happens at the cost of consistency score complement in all cases. This indicates that these objectives are not aligned in this scenario, regardless of the chosen history size. 
While the history size influences the obtained trade-offs for both the job hiring and fraud detection setting, it is not a solution for unaligned objectives.

\begin{figure}[h!]
    \centering
    \includegraphics[width=1\linewidth]{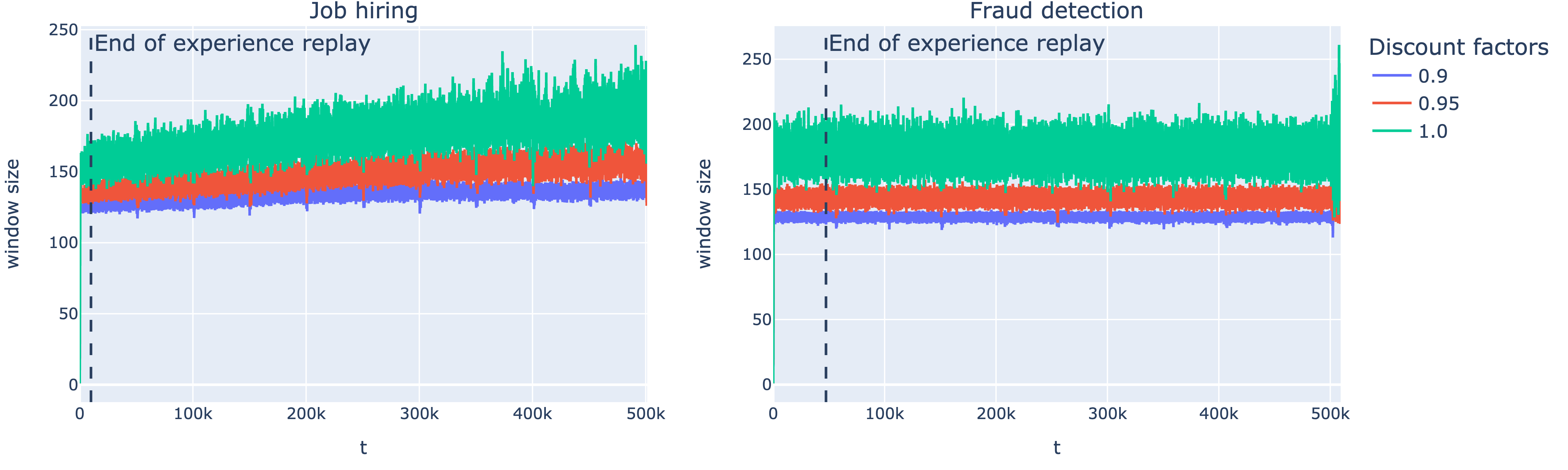}
    \caption{Average window size over time under different discount factors when optimising the performance reward (R), statistical parity (SP) and individual fairness (IF) for job hiring (left) and fraud detection (right). We observe more fluctuations as the discount factor increases. As a lower discount factor reduces the impact of earlier interactions, these interactions have a negligible impact on the fairness computations. This causes early interactions to continuously be discarded, keeping the window size more stable.}
    \label{fig:discount_t}
\end{figure}

To further investigate the discounted history, we plot the evolution of the window size over time during the training process. We present some results here and provide additional results for job hiring and fraud detection in Appendix \ref{appendix:job_results} and \ref{appendix:fraud_results}, respectively.
Figure \ref{fig:discount_t} shows the evolution of the window size under different discount factors when optimising R-SP-IF for job hiring and fraud detection. A lower discount factor results in a smaller history size in both settings. As the influence of early transactions on the fairness computation decreases faster given a lower discount factor, the average window size is more stable.
In the job hiring setting, we note a slow increase over time in the average window size. We hypothesise that this trend is caused by a change in the learned policies, where the agent performs better when fairness is considered over a larger history size.
In the fraud detection setting, the average window size remains constant throughout the experiments, with only a notable increase in variation at the end. We hypothesise that, as for the job hiring setting, this is caused by a change in the learned policies.

\begin{figure}[h!]
    \centering
    \includegraphics[width=1\linewidth]{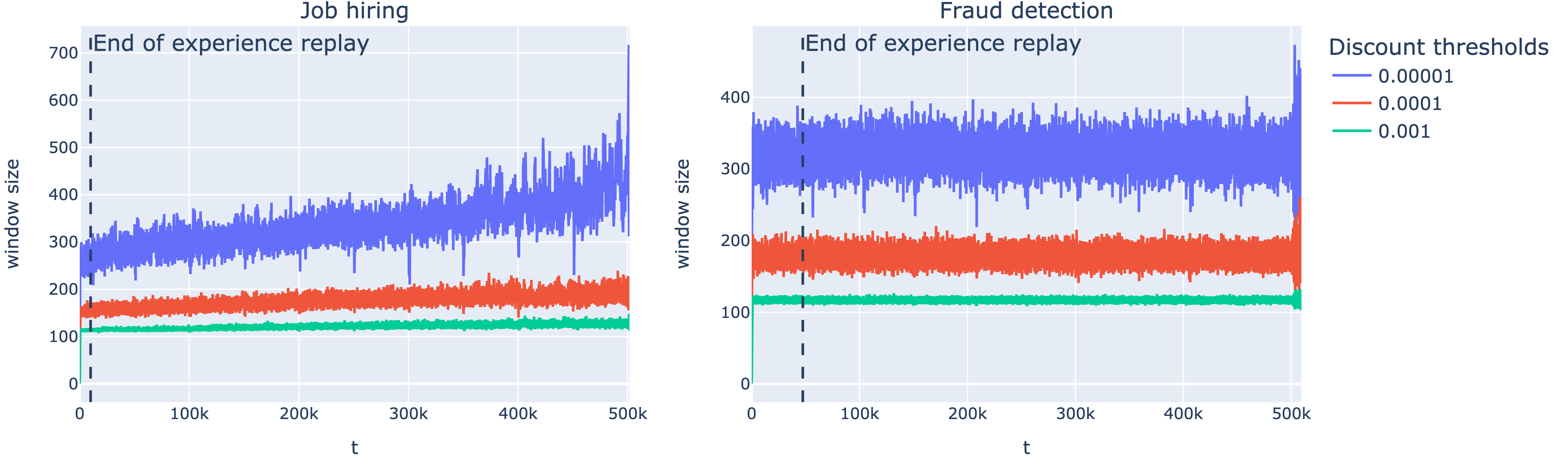}
    \caption{Average window size over time under different discount thresholds when optimising the performance reward (R), statistical parity (SP) and individual fairness (IF) for job hiring (left) and fraud detection (right). Note how the window size fluctuates less given a lower threshold in both settings, as the difference in fairness computations with and without early interactions is directly influenced by the history size. Therefore, having a higher threshold value results in more early interactions being frequently discarded, keeping the history size smaller and more consistent over time.}
    \label{fig:discount_threshold_t}
\end{figure}

Figure \ref{fig:discount_threshold_t} shows the evolution of the window size under different thresholds. Reducing the threshold tolerance results in a larger history size in both settings. Intuitively, if a larger computational margin is allowed between fairness computations with and without an interaction, less interactions will be stored as their contribution to the fairness computation is considered negligible. In contrast, a small threshold results in a larger history size. As with the discount factor results, we observe a slow increase over time in the average window size in the job hiring setting, whereas the results in the fraud detection setting remain more constant.

\section{Discussion}
\label{Section:discussion}

We propose a framework to investigate the use of fairness notions in RL. 
We establish a formulation of fairness notions that can be used as additional reward signals following a multi-objective learning approach. Based on this formulation, we classify distinct fairness settings grounded in real-world problems.
We highlight the need of multiple fairness notions, particularly ensuring both group and individual fairness simultaneously. Due to the context dependency of fairness, we show how requested fairness notions can be conflicting with the performance reward. As such, we argue the multi-objective aspect is crucial in the development of the fairness framework.

By formulating fairness notions in terms of the history defined, we establish a formal way to reason about fairness notions as reward functions. Yet, as maintaining the full history will prove computationally intractable for most real-world applications, a major challenge remains to construct approximate fairness notions. 
Individual fairness notions in particular require pair-wise comparisons of individuals, in contrast to group fairness notions that rely on the statistical measures of each group.
One research direction is to consider a sliding window approach, where the history is kept for a fixed or varying number of steps \citep{Ortiz2011}. 
%
We highlight that the choice of the particular distance metric impacts the individual fairness notions and therefore the learned policies. Moreover, the size of the history impacts the outcomes of fairness notions, especially when considering multiple fairness notions simultaneously.

Generalising fairness notions to continuous actions presents an interesting research venue to extend fairness to a wider array of problem settings.
In the field of regression, algorithms produce a scalar value rather than a discrete action from a predefined set. Consequently, regression compares actions based on how much they differ and can detect correlations between the action and one or more sensitive features \citep{Komiyama2018}. This makes it an interesting approach for comparing actions in RL.
Additionally, a hybrid notion of fairness is proposed to define other types of fairness than strictly between groups or individuals \citep{Berk2017}, which would be an interesting addition to RL as well.

Within the overarching topic of ethics, work on explainable AI focuses on making algorithms interpretable and provides explanations for their decisions \citep{Goodman2017}. While explainability aims to provide transparency regarding an agent's decisions and policy, fairness focuses on whether or not the agent makes decisions which conform to expected impartial treatment. We argue that fairness is just as crucial as other ethical principles in the development of responsible AI. To advance towards fair decision support systems, we envision the integration of fairness notions with explainable reinforcement learning, enabling fairness to be explicitly considered when presenting and interpreting policies for decision makers.

\section*{Acknowledgements}
A.C. is funded by the Fonds voor Wetenschappelijk Onderzoek (FWO) via fellowship 1SF7825N and received funding from the Research Council of the Vrije Universiteit Brussel (OZR-VUB) through OZR mandate OZR3819.
C.M.J. gratefully acknowledges support by the National Science Foundation (NWO) under Grant No. 024.004.022 (Hybrid Intelligence) and Grant No. 024.005.017 (Algorithmic Society). 
P.L. gratefully acknowledges support from FWO via postdoctoral fellowship 1242021N and the Research Council of the Vrije Universiteit Brussel (OZR-VUB via grant number OZR3863BOF). P.L. acknowledges support from FWO grant G059423N. A.N. acknowledges funding from the iBOF DESCARTES project (reference: iBOF-21-027).
This research was supported by funding from the Flemish Government under the "Onderzoeksprogramma Artifici\"{e}le Intelligentie (AI) Vlaanderen" program.
Any opinions, findings, and conclusions or recommendations expressed in this material are those of the authors and do not necessarily reflect the views of the grant organisations.

\section*{Declaration of competing interest}
The authors declare that they have no known competing financial interests or personal relationships that could have appeared to influence the work reported in this paper.

\section*{Declaration of generative AI and AI-assisted technologies in the writing process.}
During the preparation of this work the authors used generative AI in order to  improve language clarity. After using this tool/service, the authors reviewed and edited the content as needed and take full responsibility for the content of the published article.

\section*{CRediT authorship contribution statement}
Alexandra Cimpean: Conceptualization, Methodology, Software, Investigation, Writing – original draft, Data curation, Visualization. Nicole Orzan: Writing – review \& editing. Catholijn Jonker: Conceptualization, Investigation, Writing – review \& editing. Pieter Libin: Supervision, Conceptualization, Methodology, Writing – original draft, Writing – review \& editing. Ann Now\'{e}: Supervision, Conceptualization, Methodology, Writing – review \& editing.

\bibliography{references.bib}

\newpage
\appendix

\section{Job hiring $f$MDP}
\label{suppl:jobhiring_mdp}

For the job hiring setting, we create a simulator for building a team of employees that supplies at each timestep a new candidate to the agent.
To apply RL, we define the job hiring setting as a Markov Decision Process (MDP) \citep{Sutton2018}. The MDP is represented by the tuple $\{\setrlstates, \setactions, \setrewards, \rltransition\}$, consisting of a set of states $\setrlstates$, a set of actions $\setactions$, a set of rewards $\setrewards$ and a transition function $\rltransition$.

\paragraph{State}
Each timestep $t$, the agent is presented with the current state $\rlstate_t \in \setrlstates$, which specifies the company's current composition $\statepotential_t$ of hired applicants and a new job applicant $\statecandidate_t$ to assess. 
A job applicant $\statecandidate_t$ is represented by the following set of features: their gender, age, years of experience, degree, extra degree, marital status, nationality and their ability to speak four languages $\languages$. 
For the purpose of this study, we consider gender, nationality, age and marital status sensitive features, which should not be taken into account when hiring a job applicant. 
To generate realistic applicants, we sample from the distribution of the Belgian active employed and unemployed population provided by the Belgian federal government \citep{STATBELBelgianPop}. For the context of our job hiring scenario, we exclude individuals younger than 18 years from this data. To assign spoken languages to the candidates, we sample based on the most known foreign languages of adults \citep{STATBELLanguages}. 
%
We define the maximum experience of each applicant in function of their age and obtained degrees: $max_e = age - 18 - 3 * degree - 2 * extra\_degree$. We assume a linearly increasing probability for each possible year of experience $year \in [0, max_e]$ for the applicant, equal to
\begin{eqnarray}
\begin{aligned}
        P(year) = \frac{year + 1}{\sum_{y=0}^{max_e} (y + 1)}
\end{aligned}
\end{eqnarray}

The company's state $\statepotential$ is represented by a set of features focusing on the employees' skills. These features consist of the average employee potential $\potential$, the percentage of collected degrees, extra degrees, the combined years of experience and language entropy. We normalise all features based on the desired final team size $K$, such that each applicant can impact the team as much as they would in a full team. We further normalise the combined years of experience such that all features lie in the interval $[0, 1]$. 
%
Based on hired applicants, the company's team composition $\statepotential_t$ is implemented as the proportions of skill and diversity features. For example, the language diversity is represented by four values [0.6, 0.4, 0.2, 0.1] indicating 60\% of the spoken languages is Dutch, 40\% is French, 20\% is English and 10\% is German. On these values the entropy is calculated for the goodness score and reward. Therefore, the state does not contain a list of all employees, but does contain their contributions to the team's skills 
such that the agent can decide for a new candidate if they are a good fit.
Given $K$ the desired final team size and $k$ the number of employees (i.e., hired applicants), we define the company's potential based on the degree $\degree$, extra degree $\extradegree$ and experience $\experience$ each employee holds on average. Concretely, the potential of the employees follows a Gaussian with mean
\begin{eqnarray}
\begin{aligned}
        \potential = \frac{1}{K} \sum_{i=1}^k \frac{1}{3} |\{f^i \in \{\degree, \extradegree, \experience\}: f^i \neq 0\}|
\end{aligned}
\end{eqnarray}
and a standard deviation of 0.01.
For the estimated company potential given a new applicant, we use the same distribution given the assumption that an applicant's resume based on these features does not perfectly match the applicant's potential once hired for the job.

\paragraph{Goodness score}
To define how suitable each candidate is for hire, we define an objective goodness score $\goodness_t \in [-1, 1]$ based on how the estimated new company state $\hat{\statepotential}_{t+1}$ would differ from the current $\statepotential_t$, should the applicant be hired:
\begin{eqnarray}
\begin{aligned}
        \goodness_t = \frac{K}{N} \sum_{f_t \in \statepotential_t} (\hat{f}_{t+1} - f_t)
\end{aligned}
\end{eqnarray}
with $N$ the number of skill features.
Note that this goodness score is also noisy due to the noise in the current company potential and the estimated new potential. 
Intuitively, the goodness score is higher for applicants who can improve the average potential, have the requested skills and improve the language entropy of the team.

\paragraph{Action and reward}
At each timestep $t$, the agent must choose whether to reject or hire the applicant for a given state $\rlstate_t$. Given the chosen action $\action_t$ for state $\rlstate_t$, the agent receives a reward $\reward_t$ based on the goodness score $\goodness_t$ of the presented applicant. Given the goodness score $\goodness_t$ and threshold $\epsilon$, the reward for hiring an applicant is
\begin{eqnarray}
    \reward_{t,hire} = \goodness_t - \epsilon + \mathcal{N}(0, 0.01)
\end{eqnarray}
We add Gaussian noise to the reward under the assumption that the applicant's qualification may differ slightly from the estimation of the goodness score. This models the employer's uncertainty about the suitability of hired applicants.
The reward for rejecting an applicant is the negative reward of hiring the applicant:
\begin{eqnarray}
    \reward_{t,reject} = -\reward_{t,hire}
\end{eqnarray}

\paragraph{Transition function}
We define the transition function $\rltransition: \setrlstates \times \setrewards \times \setrlstates \times \setactions \rightarrow [0,1]$ as the probability of encountering the next state $\rlstate_{t+1}$ and reward $\reward_t$ given the current state $\rlstate_t$ and action $\action_t$.
To mimic a realistic team composition over time, we allow employees to leave the company based on real job transition probabilities corresponding to their age \citep{STATBELTransitions}. This provides the agent with the additional challenge of replacing lost skills of leaving employees to keep the team balanced.

\paragraph{Feedback signal}
To extend the MDP to an $\fmdpe$, we implement the feedback signal $\feedback_t$ as the correct action $\trueaction_t$ based on the goodness score:
\begin{eqnarray}
    \feedback_t = \trueaction_t
\end{eqnarray}

\section{MultiMAuS $f$MDP}
\label{suppl:frauddetection_mdp}

The fraud detection setting concerns online credit card transactions where multi-modal authentication is used to identify and reject fraudulent transactions. We make the following adaptations to the MultiMAuS simulator \citep{Zintgraf2017}, but keep their default parameters. 

\paragraph{State}
Each hour, a set of customers, both genuine and fraudulent, attempt to make transactions, where each transaction is characterised by the following features:
card id, merchant id, amount, currency, country and the date and hour when the transaction is occurring.
As the agent must check transactions on an individual basis, we consider a new timestep for every transaction request.
At each timestep $t$, the agent observes the current state $\rlstate_t$ containing information about the current company state, and a new transaction to process.
We define two company state features: the proportion of genuine to fraud transactions and the average customer satisfaction.

\paragraph{Reward + action}
For each transaction, the agent must decide whether or not to request an authentication from the customer.
Based on the chosen action $\action_t$, the agent receives a reward
\begin{eqnarray}
    \reward_t = 
        \begin{cases}
        +1 & \mathrm{if\ genuine\ authentication},\\
        -1 & \mathrm{if\ fraudulent\ authentication},\\
        0 & \mathrm{otherwise}
        \end{cases}
\end{eqnarray}

Based on this reward, always asking for authentication results in more fraudulent transactions being caught, as fraudsters are assumed to not be able to provide a second authentication \cite{Zintgraf2017}. 
However, asking for authentication too often reduces the customer's patience in completing transactions. Furthermore, too many authentication requests make it more likely for customers to leave the credit card company. Therefore, the agent must carefully select transactions to check to keep customer satisfaction high, while also catching as many fraudulent transactions as possible.


\paragraph{Feedback signal}
The reward $\reward_t$ specifies the correctness of the action if the agent requests authentication. Consequently, if the reward is positive the transaction is considered genuine, while a negative reward indicates an unsuccessful transaction, caused by a loss in commission or by stolen money requiring the credit company to repay the losses to the client.
To implement a feedback signal $\feedback$, we infer the correctness when authenticating to observe the amount of true positives and false positives.

\section{Policy Visualisation}
\label{suppl:policy_visualisation}

To easily compare the range of possible policy trade-offs across multiple objectives, all objectives are normalised and/or shifted, such that the maximum any objective can reach is 0. As the number of policies is large (more than 30 per seed), we opt to highlight a representative subset of 10 policies for each of the figures. This subset contains the policies with the highest value for each objective, along with randomly sampled policies which differ most from each other and the un-highlighted policies across all policies. Note that we do plot all policies in low opacity. As such, more shaded regions indicate a larger number of policies which obtain similar trade-offs.

\section{Job hiring}
\label{appendix:job_results}

The following subsection contains the representative policies that are visually presented in the Experiments section of the main manuscript.  

\subsection{Single-objective scenarios}
\begin{tiny}
\begin{longtable}[h!]{lllllllll}
\caption{The representative subset of job hiring policies, for different objectives. Results rounded to 5 decimals.} \label{table:job_R-SP-IF-EO-PE-PP-OAE-CSC_b0_dHEOM_w1000} \\
\toprule
Objectives & R & SP & EO & OAE & PP & PE & IF & CSC \\
\midrule
\endfirsthead
\caption[]{The representative subset of job hiring policies, for different objectives. Results rounded to 5 decimals.} \\
\toprule
Objectives & R & SP & EO & OAE & PP & PE & IF & CSC \\
\midrule
\endhead
\midrule
\multicolumn{9}{r}{Continued on next page} \\
\midrule
\endfoot
\bottomrule
\endlastfoot
\midrule
R & -0.01126 & -1.11565 & -2.20794 & -3.13321 & -14.70363 & -2.7711 & -135.21178 & -92.24007 \\
R & -2.79987 & -1.16339 & -6.36855 & -8.17393 & -15.15077 & -6.31668 & -129.69359 & -88.99381 \\
R & -4.94815 & -6.16776 & -10.7125 & -4.08187 & -9.25954 & -1.1775 & -121.77361 & -84.09399 \\
R & -3.70322 & -8.16997 & -5.3824 & -1.1377 & -8.29254 & -11.93667 & -123.49949 & -85.90979 \\
R & -4.21773 & -3.5921 & -5.57 & -9.60919 & -5.29428 & -18.32441 & -119.43388 & -82.73461 \\
R & -4.93747 & -4.1899 & -12.03186 & -10.81699 & -10.46263 & -8.08611 & -122.18026 & -86.89458 \\
\midrule
SP & -35.27956 & -0.39189 & -2.46492 & -3.12463 & -4.18598 & -3.73639 & -54.66069 & -88.51997 \\
SP & -35.00607 & -0.34989 & -2.67969 & -2.48422 & -0.8995 & -4.30345 & -143.95361 & -53.18559 \\
\end{longtable}
\end{tiny}

\clearpage
\subsection{Multi-objective scenarios}

\begin{tiny}
\begin{longtable}[h!]{lllllllll}
\caption{The representative subset of job hiring policies, for different objectives and a history size of 1000 interactions. Results rounded to 5 decimals.} \label{table:jobR.SP.IF_b0_dHEOM_w1000} \\
\toprule
Objectives & R & SP & EO & OAE & PP & PE & IF & CSC \\
\midrule
\endfirsthead
\caption[]{The representative subset of job hiring policies, for different objectives and a history size of 1000 interactions. Results rounded to 5 decimals.} \\
\toprule
Objectives & R & SP & EO & OAE & PP & PE & IF & CSC \\
\midrule
\endhead
\midrule
\multicolumn{9}{r}{Continued on next page} \\
\midrule
\endfoot
\bottomrule
\endlastfoot
\midrule
R:SP:IF & -38.55409 & -0.10997 & -0.5396 & -2.12208 & -2.30021 & -0.88142 & -46.87076 & -16.77 \\
R:SP:IF & -28.93764 & -0.52849 & -0.90892 & -0.48084 & -0.42482 & -0.33712 & -50.89976 & -29.3225 \\
R:SP:IF & -49.04152 & -1.55755 & -7.47604 & -7.97693 & -4.52488 & -9.12457 & -38.13126 & -82.79517 \\
R:SP:IF & -30.49906 & -5.27728 & -2.65579 & -0.72134 & -0.36479 & -9.4025 & -46.6758 & -32.3614 \\
R:SP:IF & -45.04024 & -0.62987 & -7.31206 & -9.59434 & -20.2168 & -10.71303 & -45.68913 & -83.86449 \\
R:SP:IF & -36.21825 & -0.71658 & -7.4431 & -10.9845 & -14.68289 & -11.96016 & -43.73351 & -70.58782 \\
R:SP:IF & -40.15923 & -3.73662 & -4.83203 & -4.32967 & -9.29859 & -3.78734 & -31.73832 & -50.2256 \\
R:SP:IF & -26.38679 & -0.39738 & -2.28354 & -0.42092 & -1.65645 & -3.3596 & -49.40085 & -31.54713 \\
\end{longtable}
\end{tiny}

\begin{figure}[h!]
    \centering
    \includegraphics[width=1\linewidth]{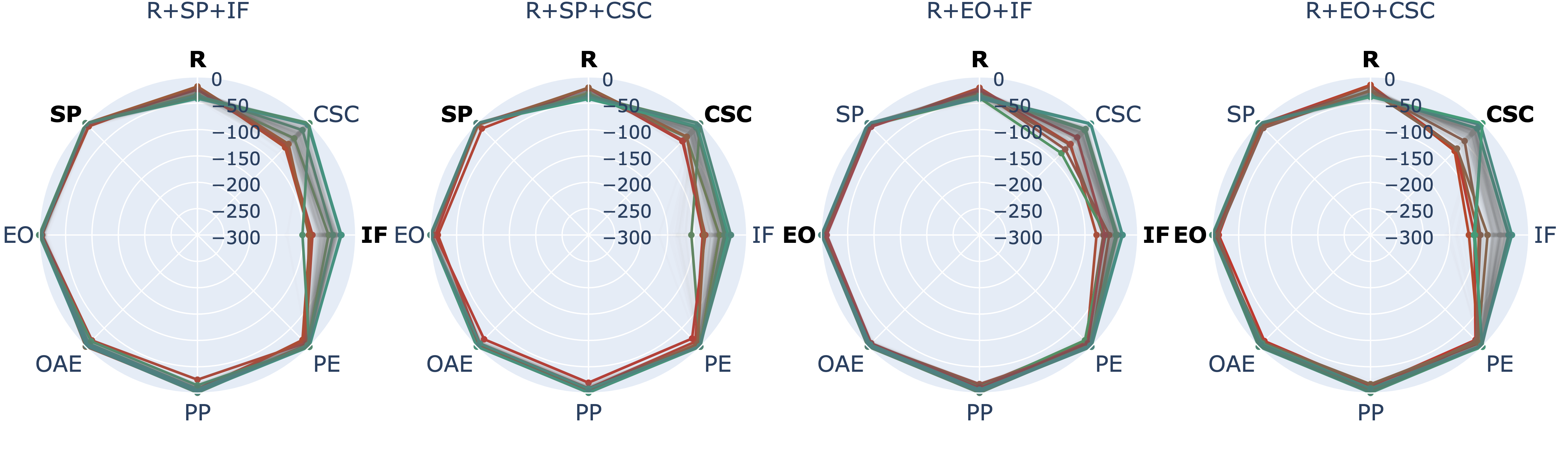}
    \caption{Representative set of learned job hiring policies when optimising the reward (R), a group fairness notion (SP or EO) and an individual fairness notion (IF or CSC), given a history size of 500 interactions.}
    \label{fig:job_multi500}
\end{figure}

\begin{tiny}
\begin{longtable}[h!]{lllllllll}
\caption{The representative subset of job hiring policies, for different objectives and a history size of 500 interactions. Results rounded to 5 decimals.} \label{table:jobR.SP.IF_b0_dHEOM_w500} \\
\toprule
Objectives & R & SP & EO & OAE & PP & PE & IF & CSC \\
\midrule
\endfirsthead
\caption[]{The representative subset of job hiring policies, for different objectives and a history size of 500 interactions. Results rounded to 5 decimals.} \\
\toprule
Objectives & R & SP & EO & OAE & PP & PE & IF & CSC \\
\midrule
\endhead
\midrule
\multicolumn{9}{r}{Continued on next page} \\
\midrule
\endfoot
\bottomrule
\endlastfoot
\midrule
R:SP:IF & -32.83704 & -0.54427 & -4.40342 & -0.52838 & -4.98103 & -5.45467 & -51.5118 & -40.5556 \\
R:SP:IF & -24.0008 & -0.82756 & -5.63892 & -17.22588 & -25.54344 & -10.3073 & -81.98038 & -60.20558 \\
R:SP:IF & -18.95935 & -8.00135 & -3.64442 & -5.54326 & -7.87233 & -16.24868 & -82.99237 & -63.79299 \\
R:SP:IF & -18.59573 & -8.64453 & -3.77416 & -1.1891 & -2.24963 & -18.46451 & -86.59753 & -65.063 \\
R:SP:IF & -18.21916 & -0.98482 & -4.39814 & -3.30643 & -2.45628 & -5.82905 & -88.94828 & -55.82098 \\
R:SP:IF & -35.11348 & -0.37711 & -3.39001 & -2.56979 & -0.59973 & -4.66196 & -43.23003 & -18.60721 \\
R:SP:IF & -36.66174 & -1.12502 & -2.52145 & -8.68367 & -7.61862 & -1.29402 & -27.12812 & -4.38524 \\
R:SP:IF & -40.35138 & -0.0395 & -0.01551 & -4.98885 & -5.03409 & -0.0 & -28.70599 & -4.046 \\
\end{longtable}
\end{tiny}

\subsubsection{Distance metrics}

\begin{tiny}
\begin{table}[h!]\tiny
\caption{The representative subset of job hiring policies, for R:SP and different distance metrics. Results rounded to 5 decimals.}
\label{table:jobR.SP_b0_dbraycurtis:HEOM:HMOM:braycurtis:HEOM:HMOM_w500}
\begin{tabular}{lllllllllllll}
\toprule
\multirow{2}{*}{Objectives} & R & SP & EO & PE & OAE & PP \\ & IF\textsubscript{braycurtis} & IF\textsubscript{HEOM} & IF\textsubscript{HMOM} & CSC\textsubscript{braycurtis} & CSC\textsubscript{HEOM} & CSC\textsubscript{HMOM} \\
\midrule
\midrule
\multirow{2}{*}{R:SP} & -23.92363 & -1.16114 & -2.95858 & -0.82576 & -2.92597 & -4.76207 \\ & -102.25606 & -65.56171 & -65.56171 & -37.79957 & -39.63509 & -39.63509 \\
\multirow{2}{*}{R:SP} & -33.20802 & -0.32851 & -4.19359 & -2.64123 & -0.67583 & -7.01809 \\ & -79.4223 & -47.22494 & -47.22494 & -27.9384 & -31.3192 & -31.3192 \\
\multirow{2}{*}{R:SP} & -40.484 & -0.2487 & -0.41582 & -7.76001 & -9.32796 & -1.00541 \\ & -69.99872 & -40.43677 & -40.43677 & -20.0372 & -17.6568 & -17.6568 \\
\multirow{2}{*}{R:SP} & -40.34898 & -0.25068 & -2.49031 & -4.27668 & -3.11852 & -3.85887 \\ & -70.79092 & -40.52792 & -40.52792 & -16.20071 & -20.38496 & -20.38496 \\
\multirow{2}{*}{R:SP} & -19.16215 & -3.4691 & -3.49375 & -8.07568 & -19.85128 & -5.95766 \\ & -138.88295 & -95.04927 & -95.04927 & -61.59741 & -65.21478 & -65.21478 \\
\multirow{2}{*}{R:SP} & -17.73706 & -1.37554 & -3.05147 & -6.48025 & -11.88248 & -4.06722 \\ & -139.51741 & -97.85275 & -97.85275 & -65.76253 & -66.15818 & -66.15818 \\
\multirow{2}{*}{R:SP} & -16.91397 & -4.01543 & -5.02638 & -3.03821 & -7.98812 & -3.19035 \\ & -128.04231 & -88.09111 & -88.09111 & -61.56036 & -62.15839 & -62.15839 \\
\multirow{2}{*}{R:SP} & -19.56787 & -4.4561 & -2.27113 & -1.17946 & -4.57868 & -9.35888 \\ & -130.17764 & -87.03747 & -87.03747 & -67.72505 & -73.99865 & -73.99865 \\
\multirow{2}{*}{R:SP} & -19.8224 & -1.27342 & -1.5569 & -5.12306 & -12.83635 & -4.83617 \\ & -132.0056 & -89.73672 & -89.73672 & -69.42762 & -73.99079 & -73.99079 \\
\multirow{2}{*}{R:SP} & -19.29517 & -1.0386 & -6.43582 & -10.28081 & -13.10043 & -10.05528 \\ & -130.60986 & -88.29499 & -88.29499 & -68.64424 & -60.50642 & -60.50642 \\
\bottomrule
\end{tabular}
\end{table}
\end{tiny}

\begin{figure}[h!]
    \centering
    \includegraphics[width=1\linewidth]{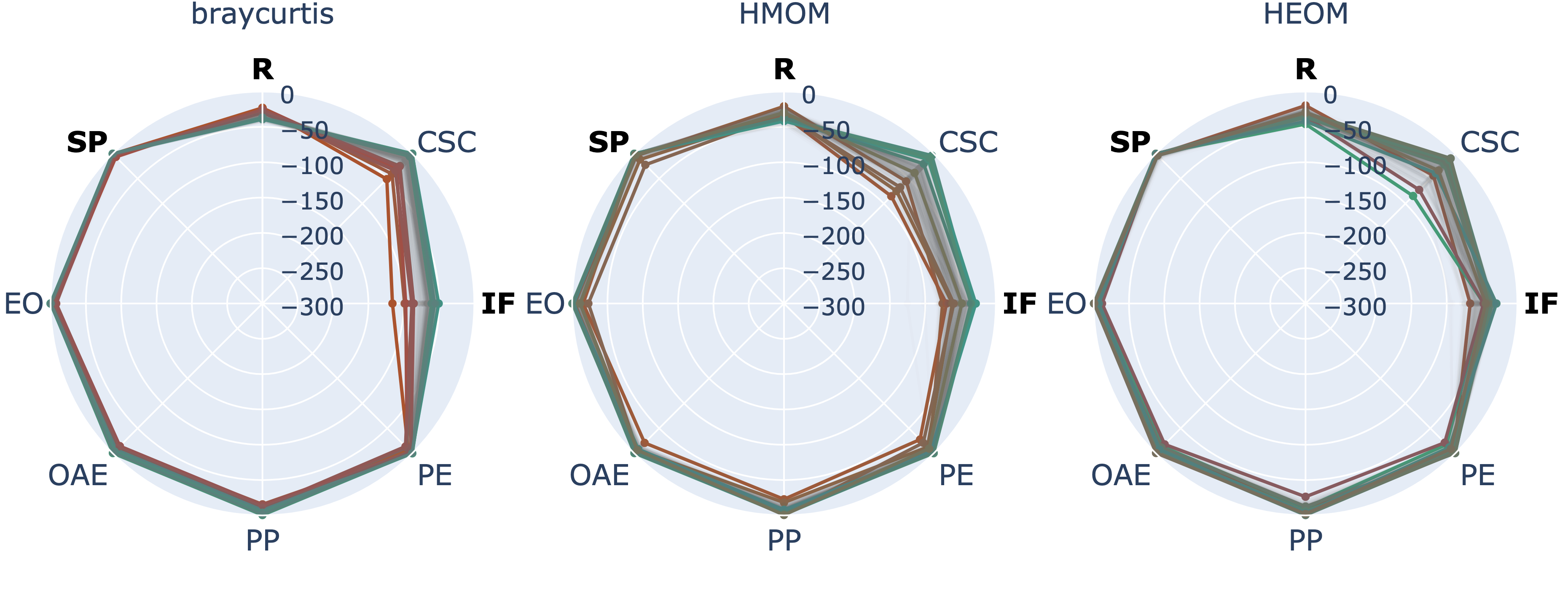}
    \caption{Representative set of learned job hiring policies when optimising the performance reward (R), statistical parity (SP) and individual fairness (IF). Showing results for individual fairness notions with different distance metrics.}
    \label{fig:job_distances}
\end{figure}

\begin{tiny}
\begin{longtable}[h!]{lllllllll}
\caption{The representative subset of job hiring policies, for different individual fairness distance metrics, when optimising for R-SP-IF. Results rounded to 5 decimals.} \label{table:jobR.SP.IF_b0_distances_w500} \\
\toprule
Distance & R & SP & EO & OAE & PP & PE & IF & CSC \\
\midrule
\endfirsthead
\caption[]{The representative subset of job hiring policies, for different individual fairness distance metrics, when optimising for R-SP-IF. Results rounded to 5 decimals.} \\
\toprule
Distance & R & SP & EO & OAE & PP & PE & IF & CSC \\
\midrule
\endhead
\midrule
\multicolumn{9}{r}{Continued on next page} \\
\midrule
\endfoot
\bottomrule
\endlastfoot
\midrule
braycurtis & -23.28194 & -0.89771 & -0.80807 & -10.01489 & -15.65259 & -1.91044 & -98.62922 & -40.6084 \\
braycurtis & -25.67471 & -4.99947 & -7.28372 & -14.44312 & -14.42495 & -2.61797 & -87.33759 & -25.00117 \\
braycurtis & -23.10118 & -6.47734 & -4.32894 & -5.29595 & -7.56851 & -10.43633 & -97.07291 & -31.51429 \\
braycurtis & -22.91382 & -1.07024 & -5.54453 & -4.33624 & -2.18834 & -8.30383 & -115.92265 & -50.97247 \\
braycurtis & -29.83329 & -0.55293 & -6.07759 & -12.30007 & -10.25855 & -11.58107 & -85.7994 & -24.68568 \\
braycurtis & -29.21212 & -0.59537 & -7.95363 & -13.05961 & -10.6939 & -13.61643 & -88.39043 & -33.922 \\
braycurtis & -38.89613 & -0.39958 & -0.0 & -6.79429 & -6.9787 & -0.86187 & -50.47314 & -0.40961 \\
braycurtis & -37.20602 & -0.27898 & -0.27629 & -0.47792 & -0.48309 & -0.59801 & -60.41361 & -5.16529 \\
\end{longtable}
\end{tiny}

\subsubsection{Historical bias}

\begin{figure}[h!]
    \centering
    \includegraphics[width=1\linewidth]{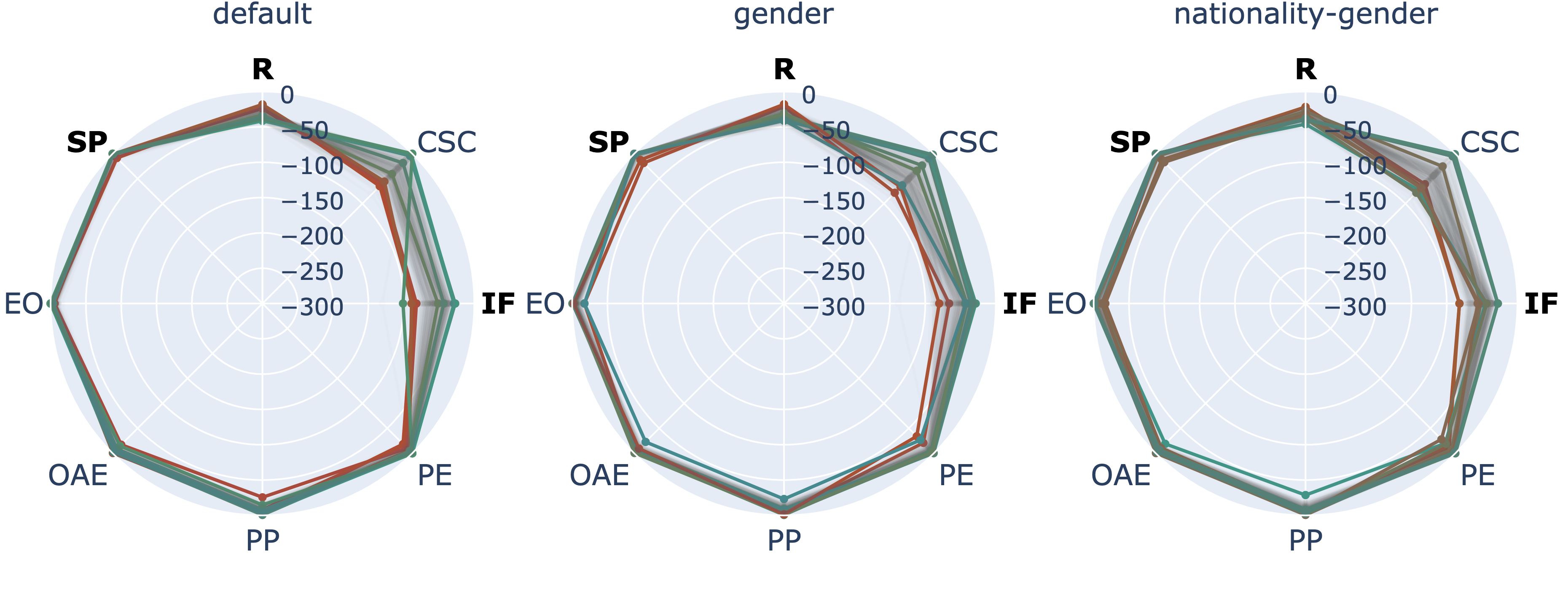}
    \caption{Representative set of learned policies when optimising the performance reward (R), statistical parity (SP) and individual fairness (IF) under different population distributions.}
    \label{fig:job_populations}
\end{figure}

\begin{tiny}
\begin{longtable}[h!]{lllllllll}
\caption{The representative subset of job hiring policies, for different populations, when optimising for R-SP-IF. Results rounded to 5 decimals.} \label{table:jobR.SP.IF_b0_dHEOM_w500_pop} \\
\toprule
Population & R & SP & EO & OAE & PP & PE & IF & CSC \\
\midrule
\endfirsthead
\caption[]{The representative subset of job hiring policies, for different populations, when optimising for R-SP-IF. Results rounded to 5 decimals.} \\
\toprule
Population & R & SP & EO & OAE & PP & PE & IF & CSC \\
\midrule
\endhead
\midrule
\multicolumn{9}{r}{Continued on next page} \\
\midrule
\endfoot
\bottomrule
\endlastfoot
\midrule
default & -32.83704 & -0.54427 & -4.40342 & -0.52838 & -4.98103 & -5.45467 & -51.5118 & -40.5556 \\
default & -24.0008 & -0.82756 & -5.63892 & -17.22588 & -25.54344 & -10.3073 & -81.98038 & -60.20558 \\
default & -18.95935 & -8.00135 & -3.64442 & -5.54326 & -7.87233 & -16.24868 & -82.99237 & -63.79299 \\
default & -18.59573 & -8.64453 & -3.77416 & -1.1891 & -2.24963 & -18.46451 & -86.59753 & -65.063 \\
default & -18.21916 & -0.98482 & -4.39814 & -3.30643 & -2.45628 & -5.82905 & -88.94828 & -55.82098 \\
default & -35.11348 & -0.37711 & -3.39001 & -2.56979 & -0.59973 & -4.66196 & -43.23003 & -18.60721 \\
default & -36.66174 & -1.12502 & -2.52145 & -8.68367 & -7.61862 & -1.29402 & -27.12812 & -4.38524 \\
default & -40.35138 & -0.0395 & -0.01551 & -4.98885 & -5.03409 & -0.0 & -28.70599 & -4.046 \\
\end{longtable}
\end{tiny}

\begin{figure}[H]\centering
  \begin{subfigure}[h!]{1\linewidth}\centering
    \includegraphics[width=\linewidth]{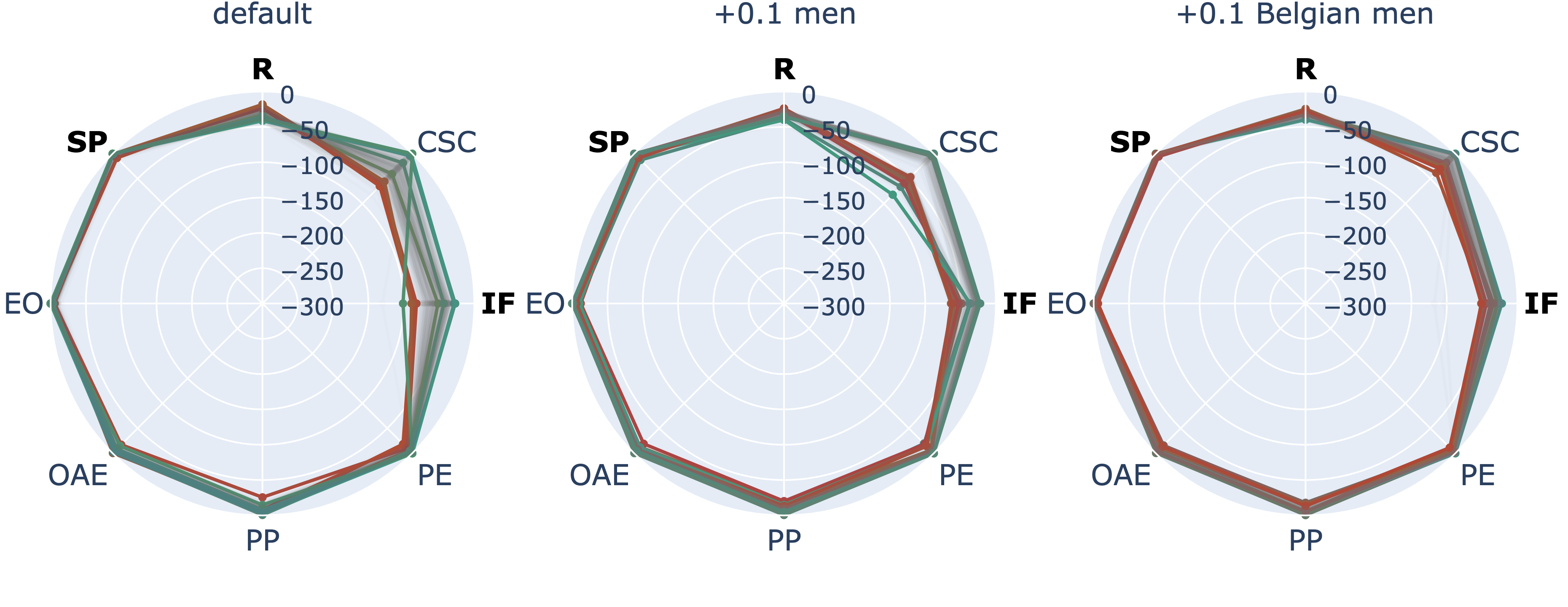}
    \caption{The default Belgian population}
    \label{fig:job_bias_default}
  \end{subfigure}
    \begin{subfigure}[h!]{1\linewidth}\centering
    \includegraphics[width=\linewidth]{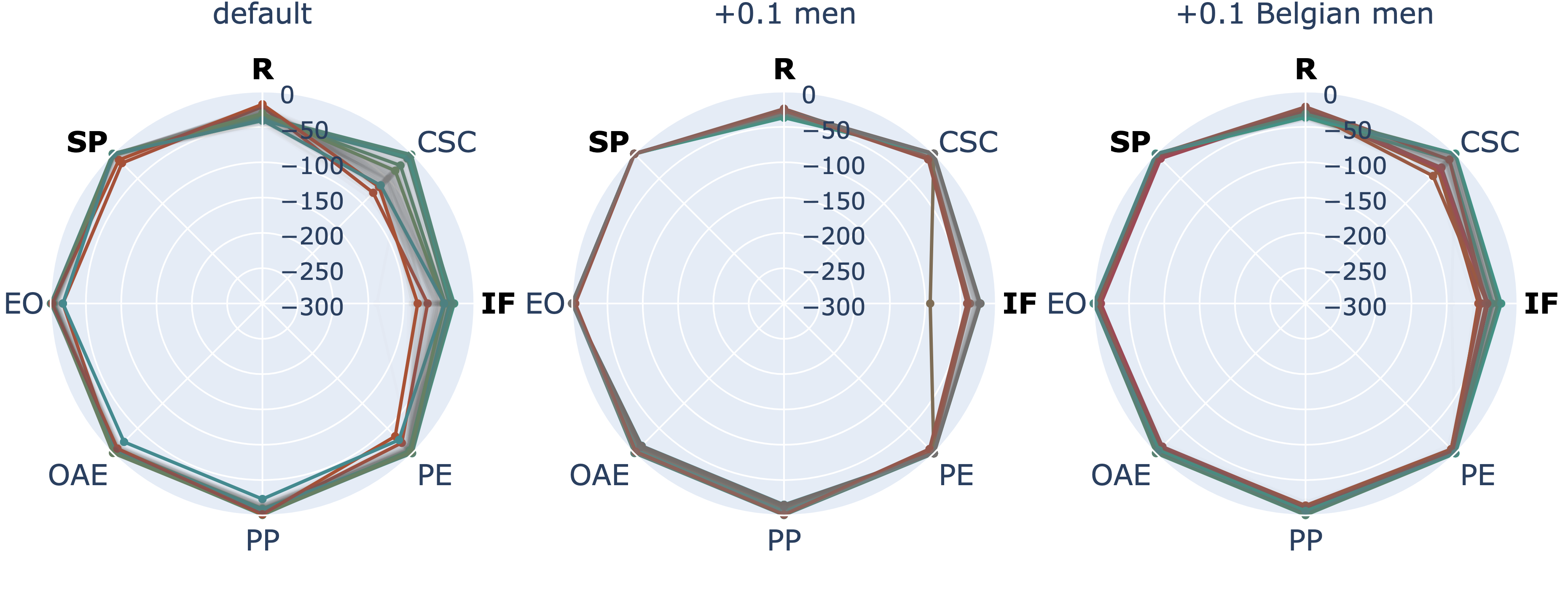}
    \caption{Belgian population, with a majority of 70\% male job applicants.}
    \label{fig:job_bias_gen}
  \end{subfigure}
      \begin{subfigure}[h!]{1\linewidth}\centering
    \includegraphics[width=\linewidth]{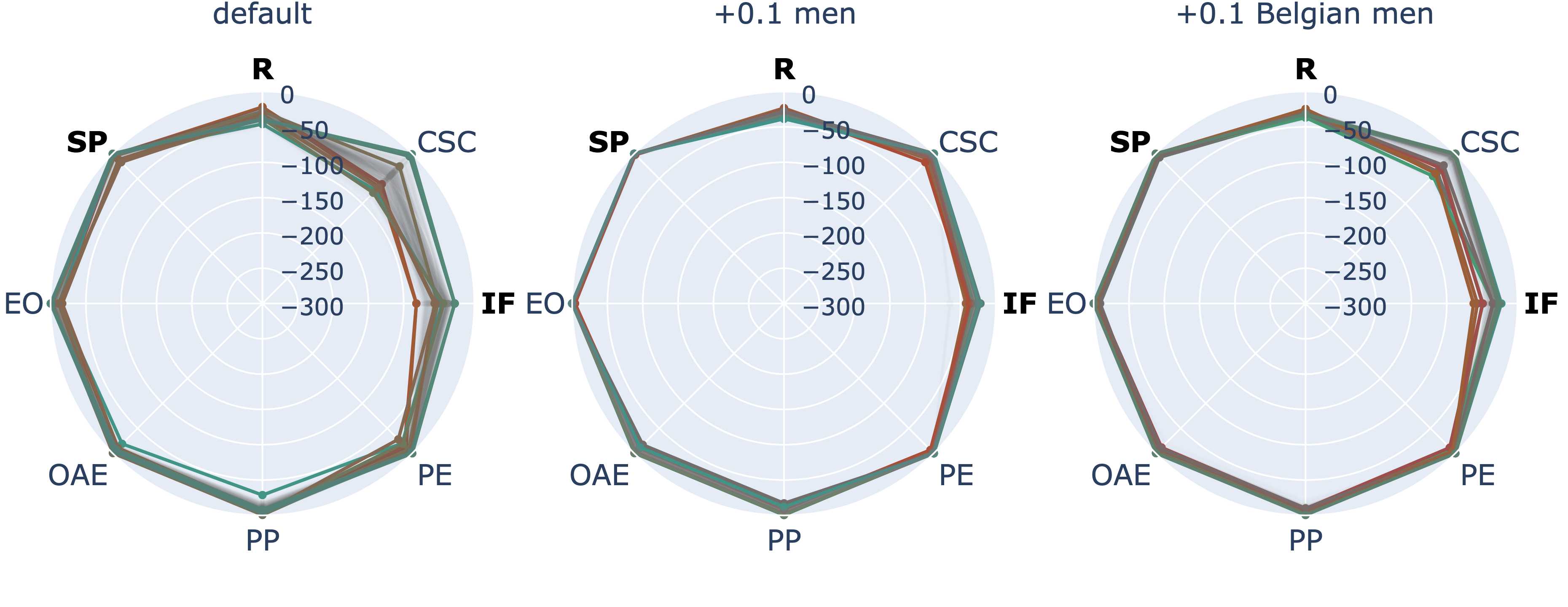}
    \caption{Belgian population, with a 5\% foreign women minority.}
    \label{fig:job_bias_nat_gen}
  \end{subfigure}
  \caption{Representative set of learned job hiring policies when optimising the performance reward (R), statistical parity (SP) and individual fairness (IF) under scenarios with different reward biases.}
    \label{figure:job_bias_all}
\end{figure}

\begin{tiny}
\begin{longtable}[h!]{lllllllll}
\caption{The representative subset of job hiring policies, for different reward biases in the default Belgian population, when optimising for R-SP-IF. Results rounded to 5 decimals.} \label{table:jobR.SP.IF_b012_dHEOM_w500} \\
\toprule
Bias & R & SP & EO & OAE & PP & PE & IF & CSC \\
\midrule
\endfirsthead
\caption[]{The representative subset of job hiring policies, for different reward biases in the default Belgian population, when optimising for R-SP-IF. Results rounded to 5 decimals.} \\
\toprule
Bias & R & SP & EO & OAE & PP & PE & IF & CSC \\
\midrule
\endhead
\midrule
\multicolumn{9}{r}{Continued on next page} \\
\midrule
\endfoot
\bottomrule
\endlastfoot
\midrule
default & -32.83704 & -0.54427 & -4.40342 & -0.52838 & -4.98103 & -5.45467 & -51.5118 & -40.5556 \\
default & -24.0008 & -0.82756 & -5.63892 & -17.22588 & -25.54344 & -10.3073 & -81.98038 & -60.20558 \\
default & -18.95935 & -8.00135 & -3.64442 & -5.54326 & -7.87233 & -16.24868 & -82.99237 & -63.79299 \\
default & -18.59573 & -8.64453 & -3.77416 & -1.1891 & -2.24963 & -18.46451 & -86.59753 & -65.063 \\
default & -18.21916 & -0.98482 & -4.39814 & -3.30643 & -2.45628 & -5.82905 & -88.94828 & -55.82098 \\
default & -35.11348 & -0.37711 & -3.39001 & -2.56979 & -0.59973 & -4.66196 & -43.23003 & -18.60721 \\
default & -36.66174 & -1.12502 & -2.52145 & -8.68367 & -7.61862 & -1.29402 & -27.12812 & -4.38524 \\
default & -40.35138 & -0.0395 & -0.01551 & -4.98885 & -5.03409 & -0.0 & -28.70599 & -4.046 \\
\end{longtable}
\end{tiny}

\begin{tiny}
\begin{longtable}[h!]{lllllllll}
\caption{The representative subset of job hiring policies, for different reward biases in the skewed gender population, when optimising for R-SP-IF. Results rounded to 5 decimals.} \label{table:jobR.SP.IF_b012_dHEOM_w500_gen} \\
\toprule
Bias & R & SP & EO & OAE & PP & PE & IF & CSC \\
\midrule
\endfirsthead
\caption[]{The representative subset of job hiring policies, for different reward biases in the skewed gender population, when optimising for R-SP-IF. Results rounded to 5 decimals.} \\
\toprule
Bias & R & SP & EO & OAE & PP & PE & IF & CSC \\
\midrule
\endhead
\midrule
\multicolumn{9}{r}{Continued on next page} \\
\midrule
\endfoot
\bottomrule
\endlastfoot
\midrule
default & -21.29839 & -18.77027 & -17.3662 & -3.20767 & -2.4 & -20.98858 & -66.14426 & -78.2583 \\
default & -36.90632 & -0.75636 & -1.36396 & -5.29821 & -5.014 & -0.32597 & -30.51519 & -3.11525 \\
default & -33.70183 & -0.3236 & -0.39671 & -8.01912 & -9.4209 & -0.6648 & -35.53955 & -9.72134 \\
default & -41.05521 & -0.15477 & -1.79464 & -6.67046 & -8.17939 & -2.3724 & -33.4979 & -2.9532 \\
default & -39.51364 & -1.87862 & -4.02255 & -3.15679 & -5.46843 & -1.5063 & -28.57055 & -7.85218 \\
default & -35.2882 & -2.57834 & -3.36746 & -2.63898 & -1.69343 & -1.39954 & -30.41273 & -2.00204 \\
default & -33.85126 & -0.31494 & -3.14944 & -2.42771 & -0.60196 & -5.78202 & -41.34256 & -23.6468 \\
default & -30.90425 & -2.55622 & -2.64348 & -0.68285 & -1.90576 & -2.34246 & -43.59032 & -34.15721 \\
\end{longtable}
\end{tiny}

\begin{tiny}
\begin{longtable}[h!]{lllllllll}
\caption{The representative subset of job hiring policies, for different reward biases in the skewed nationality-gender population, when optimising for R-SP-IF. Results rounded to 5 decimals.} \label{table:jobR.SP.IF_b012_dHEOM_nat_gen_w500} \\
\toprule
Bias & R & SP & EO & OAE & PP & PE & IF & CSC \\
\midrule
\endfirsthead
\caption[]{The representative subset of job hiring policies, for different reward biases in the skewed nationality-gender population, when optimising for R-SP-IF. Results rounded to 5 decimals.} \\
\toprule
Bias & R & SP & EO & OAE & PP & PE & IF & CSC \\
\midrule
\endhead
\midrule
\multicolumn{9}{r}{Continued on next page} \\
\midrule
\endfoot
\bottomrule
\endlastfoot
\midrule
default & -21.98471 & -2.60183 & -13.09889 & -10.99106 & -2.72631 & -15.48665 & -82.06024 & -60.9962 \\
default & -21.80321 & -4.37759 & -1.34972 & -5.38558 & -6.59804 & -12.36914 & -82.08882 & -66.3998 \\
default & -27.90065 & -4.50498 & -2.90482 & -0.44408 & -0.60262 & -7.36999 & -51.90261 & -25.58851 \\
default & -45.98682 & -0.8808 & -11.71366 & -18.98683 & -28.76968 & -20.14859 & -45.33498 & -73.31568 \\
default & -38.1039 & -15.00093 & -12.44673 & -0.98466 & -2.53822 & -19.0313 & -43.3643 & -78.19427 \\
default & -32.31672 & -12.0008 & -16.54558 & -10.51602 & -6.94122 & -3.43324 & -56.20597 & -68.59617 \\
default & -34.01614 & -17.15549 & -11.18996 & -1.15021 & -2.52682 & -27.95816 & -54.84369 & -68.46001 \\
default & -38.51683 & -0.17652 & -0.25418 & -4.99693 & -5.07205 & -0.0 & -27.85228 & -5.33803 \\
\end{longtable}
\end{tiny}

\subsubsection{History size}
\begin{tiny}
\begin{longtable}[h!]{rllllllll}
\caption{The representative subset of job hiring policies, for different window sizes, when optimising for R-SP-IF. Results rounded to 5 decimals.} \label{table:jobR.SP.IF_b0_dHEOM_windows} \\
\toprule
Window & R & SP & EO & OAE & PP & PE & IF & CSC \\
\midrule
\endfirsthead
\caption[]{The representative subset of job hiring policies, for different window sizes, when optimising for R-SP-IF. Results rounded to 5 decimals.} \\
\toprule
Window & R & SP & EO & OAE & PP & PE & IF & CSC \\
\midrule
\endhead
\midrule
\multicolumn{9}{r}{Continued on next page} \\
\midrule
\endfoot
\bottomrule
\endlastfoot
\midrule
100 & -30.81397 & -4.15814 & -6.20586 & -1.61295 & -3.76896 & -5.70434 & -44.09631 & -17.3763 \\
100 & -19.93607 & -12.65046 & -27.45694 & -20.78539 & -11.62838 & -15.71096 & -79.54784 & -62.42128 \\
100 & -35.96583 & -2.15421 & -1.76129 & -2.24708 & -2.12711 & -2.78107 & -26.31224 & -4.789 \\
100 & -34.71499 & -0.08117 & -0.11683 & -3.52936 & -3.47337 & -0.0 & -27.37421 & -0.05 \\
100 & -39.49371 & -4.2829 & -6.27967 & -9.11281 & -9.33388 & -1.44108 & -24.4035 & -8.4042 \\
100 & -40.09329 & -24.26039 & -23.83035 & -8.2128 & -16.28096 & -31.9161 & -35.0319 & -68.10789 \\
100 & -18.09346 & -10.97535 & -7.98396 & -12.00867 & -21.23272 & -40.91113 & -89.21693 & -68.01544 \\
100 & -15.26359 & -4.20476 & -14.06081 & -19.02084 & -27.0078 & -17.5522 & -93.89771 & -76.03551 \\
\end{longtable}
\end{tiny}

\subsubsection{Discounted history}

\begin{tiny}
\begin{longtable}{lllllllll}
\caption{The representative subset of job hiring policies, for different discounted histories under different discount factors, when optimising for R-SP-IF. Results rounded to 5 decimals.} \label{table:jobR.SP.IF_b0_dHEOM_windows_discount} \\
\toprule
Window & R & SP & EO & OAE & PP & PE & IF & CSC \\
\midrule
\endfirsthead
\caption[]{The representative subset of job hiring policies, for different discounted histories under different discount factors, when optimising for R-SP-IF. Results rounded to 5 decimals.} \\
\toprule
Window & R & SP & EO & OAE & PP & PE & IF & CSC \\
\midrule
\endhead
\midrule
\multicolumn{9}{r}{Continued on next page} \\
\midrule
\endfoot
\bottomrule
\endlastfoot
\midrule
discount 0.9 & -43.283 & -0.54309 & -0.52527 & -1.39242 & -4.41871 & -0.55774 & -31.70598 & -29.67035 \\
discount 0.9 & -41.17553 & -0.00022 & -0.04288 & -2.27227 & -3.96416 & -0.07603 & -40.11693 & -13.07397 \\
discount 0.9 & -38.46206 & -0.00781 & -0.01576 & -2.32957 & -4.0028 & -0.05496 & -34.0481 & -3.56623 \\
discount 0.9 & 7.50933 & -0.11656 & -0.35584 & -7.37448 & -42.11417 & -2.45181 & -292.46518 & -76.73218 \\
discount 0.9 & -24.40786 & -0.08567 & -0.92962 & -8.95482 & -43.05805 & -2.57908 & -311.85278 & -83.60093 \\
discount 0.9 & -12.19959 & -0.36258 & -0.00737 & -6.62913 & -42.2862 & -2.91355 & -290.42737 & -78.97668 \\
discount 0.9 & -26.72007 & -0.02322 & -0.00071 & -2.03873 & -5.09575 & -0.03909 & -62.7697 & -39.88792 \\
discount 0.9 & -33.13877 & -0.0316 & -0.05444 & -1.32909 & -6.28756 & -0.00067 & -88.38697 & -58.66058 \\
\end{longtable}
\end{tiny}

\begin{tiny}
\begin{longtable}{lllllllll}
\caption{The representative subset of job hiring policies, for different discounted histories under different discount thresholds, when optimising for R-SP-IF. Results rounded to 5 decimals.} \label{table:jobR.SP.IF_b0_dHEOM_windows_threshold} \\
\toprule
Window & R & SP & EO & OAE & PP & PE & IF & CSC \\
\midrule
\endfirsthead
\caption[]{The representative subset of job hiring policies, for different discounted histories under different discount thresholds, when optimising for R-SP-IF. Results rounded to 5 decimals.} \\
\toprule
Window & R & SP & EO & OAE & PP & PE & IF & CSC \\
\midrule
\endhead
\midrule
\multicolumn{9}{r}{Continued on next page} \\
\midrule
\endfoot
\bottomrule
\endlastfoot
\midrule
threshold 0.001 & -36.6166 & -0.16751 & -0.27415 & -1.2356 & -4.58649 & -0.00491 & -47.48212 & -27.13823 \\
threshold 0.001 & -36.64317 & -0.2505 & -0.30631 & -2.68412 & -4.3848 & -0.16195 & -36.30157 & -5.00945 \\
threshold 0.001 & -34.74876 & -0.00026 & -0.03738 & -2.65028 & -4.3315 & -0.07137 & -45.52317 & -24.49959 \\
threshold 0.001 & -41.54522 & -0.13647 & -0.00195 & -2.4171 & -5.07481 & -0.45852 & -32.80744 & -35.77768 \\
threshold 0.001 & -37.28486 & -0.00026 & -0.04252 & -2.62467 & -4.11202 & -0.07654 & -40.83262 & -12.41377 \\
threshold 0.001 & -41.33636 & -0.24664 & -0.1417 & -0.07698 & -4.62396 & -0.35033 & -49.69037 & -95.332 \\
threshold 0.001 & -16.68753 & -0.2586 & -0.1824 & -2.55558 & -6.50897 & -0.39547 & -112.89635 & -82.85616 \\
threshold 0.001 & -15.58067 & -0.26608 & -0.18952 & -2.67958 & -6.82191 & -0.40357 & -114.42494 & -86.16483 \\
\end{longtable}
\end{tiny}


\begin{figure}
    \centering
    \includegraphics[width=1\linewidth]{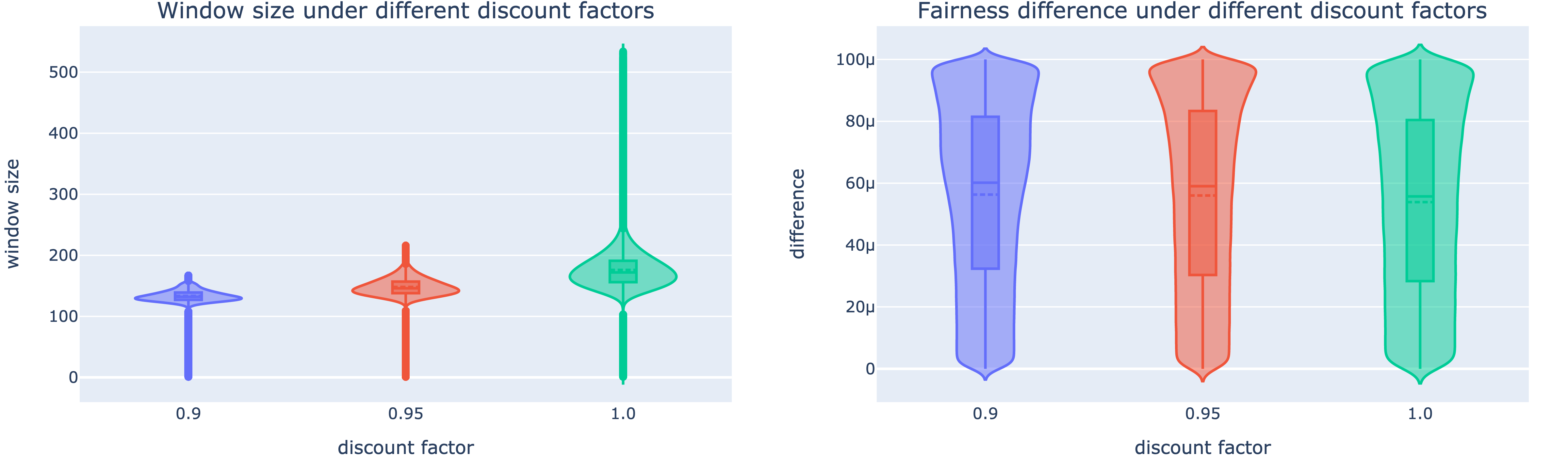}
    \caption{History window size and fairness difference of job hiring policies when optimising the reward (R), statistical parity (SP) and individual fairness (IF). Showing results for histories with different discount factors.}
    \label{fig:job_discount}
\end{figure}

\begin{figure}
    \centering
    \includegraphics[width=1\linewidth]{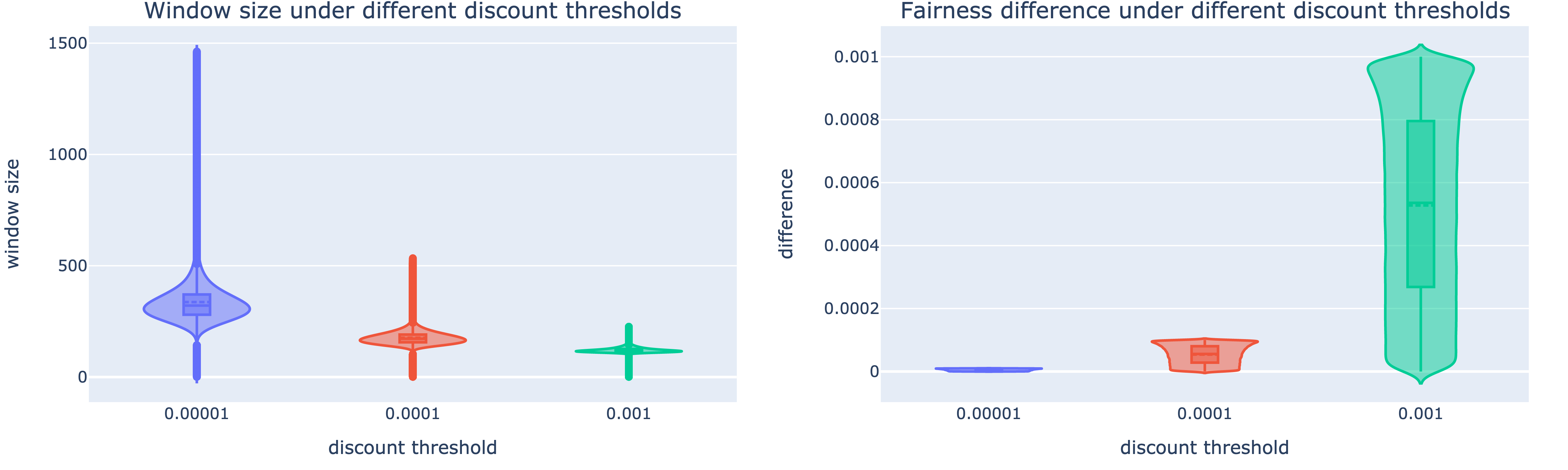}
    \caption{History window size and fairness difference of job hiring policies when optimising the reward (R), statistical parity (SP) and individual fairness (IF). Showing results for histories with different discount thresholds.}
    \label{fig:job_threshold}
\end{figure}

\section{Fraud detection}
\label{appendix:fraud_results}

\subsection{Single-objective scenarios}
\begin{tiny}
\begin{longtable}[h!]{lllllllll}
\caption{The representative subset of fraud detection policies, for different objectives. Results rounded to 5 decimals.} \label{table:fraud_R-SP-IF-EO-PE-PP-OAE-CSC_b0_dHEOM_w1000} \\
\toprule
Objectives & R & SP & EO & OAE & PP & PE & IF & CSC \\
\midrule
\endfirsthead
\caption[]{The representative subset of fraud detection policies, for different objective. Results rounded to 5 decimals.} \\
\toprule
Objectives & R & SP & EO & OAE & PP & PE & IF & CSC \\
\midrule
\endhead
\midrule
\multicolumn{9}{r}{Continued on next page} \\
\midrule
\endfoot
\bottomrule
\endlastfoot
\midrule
R & -178.0 & -14.06121 & -143.11859 & -10.53394 & -24.18176 & -9.89698 & -144.79916 & -497.84805 \\
R & -176.0 & -25.56806 & -42.25323 & -16.5029 & -228.35214 & -29.27623 & -145.10907 & -499.35535 \\
R & -212.0 & -40.12577 & -132.34218 & -26.45457 & -69.94877 & -28.49761 & -159.3093 & -489.89459 \\
R & -197.0 & -63.23713 & -128.12108 & -47.98624 & -29.25284 & -59.87572 & -150.52829 & -503.72232 \\
R & -203.0 & -52.8538 & -85.32261 & -59.18287 & -49.25005 & -60.3217 & -149.42546 & -504.31689 \\
\midrule
SP & -364.0 & -7.49004 & -150.20654 & -51.84356 & -23.83959 & -28.68732 & -131.37476 & -437.50308 \\
SP & -326.0 & -7.19406 & -54.82491 & -21.21576 & -299.2641 & -19.54623 & -144.55959 & -490.85565 \\
SP & -307.0 & -8.40817 & -95.71431 & -28.67083 & -141.32486 & -17.75563 & -143.99983 & -472.9917 \\
\end{longtable}
\end{tiny}

\clearpage
\subsection{Multi-objective scenarios}

\begin{tiny}
\begin{longtable}[h!]{lllllllll}
\caption{The representative subset of fraud detection policies, for different objectives and a history size of 1000 interactions. Results rounded to 5 decimals.} \label{table:fraudR.SP.IF_b0_dHEOM_w1000} \\
\toprule
Objectives & R & SP & EO & OAE & PP & PE & IF & CSC \\
\midrule
\endfirsthead
\caption[]{The representative subset of fraud detection policies, for different objectives and a history size of 1000 interactions. Results rounded to 5 decimals.} \\
\toprule
Objectives & R & SP & EO & OAE & PP & PE & IF & CSC \\
\midrule
\endhead
\midrule
\multicolumn{9}{r}{Continued on next page} \\
\midrule
\endfoot
\bottomrule
\endlastfoot
\midrule
R:SP:IF & -145.0 & -109.44725 & -222.07727 & -98.53996 & -56.13454 & -107.27169 & -136.86882 & -495.10754 \\
R:SP:IF & -235.0 & -10.67722 & -57.79453 & -11.44176 & -65.18379 & -7.54082 & -129.64458 & -430.83289 \\
R:SP:IF & -207.0 & -9.04047 & -39.77442 & -6.35331 & -95.52411 & -9.23981 & -136.89569 & -472.24136 \\
R:SP:IF & -327.0 & -12.9746 & -69.71426 & -19.64719 & -93.94595 & -17.50805 & -108.55624 & -385.42993 \\
R:SP:IF & -171.0 & -19.60403 & -57.70545 & -12.18851 & -9.9124 & -17.07203 & -142.25101 & -496.66132 \\
R:SP:IF & -82.0 & -25.56806 & -42.25323 & -16.5029 & -228.35214 & -29.27623 & -145.10907 & -499.35535 \\
R:SP:IF & -157.0 & -14.67714 & -196.19548 & -124.00999 & -146.45078 & -28.27451 & -711.88477 & -334.62921 \\
R:SP:IF & -271.0 & -18.57964 & -18.631 & -16.15601 & -160.35503 & -21.72479 & -121.3085 & -418.12573 \\
\end{longtable}
\end{tiny}

\begin{figure}[h!]
    \centering
    \includegraphics[width=1\linewidth]{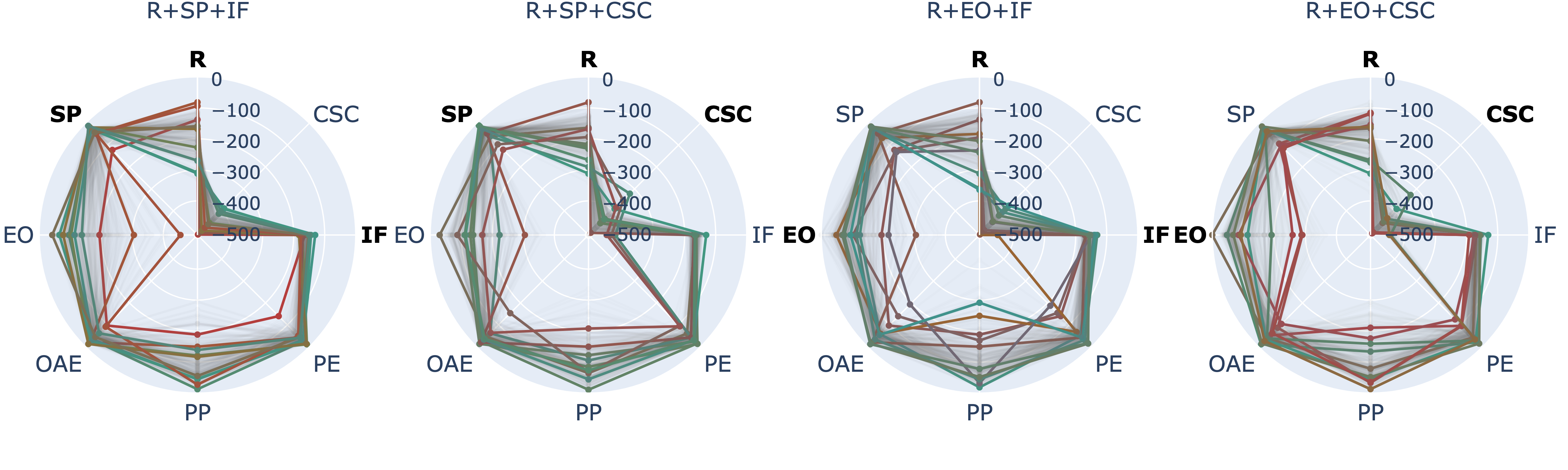}
    \caption{Representative set of learned fraud detection policies when optimising the reward (R), a group fairness notion (SP or EO) and an individual fairness notion (IF or CSC) given a history size of 500 interactions.}
    \label{fig:fraud_multi500}
\end{figure}

\begin{tiny}
\begin{longtable}[h!]{lllllllll}
\caption{The representative subset of fraud detection policies, for different objectives and a history size of 500 interactions. Results rounded to 5 decimals.} \label{table:fraudR.SP.IF_b0_dHEOM_w500} \\
\toprule
Objectives & R & SP & EO & OAE & PP & PE & IF & CSC \\
\midrule
\endfirsthead
\caption[]{The representative subset of fraud detection policies, for different objectives and a history size of 500 interactions. Results rounded to 5 decimals.} \\
\toprule
Objectives & R & SP & EO & OAE & PP & PE & IF & CSC \\
\midrule
\endhead
\midrule
\multicolumn{9}{r}{Continued on next page} \\
\midrule
\endfoot
\bottomrule
\endlastfoot
\midrule
R:SP:IF & -138.0 & -122.14719 & -194.15494 & -97.09627 & -188.77664 & -139.86931 & -163.9416 & -507.65598 \\
R:SP:IF & -270.0 & -19.76074 & -95.76708 & -31.97537 & -12.54168 & -26.88852 & -151.25174 & -413.00012 \\
R:SP:IF & -228.0 & -15.55026 & -102.89066 & -16.21072 & -120.96671 & -11.33415 & -171.87035 & -455.09711 \\
R:SP:IF & -82.0 & -47.37895 & -305.065 & -20.67384 & -149.64117 & -42.81386 & -170.89844 & -499.71588 \\
R:SP:IF & -312.0 & -25.88358 & -113.83823 & -33.29543 & -44.36665 & -30.82785 & -130.72705 & -391.14267 \\
R:SP:IF & -309.0 & -27.44427 & -64.93825 & -27.12269 & -46.05789 & -27.60119 & -144.33105 & -388.65131 \\
R:SP:IF & -168.0 & -19.42361 & -77.30825 & -12.81878 & -116.77959 & -14.68172 & -182.69884 & -490.1467 \\
R:SP:IF & -94.0 & -47.17715 & -455.11295 & -91.46596 & -27.55221 & -50.7003 & -176.42862 & -476.20084 \\
\end{longtable}
\end{tiny}

\subsubsection{Distance metrics}
\begin{tiny}
\begin{table}[h!]\tiny
\caption{The representative subset of fraud detection policies, for different objectives, for R:SP and different distance metrics. Results rounded to 5 decimals.}
\label{table:fraudR.SP_b0_dbraycurtis:HEOM:HMOM:braycurtis:HEOM:HMOM_w500}
\begin{tabular}{lllllllllllll}
\toprule
\multirow{2}{*}{Objectives} & R & SP & EO & PE & OAE & PP \\ & IF\textsubscript{braycurtis} & IF\textsubscript{HEOM} & IF\textsubscript{HMOM} & CSC\textsubscript{braycurtis} & CSC\textsubscript{HEOM} & CSC\textsubscript{HMOM} \\
\midrule
\midrule
\multirow{2}{*}{R:SP} & -117.0 & -60.08014 & -108.80129 & -80.4536 & -287.0166 & -83.83255 \\ & -680.77411 & -147.88969 & -147.88969 & -494.36685 & -500.95273 & -500.95273 \\
\multirow{2}{*}{R:SP} & -222.0 & -13.98716 & -254.371 & -31.96294 & -79.67577 & -19.688 \\ & -562.82581 & -124.80939 & -124.80939 & -419.53494 & -415.62103 & -415.62103 \\
\multirow{2}{*}{R:SP} & -111.0 & -88.57816 & -188.54306 & -112.3479 & -86.84702 & -109.92289 \\ & -688.49493 & -140.68575 & -140.68575 & -506.20004 & -507.92996 & -507.92996 \\
\bottomrule
\end{tabular}
\end{table}
\end{tiny}

\begin{figure}[h!]
    \centering
    \includegraphics[width=1\linewidth]{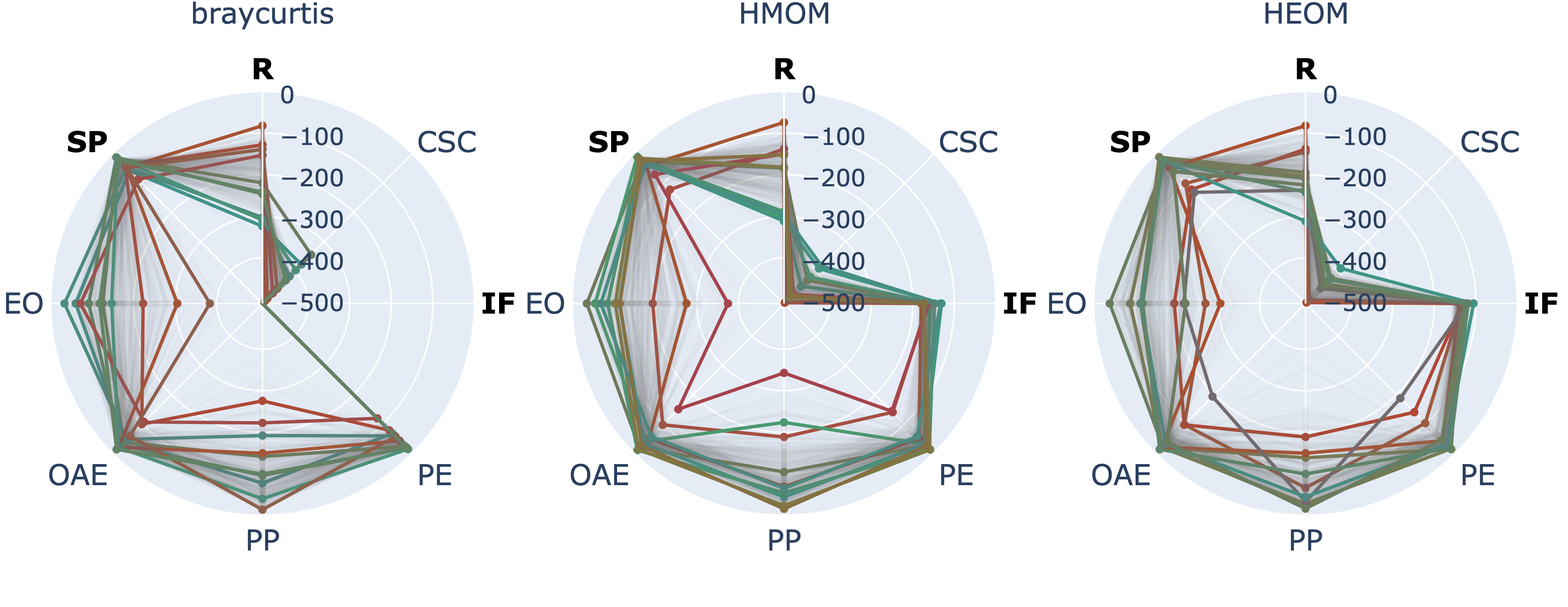}
    \caption{Representative set of learned fraud detection policies when optimising the performance reward (R), statistical parity (SP) and individual fairness (IF). Showing results for individual fairness notions with different distance metrics.}
    \label{fig:fraud_distances}
\end{figure}

\begin{tiny}
\begin{longtable}[h!]{lllllllll}
\caption{The representative subset of fraud detection policies, for different individual fairness distance metrics, when optimising for R-SP-IF. Results rounded to 5 decimals.} \label{table:fraudR.SP.IF_b0_distances_w500} \\
\toprule
Distance & R & SP & EO & OAE & PP & PE & IF & CSC \\
\midrule
\endfirsthead
\caption[]{The representative subset of fraud detection policies, for different individual fairness distance metrics, when optimising for R-SP-IF. Results rounded to 5 decimals.} \\
\toprule
Distance & R & SP & EO & OAE & PP & PE & IF & CSC \\
\midrule
\endhead
\midrule
\multicolumn{9}{r}{Continued on next page} \\
\midrule
\endfoot
\bottomrule
\endlastfoot
\midrule
braycurtis & -324.0 & -57.43885 & -60.46151 & -42.45512 & -77.42668 & -67.27528 & -510.58569 & -377.16336 \\
braycurtis & -220.0 & -15.61738 & -125.77898 & -37.13266 & -141.44695 & -23.63807 & -725.56421 & -344.15695 \\
braycurtis & -128.0 & -42.36214 & -222.71947 & -99.33068 & -275.60986 & -78.12498 & -689.01746 & -493.15265 \\
braycurtis & -303.0 & -46.3716 & -33.38168 & -47.60442 & -192.04443 & -59.4381 & -548.69971 & -417.85205 \\
braycurtis & -246.0 & -12.60961 & -92.10497 & -27.9348 & -96.02985 & -18.63959 & -584.56213 & -430.16708 \\
braycurtis & -82.0 & -47.37895 & -305.065 & -20.67384 & -149.64117 & -42.81386 & -688.71338 & -501.73135 \\
braycurtis & -153.0 & -87.44024 & -71.77389 & -106.87981 & -222.54231 & -118.94775 & -638.48425 & -474.59418 \\
braycurtis & -309.0 & -17.6025 & -147.12215 & -20.49753 & -39.84829 & -14.47666 & -549.8299 & -396.54648 \\
\end{longtable}
\end{tiny}

\subsubsection{Historical bias}

\begin{figure}[h!]
    \centering
    \includegraphics[width=1\linewidth]{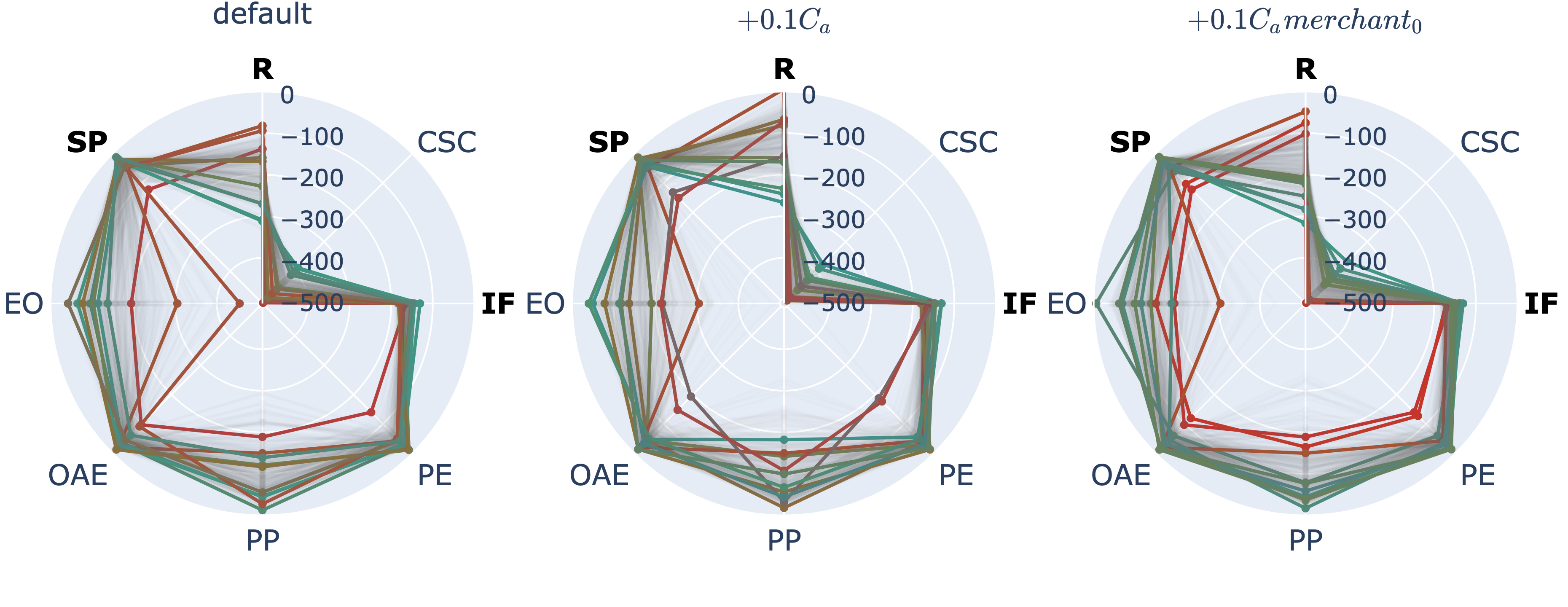}
    \caption{Representative set of learned fraud detection policies when optimising the performance reward (R), statistical parity (SP) and individual fairness (IF). Showing results under different reward biases.}
    \label{fig:fraud_bias}
\end{figure}

\begin{tiny}
\begin{longtable}[h!]{lllllllll}
\caption{The representative subset of fraud detection policies, for different reward biases, when optimising for R-SP-IF. Results rounded to 5 decimals.} \label{table:fraudR.SP.IF_b012_dHEOM_w500} \\
\toprule
Bias & R & SP & EO & OAE & PP & PE & IF & CSC \\
\midrule
\endfirsthead
\caption[]{The representative subset of fraud detection policies, for different reward biases, when optimising for R-SP-IF. Results rounded to 5 decimals.} \\
\toprule
Bias & R & SP & EO & OAE & PP & PE & IF & CSC \\
\midrule
\endhead
\midrule
\multicolumn{9}{r}{Continued on next page} \\
\midrule
\endfoot
\bottomrule
\endlastfoot
\midrule
default & -138.0 & -122.14719 & -194.15494 & -97.09627 & -188.77664 & -139.86931 & -163.9416 & -507.65598 \\
default & -270.0 & -19.76074 & -95.76708 & -31.97537 & -12.54168 & -26.88852 & -151.25174 & -413.00012 \\
default & -228.0 & -15.55026 & -102.89066 & -16.21072 & -120.96671 & -11.33415 & -171.87035 & -455.09711 \\
default & -82.0 & -47.37895 & -305.065 & -20.67384 & -149.64117 & -42.81386 & -170.89844 & -499.71588 \\
default & -312.0 & -25.88358 & -113.83823 & -33.29543 & -44.36665 & -30.82785 & -130.72705 & -391.14267 \\
default & -309.0 & -27.44427 & -64.93825 & -27.12269 & -46.05789 & -27.60119 & -144.33105 & -388.65131 \\
default & -168.0 & -19.42361 & -77.30825 & -12.81878 & -116.77959 & -14.68172 & -182.69884 & -490.1467 \\
default & -94.0 & -47.17715 & -455.11295 & -91.46596 & -27.55221 & -50.7003 & -176.42862 & -476.20084 \\
\end{longtable}
\end{tiny}

\subsubsection{History size}
\begin{tiny}
\begin{longtable}[h!]{rllllllll}
\caption{The representative subset of fraud detection policies, for different window sizes, when optimising for R-SP-IF. Results rounded to 5 decimals.} \label{table:fraudR.SP.IF_b0_dHEOM_windows} \\
\toprule
Window & R & SP & EO & OAE & PP & PE & IF & CSC \\
\midrule
\endfirsthead
\caption[]{The representative subset of fraud detection policies, for different window sizes, when optimising for R-SP-IF. Results rounded to 5 decimals.} \\
\toprule
Window & R & SP & EO & OAE & PP & PE & IF & CSC \\
\midrule
\endhead
\midrule
\multicolumn{9}{r}{Continued on next page} \\
\midrule
\endfoot
\bottomrule
\endlastfoot
\midrule
100 & -134.0 & -110.82684 & -20.76191 & -110.18678 & -21.18071 & -110.91719 & -116.51359 & -456.66241 \\
100 & -160.0 & -76.76062 & -243.49571 & -55.15492 & -96.82788 & -67.41191 & -182.88638 & -443.27899 \\
100 & -268.0 & -44.33103 & -244.17125 & -102.72358 & -153.30838 & -83.0055 & -119.28946 & -427.96109 \\
100 & -121.0 & -175.88464 & -360.93869 & -196.7216 & -138.23665 & -200.94218 & -131.44482 & -505.57571 \\
100 & -200.0 & -90.84322 & 0.0 & -93.2466 & -53.33629 & -92.73228 & -117.6415 & -444.74252 \\
100 & -206.0 & -215.88974 & -296.20059 & -272.55972 & -195.57675 & -289.63034 & -113.0688 & -446.12732 \\
100 & -81.0 & -89.33797 & -54.09002 & -76.72498 & -79.7075 & -83.5561 & -129.91655 & -506.85449 \\
100 & -179.0 & -222.35521 & -115.96928 & -220.49361 & -182.8539 & -247.05974 & -109.46583 & -457.64645 \\
\end{longtable}
\end{tiny}

\subsubsection{Discounted history}

\begin{tiny}
\begin{longtable}[h!]{lllllllll}
\caption{The representative subset of fraud detection policies, for different discounted histories under different discount factors, when optimising for R-SP-IF. Results rounded to 5 decimals.} \label{table:fraudR.SP.IF_b0_dHEOM_windows_discount} \\
\toprule
Window & R & SP & EO & OAE & PP & PE & IF & CSC \\
\midrule
\endfirsthead
\caption[]{The representative subset of fraud detection policies, for different discounted histories under different discount factors, when optimising for R-SP-IF. Results rounded to 5 decimals.} \\
\toprule
Window & R & SP & EO & OAE & PP & PE & IF & CSC \\
\midrule
\endhead
\midrule
\multicolumn{9}{r}{Continued on next page} \\
\midrule
\endfoot
\bottomrule
\endlastfoot
\midrule
discount 0.9 & -125.0 & -40.06539 & -75.29479 & -18.62443 & -31.32059 & -33.86568 & -158.89175 & -500.23251 \\
discount 0.9 & -260.0 & -14.92101 & -13.58911 & -9.69973 & -53.28425 & -14.26239 & -127.90926 & -405.16141 \\
discount 0.9 & -331.0 & -14.01515 & -15.06201 & -20.10114 & -196.5433 & -9.65232 & -123.93516 & -422.39969 \\
discount 0.9 & -165.0 & -6.07151 & -24.86858 & -0.32446 & -107.8061 & -3.59344 & -149.22353 & -473.37967 \\
discount 0.9 & -82.0 & -13.08342 & -8.44657 & -10.66294 & -171.99104 & -14.82321 & -153.31599 & -499.83545 \\
discount 0.9 & -302.0 & -1.8196 & -1.01586 & -63.63331 & -84.1 & -2.15443 & -126.83976 & -397.35928 \\
discount 0.9 & -191.0 & -0.14776 & -0.13192 & -84.45712 & -99.71792 & -0.13409 & -154.66788 & -477.16464 \\
discount 0.9 & -312.0 & -10.20967 & -4.30884 & -10.24784 & -79.90668 & -12.22036 & -116.29398 & -384.94092 \\
\end{longtable}
\end{tiny}

\begin{tiny}
\begin{longtable}[h!]{lllllllll}
\caption{The representative subset of fraud detection policies, for different discounted histories under different discount thresholds, when optimising for R-SP-IF. Results rounded to 5 decimals.} \label{table:fraudR.SP.IF_b0_dHEOM_windows_threshold} \\
\toprule
Window & R & SP & EO & OAE & PP & PE & IF & CSC \\
\midrule
\endfirsthead
\caption[]{The representative subset of fraud detection policies, for different discounted histories under different discount thresholds, when optimising for R-SP-IF. Results rounded to 5 decimals.} \\
\toprule
Window & R & SP & EO & OAE & PP & PE & IF & CSC \\
\midrule
\endhead
\midrule
\multicolumn{9}{r}{Continued on next page} \\
\midrule
\endfoot
\bottomrule
\endlastfoot
\midrule
threshold 0.001 & -208.0 & -0.00958 & -5.68211 & -58.60069 & -78.2494 & -3.40947 & -144.32179 & -430.00775 \\
threshold 0.001 & -260.0 & -14.92101 & -13.58911 & -9.69973 & -53.28425 & -14.26239 & -127.95671 & -403.75235 \\
threshold 0.001 & -142.0 & -9.66441 & -99.40194 & -12.96831 & -125.17337 & -9.10954 & -147.25964 & -463.17862 \\
threshold 0.001 & -186.0 & -0.6262 & -2.83293 & -82.40456 & -110.20138 & -0.48523 & -235.49573 & -470.14545 \\
threshold 0.001 & -176.0 & -2.16789 & -0.14738 & -63.58099 & -98.07688 & -2.86416 & -134.54185 & -451.49033 \\
threshold 0.001 & -147.0 & -49.8673 & -26.08375 & -27.96697 & -88.52016 & -49.70921 & -146.57738 & -471.30225 \\
threshold 0.001 & -82.0 & -13.08342 & -8.44657 & -10.66294 & -171.99104 & -14.82321 & -152.30988 & -500.328 \\
threshold 0.001 & -312.0 & -10.20967 & -4.30884 & -10.24784 & -79.90668 & -12.22036 & -115.5029 & -388.1792 \\
\end{longtable}
\end{tiny}

\begin{figure}[h!]
    \centering
    \includegraphics[width=1\linewidth]{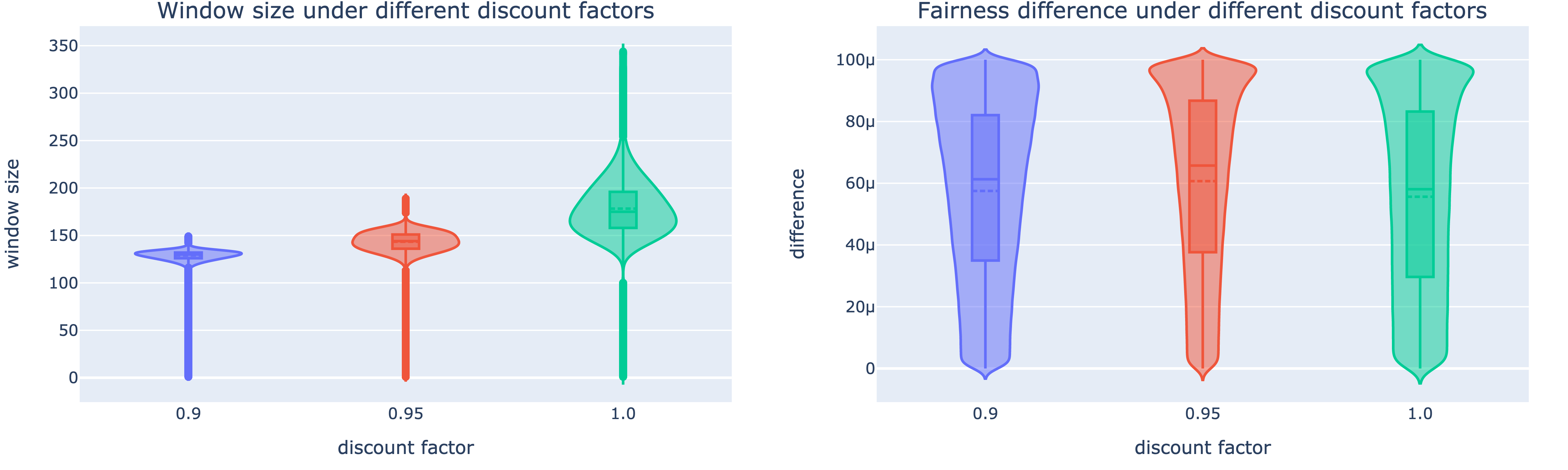}
    \caption{History window size and fairness difference of fraud detection policies when optimising the reward (R), statistical parity (SP) and individual fairness (IF). Showing results for histories with different discount factors.}
    \label{fig:fraud_discount}
\end{figure}

\begin{figure}[h!]
    \centering
    \includegraphics[width=1\linewidth]{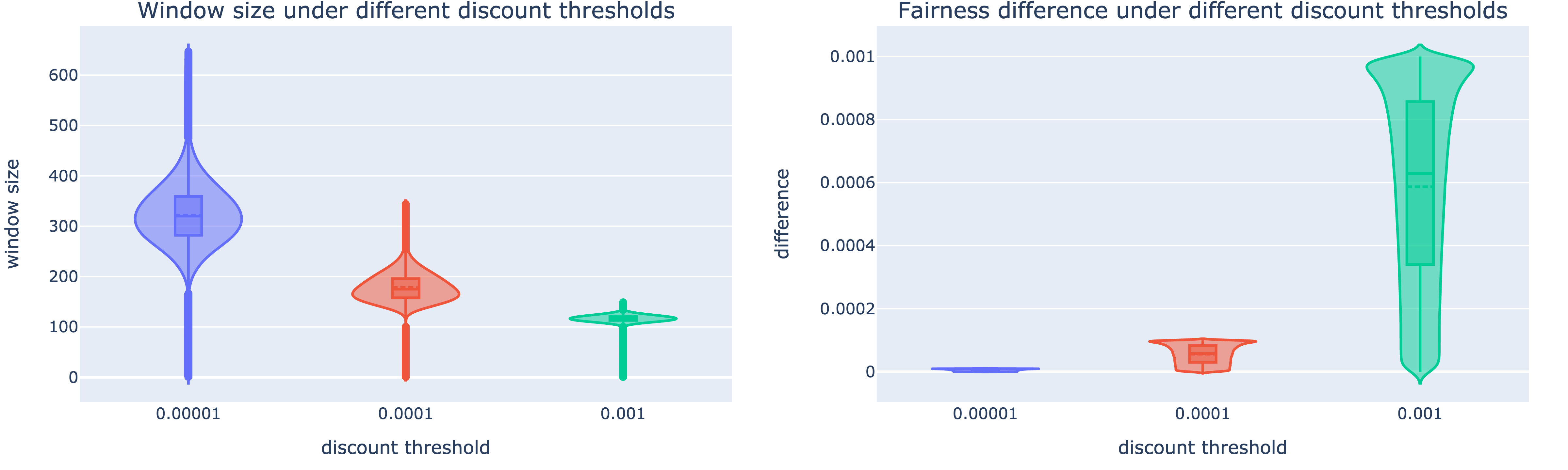}
    \caption{History window size and fairness difference of fraud detection policies when optimising the reward (R), statistical parity (SP) and individual fairness (IF). Showing results for histories with different discount thresholds.}
    \label{fig:fraud_threshold}
\end{figure}


\end{document}